\documentclass[10pt,twocolumn,letterpaper]{article}

\usepackage[pagenumbers]{cvpr} % To force page numbers, e.g. for an arXiv version

\usepackage[dvipsnames]{xcolor}

\usepackage{amsthm,amsfonts,mathtools}
\usepackage{soul}
\allowdisplaybreaks
\usepackage{multirow} 
\usepackage{adjustbox}
\usepackage{microtype}
\usepackage[dvipsnames]{xcolor}
\usepackage{comment}
\usepackage{verbatim}

\newcommand{\KL}{{\rm KL}}
\newcommand{\JSD}{{\rm JSD}}
\newcommand{\JD}{{\rm JD}}

\newcommand{\hd}[1]{\vspace{1ex}\noindent \textbf{#1}}

\definecolor{cvprblue}{rgb}{0.21,0.49,0.74}
\usepackage[pagebackref,breaklinks,colorlinks,citecolor=cvprblue]{hyperref}

\def\paperID{38} % *** Enter the Paper ID here
\def\confName{3DV\xspace}
\def\confYear{2026\xspace}

\title{Text-to-3D Generation using Jensen-Shannon Score Distillation}

\author{%
Khoi Do \qquad  Binh-Son Hua 
\\ 
\\
Trinity College Dublin, Ireland
\tt\small \{dokh, binhson.hua\}@tcd.ie
}

\begin{document}
\maketitle
\begin{abstract}
Score distillation sampling is an effective technique to generate 3D models from text prompts, utilizing pre-trained large-scale text-to-image diffusion models as guidance. However, the produced 3D assets tend to be oversaturated, over-smoothed, and have limited diversity. These issues are a result of a reverse Kullback–Leibler (KL) divergence objective, which makes the optimization unstable and results in mode-seeking behavior. In this paper, we derive a bounded score distillation objective based on Jensen-Shannon divergence (JSD), which stabilizes the optimization process and produces high-quality 3D generation. 
JSD can match the generated and target distributions well, therefore mitigating mode seeking. 
We provide a practical implementation of JSD by utilizing the theory of generative adversarial networks to define an approximate objective function for the generator, assuming the discriminator is well-trained. 
By assuming the discriminator follows a log-odds classifier, we propose a minority sampling algorithm to estimate the gradients of our proposed objective, providing a practical implementation for JSD. 
We conduct both theoretical and empirical studies to validate our method. Experimental results on T3Bench demonstrate that our method can produce high-quality and diversified 3D assets. The code is available at {\color{Salmon}\url{https://github.com/KhoiDOO/jsddreamer}}.
\end{abstract}    
\section{Introduction}\label{sec:intro}
\begin{figure*}[!htp]
    \centering
    \includegraphics[width=0.95\linewidth]{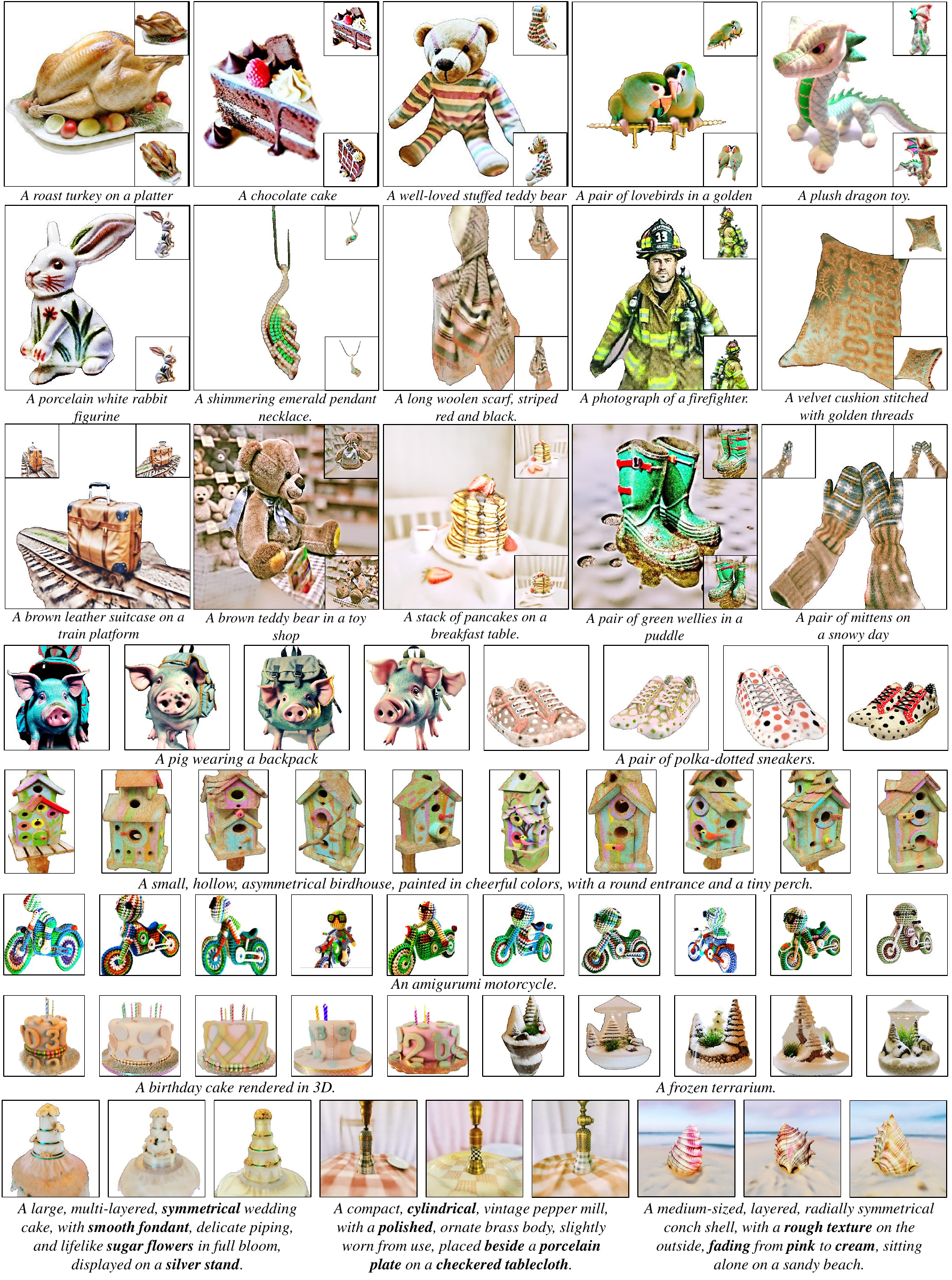}
    \caption{Our method can improve quality and diversity generation with multiple seeds starting from a fixed initialization point.}
    \label{fig:teaser}
\end{figure*}

Text-to-3D generation has become an impactful and leading research field in computer vision, contributing to various applications. Creating high-quality 3D content with view consistency and diversity is resource-intensive, making automated 3D generation a crucial research goal. The developments in the neural radiance field \cite{glnerf, nerf}, Gaussian splatting~\cite{gs, half_gs}, and multimodal latent diffusion models (LDM) \cite{sd, controlnet}, have driven substantial advancements in generating imaginative 3D content from text prompts \cite{sdi, luciddreamer}. 

A straightforward but expensive approach to generating 3D assets given a single text prompt is to train a large-scale generative model on a large-scale 3D shape dataset. 
Another approach is to learn 3D assets by distilling from a large pre-trained model. Score distillation sampling (SDS) utilizes a pre-trained model to learn a neural network (particle), which can synthesize different views of an object \cite{sds}. SDS optimizes a KL divergence~\cite{kld} (KLD) between Gaussian distributions in the forward and backward processes in LDM. Owing to the integration of a 2D pre-trained model, optimizing the KLD leads to an improved 3D representation with more consistent views.

Nevertheless, as highlighted in \cite{vsd, asd, sdi}, SDS tends to produce 3D assets that are oversaturating, oversmoothing, and lack diversity. Other existing approaches aim to produce higher quality 3D assets by utilizing variational score distillation \cite{vsd, asd}, multi-stage training strategies \cite{vp3d, scaledreamer, jointdreamer}. However, those methods are highly costly in computation, which requires fine-tuning pre-trained models or performing mesh extraction or texture fine-tuning. 

Mode-seeking in SDS can be attributed to the asymmetry of the KLD~\cite{kld}, which matches the Gaussians in the forward process to modes of the score functions in a diffusion process~\cite{sds}. 
To mitigate this problem, one potential idea is to symmetrize the objective function, as inspired by the literature of Generative Adversarial Networks (GANs)~\cite{gan, wgan, wgan-gp}. 
We propose to use Jensen-Shannon divergence \cite{jsd} (JSD), a bounded divergence, as our objective function.
Multiple variants of JSD exist, but it remains challenging to perform optimizations using JSD as it requires estimating a mixture of the probability density in JSD. 
Some variants of JSD allow such estimation, such as Geometric JSD, the mixture density of which can be logarithmically derived. However, this objective is unbounded and, therefore, unstable to optimize. 
To implement JSD in practice, we aim to approximate JSD instead. 
Our key insight is to define a discriminator based on log-odds classifiers \cite{bishop} and use minority sampling \cite{dont-play-fav} to perform distillation with an approximated JSD derived via GAN criterion. We show that this technique can well approximate JSD, leading to score distillation with improved stability, enabling the optimization to converge to different modes on the latent manifold.

To support our theoretical derivations, we conduct empirical analysis on a toy dataset by training a toy diffusion model and performing score distillation toward a specific cluster. The obtained results illustrate that the gradient of our proposed method is more stable than SDS owing to an effective control variate, which is positively correlated with the estimated noise. These experiments also show that our method can enhance the diversity of the generated 3D objects.  

We evaluate our method on the recently introduced T3Bench \cite{t3bench}, including a wide range of prompts for the 3D generation task. Our quantitative comparisons with state-of-the-art text-to-3D methods show that our method can generate high-quality 3D assets with a strong alignment with the given prompt. Our contribution can be summarized as follows:
\begin{itemize}
    \item We use Jensen-Shannon divergence (JSD) for score distillation and estimate JSD by leveraging GAN theories and a minority sampling technique.
    \item We validate our theory with empirical experiments, training a toy diffusion model on a toy Gaussian dataset and comparing SDS with our method, confirming the stability and diversity of the generated samples.
    \item We conduct evaluations on common benchmarks and show that our proposed method can generate high-fidelity and diverse 3D assets.  
\end{itemize}

\section{Related Works}\label{sec:related-work}

\hd{Text-to-image generation.} Early text-to-image generation is based on Generative Adversarial Networks (GAN) \cite{gan}, where images are generated conditioned on a textual inputs. The objective criterion in ordinary GAN is Jensen-Shannon Divergence~\cite{jsd} (JSD), approximated by a min-max game optimization. This approach results in mode collapse due to the discontinuous region between the generator and discriminator distribution~\cite{gan-mode-collapse-theory}. To address this problem, alternative divergences~\cite{fgan, wgan, wgan-gp, imp-wgan-gp}, multiple generators~\cite{madgan, mgan, megan, mg-gan, pgan, mclgan}, manifold learning~\cite{mggan, mlbgan, omeegan}, and score matching~\cite{diff-gan, smart} are proposed to match the fake and real distributions. Particularly, approaches that bridge ordinary GAN and score generative models \cite{ddpm, sbgm} achieve SOTA results by diffusing all data points to the same manifold, thus making JSD continuous everywhere~\cite{diff-gan}, while metric-based distance is not always converged~\cite{swgan,gan-converge}. Diffusion models~\cite{diff, sbgm, sd, controlnet}, otherwise learn the relationship between image and textual distribution via a stochastic denoising process, improving high fidelity and diversified generation. Recently, there has been research aiming to improve diversity by guiding the estimated score toward low-density regions in diffusion models~\cite{dont-play-fav, dream-sampler, self-guide}.

\hd{Text-to-3D generation.} 3D content can be generated by several methods. Based on feed-forward inference, 3D content can be generated by a reconstruction model trained on large-scale datasets~\cite{lgm, li2024instantd, richdreamer, latte3d, cad, trellis}, which comes at a cost of extensive annotated data and computational resources. Distillation-based methods, otherwise, optimize a 3D representation to learn an asset aligned with the text prompts from a pre-trained text-to-image model~\cite {sds, dreamfield}. However, these methods are per-prompt optimization, thus requiring a large amount of resources in time and computation. Amortized optimization~\cite{att3d,scaledreamer} trains a unified model on many prompts and 3D asset pairs. Besides, score distillation methods also focus on improving quality~\cite{vsd,asd,luciddreamer,sdi,isd,recdreamer}, view consistency~\cite{mvdream,scaledreamer,jointdreamer,coser,sculpt3d}, diversity~\cite{vsd,asd,sdi,isd}, and faster distillation~\cite{compgs} of 3D asset generation. 
Recently, Adversarial Score Distillation~\cite{asd} (ASD) bridges Variational Score Distillation~\cite{vsd} (VSD) and GAN theory to perform score distillation based on Wasserstein Probability Flow via alternative training. However, the $\ell_1$ transport cost, which restricts the discriminator (e.g., LORA~\cite{lora}) to be 1-Lipschitz~\cite{basic-real-analysis}, is shown not to always converge~\cite {swgan,gan-converge}. This circumstance leads to uncontrollable artifacts and low-quality features in their results. Our approach also leverages GAN theories relevant to Jensen-Shannon divergence, enabling the generated and target distributions to lie on the same support space~\cite{diff-gan,smart}, which returns more stable gradients for 3D generation. 
Our method is also related to variance reduction techniques for score distillation using control variates~\cite{anonymous2023taming,wang2023steindreamer,tang2024ssd,cfd}, but our derivation is via the JSD objective and GAN theories.

\section{Backgrounds}\label{sec:background}

\subsection{Score Distillation Sampling}

Score distillation sampling (SDS)~\cite{sds} has shown great promise in text-to-3D generation by distilling pre-trained large-scale text-to-image diffusion models. SDS optimizes a 3D model parameterized by $\theta\sim p(\Theta)$ by score distillation gradients derived from a large pre-trained model~\cite{sd, controlnet}. Particularly, given text prompt $y$ and the rendered image $\hat{x}_0 = g(\theta, c)$, where $g$ and $c$ are render function and camera pose, the SDS loss function can be written as:
\begin{align}
    \mathcal{L}_{\rm SDS} = \mathbb{E}_{t, \epsilon}\Big[w(t)\frac{\sigma_t}{\alpha_t}\KL(q(\hat{x}_t|\hat{x}_0)\|p_\psi(\hat{x}_t|y)\Big].
\end{align}
The gradient estimated through the score model is shown in Eq. \eqref{eq:sds}, where $\hat{\epsilon}_\psi(\hat{x}_t, y)$ and $\epsilon$ are the estimated noise and the control variate.
\begin{align}\label{eq:sds}
    \nabla_\theta \mathcal{L}_{\rm SDS} = \mathbb{E}_{t, \epsilon}\Big[w(t)(\hat{\epsilon}_\psi(\hat{x}_t, y) - \epsilon)\frac{\partial \hat{x}_0}{\partial\theta}\Big].
\end{align}
It is commonly known that SDS often suffers from over-saturation, over-smoothing, and low-diversity problems. 
Previous work~\cite{asd,vsd,latentnerf} attributed these phenomena to the objective based on the reverse KLD. 
It is therefore beneficial to explore variants of KLD to seek improved convergence and stability in the generation process. 

\subsection{KL Divergence Symmetrization}
KL Divergence \cite{kld} (KLD) or relative entropy is the most fundamental distance. KLD is an asymmetric distance (i.e., $\KL(p, q) \neq \KL(q, p)~ \forall p, q$), which is unbounded and may be infinite. 
\begin{align}
    \KL(p, q) = \sum p\log(p/q),
\end{align}
where $p$ and $q$ are two arbitrary distributions. Jeffreys Divergence \cite{jd} (JD) is a symmetric divergence combining forward and reverse KLD: 
\begin{align}
    \JD(p, q) = \sum \Big[p\log(p/q) + q\log(p/q)\Big].
\end{align}
However, due to the upper-unbounded and numerical instability characteristics, optimizing JD objectives is challenging.  
Another popular symmetrization of the KLD is the Jensen-Shannon Divergence \cite{jsd} (JSD), which can be defined as follows:
\begin{align}
    \JSD(p, q) = \frac{1}{2}\sum \Big[p\log\frac{2p}{p + q} + q\log\frac{2q}{p + q}\Big].
\end{align}
JSD is naturally lower-bounded and upper-bounded within $(0, \log_b2)$ whose base is $b$. 
In the literature of generative adversarial networks, JSD was used to learn diversified  generators~\cite{gan,wgan}.
Given the boundedness of JSD, we aim to utilize JSD for our 3D generation. 
A visualization of all divergences is provided in the supplementary material. 

\section{Methodology}\label{sec:methodology}

\begin{figure*}
    \centering
    \includegraphics[width=\linewidth]{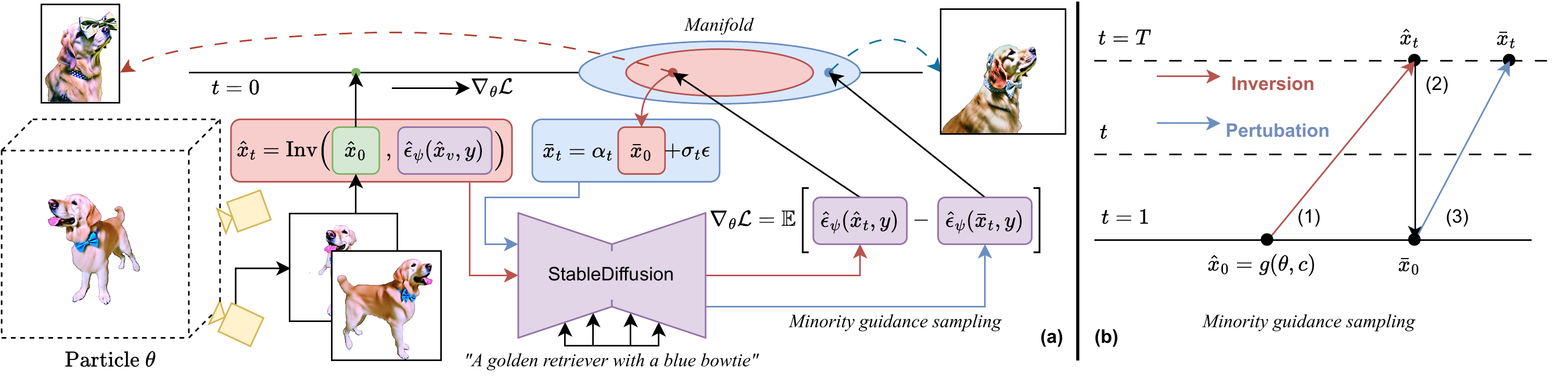}
    \caption{\textbf{a) Methodology overview}. Initially, an image $\hat{x}_0$ is generated via a render function $g(\theta, c)$. We obtain a noisy latent of common mode $\hat{x}_t$ conditioned on a text prompt $y$ ({\color{Salmon}{$\blacksquare$}}) by using the inversion technique. To gather low-density ({\color{CornflowerBlue}{$\blacksquare$}}) samples, we acquire $\bar{x}_0$ via the reverse process, then diffuse it by using SDE. The estimated score of both high- and low-density samples guides $\theta$ toward convergence. \textbf{b)~Minority guidance sampling}. The process of obtaining minority sample $\bar{x}_t$ in three steps: (1)~DDIM inversion to map $\hat{x}_0$ to $\hat{x}_t$, (2)~reverse sampling from $\hat{x}_t$ to obtain $\bar{x}_0$, and (3)~a random diffusion to obtain $\bar{x}_t$.}
    \label{fig:method}
    \vspace{-10pt}
\end{figure*}

\subsection{Jensen-Shannon Divergence Distillation}
We propose to use JSD as the objective function for text-to-3D generation:
\begin{align}
    \mathcal{L}_{\mathrm{JSD}} &\coloneqq \mathbb{E}_{t, \epsilon}\Big[w(t)\frac{\sigma_t}{\alpha_t}\JSD(q(\hat{x}_t|\hat{x}_0)\|p_\psi(\hat{x}_t|y)\Big] 
\end{align} 
JSD differs from reverse KL in the following properties. 
\\
\noindent\textbf{Boundedness.} KLD is unstable and less robust to noise due to its unboundedness~\cite{jsdg-kl-unstable,jsdg-noisy}. KLD can reach extreme values, producing unstable gradients during training, thus preventing $\theta$ from converging to an optimal solution~\cite{fd-vi}. JSD~\cite{jsd}, otherwise, has a bounded and thus more numerically stable loss landscape than KLD, which stabilizes the training procedure~\cite{stable-loss}, encouraging $\theta$ to reach an optimal solution~\cite{bounded-smooth-loss}. We provided a boundedness analysis in the supplementary material, which shows that JSD is in fact a lower bound of the reverse KL objective: 
\begin{align}\label{eq:sds-jsd}
    \mathcal{L}_{\mathrm{JSD}} \leq \mathbb{E}_{t, \epsilon}\Big[w(t)\frac{\sigma_t}{\alpha_t}\KL(q(\hat{x}_t|\hat{x}_0)\|p_\psi(\hat{x}_t|y)\Big]. \notag
\end{align} 

\noindent\textbf{Mode coverages.} SDS faces a problem of low-diversity generation due to mode collapses from the reverse KLD formulation~\cite{vsd,asd}. 
%Unlike reverse KLD, JSD matches well distributions~\cite{gan, diff-gan}. 
Unlike reserve KLD, JSD can deal with null mass probability and satisfies the triangle inequality~\cite{jsd-triangle}, therefore being a good metric distance to match two insignificantly non-overlapping distributions. JSD is, therefore, a potential divergence to mitigate mode collapses, enabling the learning of diversified 3D representations. 

\noindent\textbf{Challenges.}
Despite JSD's benefits, applying JSD for score distillation remains a challenge because sampling from the mixture distribution in JSD is not straightforward, as it is the arithmetic mean $(q(\hat{x}_t|x) + p_\psi(\hat{x}_t|y))/2$ that cannot be logarithmically derived. An alternative way is to use geometric mean to perform mixture distribution~\cite{jsdg,jsdg-star,jsds} $\sqrt{q(\hat{x}_t|x)p_\psi(\hat{x}_t|y)}$, which can be derived easily by the logarithmic function. However, this geometric JSD is unbounded, being an upper-bounded version of the ordinary JSD~\cite{jsds}. In this paper, we instead derive a new objective approximating JSD for score distillation.

\subsection{A Discriminator-based Objective}
We leverage the GAN~\cite{gan, gan-mode-collapse-theory} training strategy to approximate JSD. In GAN learning theory, there are two steps of training, including training a discriminator $\mathcal{D}$ (a binary classifier) and a generator $\mathcal{G}$, which are both parameterized models. This learning process is a minimax two-player game:
\begin{align}
% \label{eq:gan}
V(\mathcal{G}, \mathcal{D}) &= \int_x \Big[ p_{data}(x) \log \mathcal{D}(x)
\notag\\ &
+ p_{\mathcal{G}}(x) \log(1 - \mathcal{D}(x)) \Big] dx.
\end{align}
When $\mathcal{D}$ reaches the optimal solution, the criterion $V(\mathcal{G}, \mathcal{D})$ with respect to $\mathcal{G}$ is equivalent to a JSD objective function~\cite{gan, gan-mode-collapse-theory}. 
Following the vanilla GAN~\cite{gan}, to train the generator, instead of minimizing $\log(1 - D(x))$, we minimize $-\log D(x)$, resulting in the following objective:
\begin{align}
    \mathcal{L}_{\mathcal{G}}(\theta) &= \mathbb{E}_{t, \epsilon}\Big[- \log D(\hat{x}_t; y)\Big] 
\end{align}
To define the discriminator, we drew inspiration from the fact that our pre-trained text-to-image diffusion model can be regarded as a robust classifier~\cite{DAS, diff-zero-shot, diff-certifiable-zero-shot}, making the model naturally a discriminator. We assume that our discriminator is optimal, and formulate it such that the discriminator follows a log-odds binary classifier~\cite{bishop}:
\begin{align}\label{eq:minmax}
    \mathcal{D}(\hat{x}_t; y) = \frac{p_\psi(y|\hat{x}_t)}{1 - p_\psi(y|\hat{x}_t)},
\end{align}
where $p_\psi(y|\hat{x}_t)$ is the likelihood of $\hat{x}_t$ classified as text prompt $y$. In this discriminator, we note that the term $(1 - p_\psi(y|\hat{x}_t))$ cannot be directly computed. For simplicity, we denote this term by a density function $p(\phi| \hat{x}_t) \approx 1 - p_\psi(y|\hat{x}_t)$. We assume the input prompt $y \in \Omega$ (space of all prompts). Given a prompt $y$, we assume multiple solutions $\hat{x}_t$, aligned to $y$ due to diversity. Such solutions can be captioned by an extended set of prompts $\mathcal{Y}$ such that $y \in \mathcal{Y} \subset \Omega$. The complementary prompt $\phi$ can be defined by
$\phi \in \Omega \setminus \mathcal{Y}$, capturing irrelevant descriptions of $\hat{x}_t$. We define such irrelevance between $y$ and $\phi$ via a complement density $p(\phi|\hat{x}_t) = 1 - p(y|\hat{x}_t)$. Expanding the logarithm function and taking the derivative, the gradient of our objective function becomes:
\begin{flalign}
    \nabla_\theta\mathcal{L}_\mathcal{G} = \mathbb{E}_{t, \epsilon}\Big[\nabla_\theta\log p_\psi(\phi| \hat{x}_t) - \nabla_\theta\log p_\psi(y | \hat{x}_t) \Big].
\end{flalign}
Let us now proceed to relate this gradient to the score function of the pre-trained text-to-image model below.

\subsection{Gradient Approximation}

Let us proceed to derive each term in the gradient. The right term can be factorized using the Bayesian theorem, such that $\nabla_\theta\log p_\psi(y|\hat{x}_t) \propto \nabla_\theta\log p_\psi(\hat{x}_t|y) - \nabla_\theta\log p_\psi(\hat{x}_t|\oslash)$. 
The left term can be derived via multiclass generalization of the logistic sigmoid \cite{bishop} to yield:
\begin{flalign}
& \nabla_\theta\log p_\psi(\phi|\hat{x}_t) 
\!\propto\!
\nabla_\theta[\log p_\psi(\bar{x}_t|y)\! - \!\log p_\psi(\bar{x}_t|\oslash)].
\end{flalign}
This formulation means that instead of using the prompt $\phi$ and the noised image $\hat{x}_t$ for estimating the gradient, we estimate a minority sample $\bar{x}_t$ from $\hat{x}_t$, such that $p(\phi | \hat{x}_t) \approx p(y | \bar{x}_t)$ and therefore we use the same prompt $y$ to approximate the gradients. The minority sampling process is illustrated in Fig.~\ref{fig:method}. 

Our minority sampling has three steps. 
(1) We first apply DDIM inversion~\cite{ddim} to estimate $\hat{x}_t$ from the rendered image $\hat{x}_0$, preserving the conditioning on text prompt $y$.
(2) We then estimate the denoised rendered image $\bar{x}_0$ from $\hat{x}_t$ by following the reverse process $\bar{x}_0 = (1/\alpha_t)\big(\hat{x}_t - \sigma_t \epsilon_{\psi}(\hat{x}_t, y)\big)$.
(3) We then diffuse $\bar{x}_0$ to obtain the perturbed sample $\bar{x}_t = \alpha_t\bar{x}_0 + \sigma_t \epsilon$ where $\epsilon\sim\mathcal{N}(0, \mathbb{I})$ is random noise.
As the perturbation to produce $\bar{x}_0$ is random, the noised sample $\bar{x}_t$ becomes less well aligned with the original prompt $y$, meaning that the density $p_\psi(y | \bar{x}_t)$ is low~\cite{dont-play-fav} and hence a good approximation to $p(\phi | \hat{x}_t)$. The full derivation can be found in the supplementary material. 

The estimated gradient is therefore:
\begin{align}\label{eq:simple-jsd}
    \nabla_\theta\mathcal{L} = \mathbb{E}_{t, \epsilon}\Big[w(t)\frac{\alpha_t}{\sigma_t}(\hat{\epsilon}_\psi(\hat{x}_t, y) - \hat{\epsilon}_\psi(\bar{x}_t, y))\frac{\partial\hat{x}_0}{\partial\theta}\Big],
\end{align}
where $\hat{\epsilon}_\psi(\hat{x}_t, y)$ and $\hat{\epsilon}_\psi(\bar{x}_t, y)$ are the predicted scores. It is worth noting that it is natural to integrate classifier guidance scale~\cite{ho2021classifierfree}, denoted by $s$, into this gradient by representing
$\hat{\epsilon}_\psi(\hat{x}_t, y) = \epsilon_\psi(\hat{x}_t, \oslash) + s(\epsilon_\psi(\hat{x}_t, y) - \epsilon_\psi(\hat{x}_t, \oslash))$ and $\hat{\epsilon}_\psi(\bar{x}_t, y) = \epsilon_\psi(\bar{x}_t, \oslash) + s(\epsilon_\psi(\bar{x}_t, y) - \epsilon_\psi(\bar{x}_t, \oslash))$.
It can be observed that the gradient formulated in Eq.~\ref{eq:simple-jsd} leads to increased diversity of the generated samples because the estimated noise $\hat{\epsilon}_\psi(\bar{x}_t, y)$ acts as an effective control variate. The detailed derivation is provided in the supplementary material. In the next section, we will present an empirical analysis of this gradient, connecting it to the control variate perspective on improved optimization.
\section{Experimental Results}
\subsection{Empirical Analysis}\label{sec:ana}
We conducted empirical experiments using a toy diffusion model to analyze the optimization behavior of our proposed objective. We illustrate the convergence of the optimization, demonstrating its gradient stability and trajectory diversity. 

\begin{figure}[t]
    \centering
    \begin{subfigure}[t]{0.40\linewidth}
        \centering
        \includegraphics[width=\linewidth]{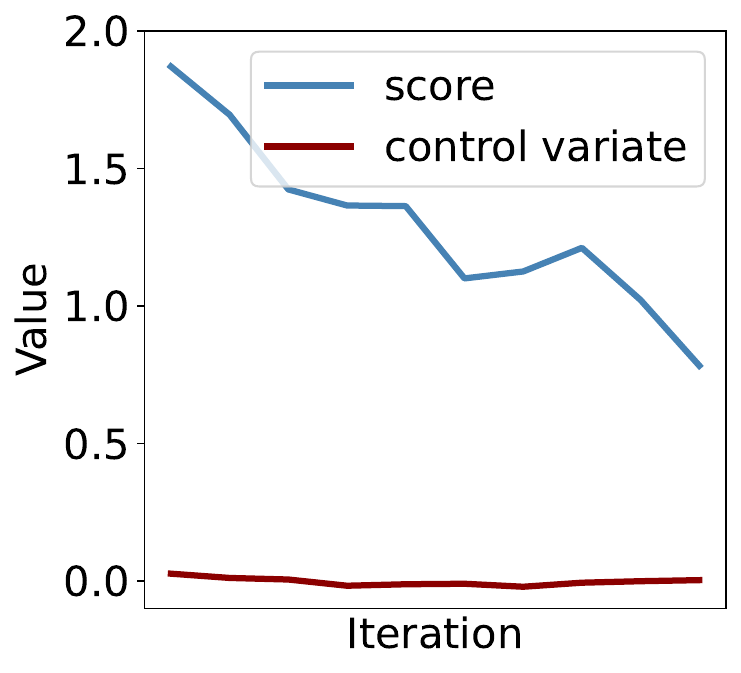}
        \caption{SDS}
        \label{fig:exam-ns-ns-sds}
    \end{subfigure}
    \hspace{1cm}
    \begin{subfigure}[t]{0.40\linewidth}
        \centering
        \includegraphics[width=\linewidth]{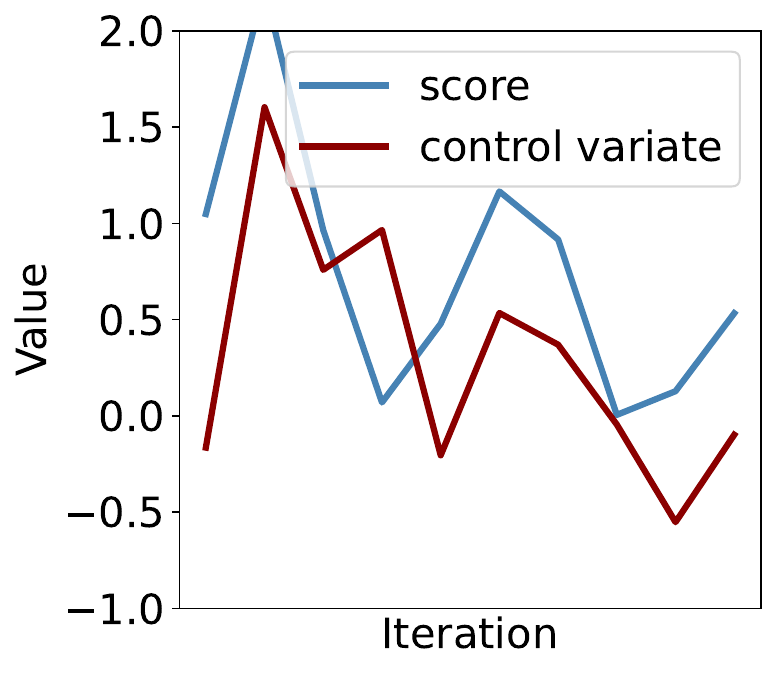}
        \caption{JSD}
        \label{fig:exam-ns-ns-jsd}
    \end{subfigure}
    \caption{Quantitative comparison between estimated noise and control variate in SDS and JSD. It can be seen that our control variate is positively correlated to the estimated noise, hence reducing variances in gradient estimation.}
    \label{fig:exam-ns-ns}
    \vspace{-10pt}
\end{figure}

\hd{Gradient Stability.}
We trained a simple diffusion model on a dataset of eight cluster samples drawn from a mixture of eight two-dimensional Gaussian distributions. We then performed optimization using JSD and SDS to sample a data point toward a specific cluster. Each cluster is considered a class, which will be used for classifier-free guidance during score distillation. We examine the value of the estimated score and the control variate (the left and right term in the gradient in Eq.~\ref{eq:simple-jsd}). It can be observed that both terms are positively correlated, making the gradient values close to zero, hence reducing variances and stabilizing the optimization (Fig.~\ref{fig:exam-ns-ns}). 

\hd{Gradient Trajectory Diversity.}
\begin{figure}[b]
    \centering
    \begin{subfigure}[b]{0.09\textwidth}
        \includegraphics[width=\textwidth]{imgs/analysis/distillation/sds/noaxis_trials_10_start__1.0_1.0_.pdf}
    \end{subfigure}
    \begin{subfigure}[b]{0.09\textwidth}
        \includegraphics[width=\textwidth]{imgs/analysis/distillation/sds/noaxis_trials_100_start__1.0_1.0_.pdf}
    \end{subfigure}
    \begin{subfigure}[b]{0.09\textwidth}
        \includegraphics[width=\textwidth]{imgs/analysis/distillation/sds/noaxis_trials_200_start__1.0_1.0_.pdf}
    \end{subfigure}
    \begin{subfigure}[b]{0.09\textwidth}
        \includegraphics[width=\textwidth]{imgs/analysis/distillation/sds/noaxis_trials_500_start__1.0_1.0_.pdf}
    \end{subfigure}
    \begin{subfigure}[b]{0.09\textwidth}
        \includegraphics[width=\textwidth]{imgs/analysis/distillation/sds/noaxis_trials_1000_start__1.0_1.0_.pdf}
    \end{subfigure}
    \begin{subfigure}[b]{0.09\textwidth}
        \includegraphics[width=\textwidth]{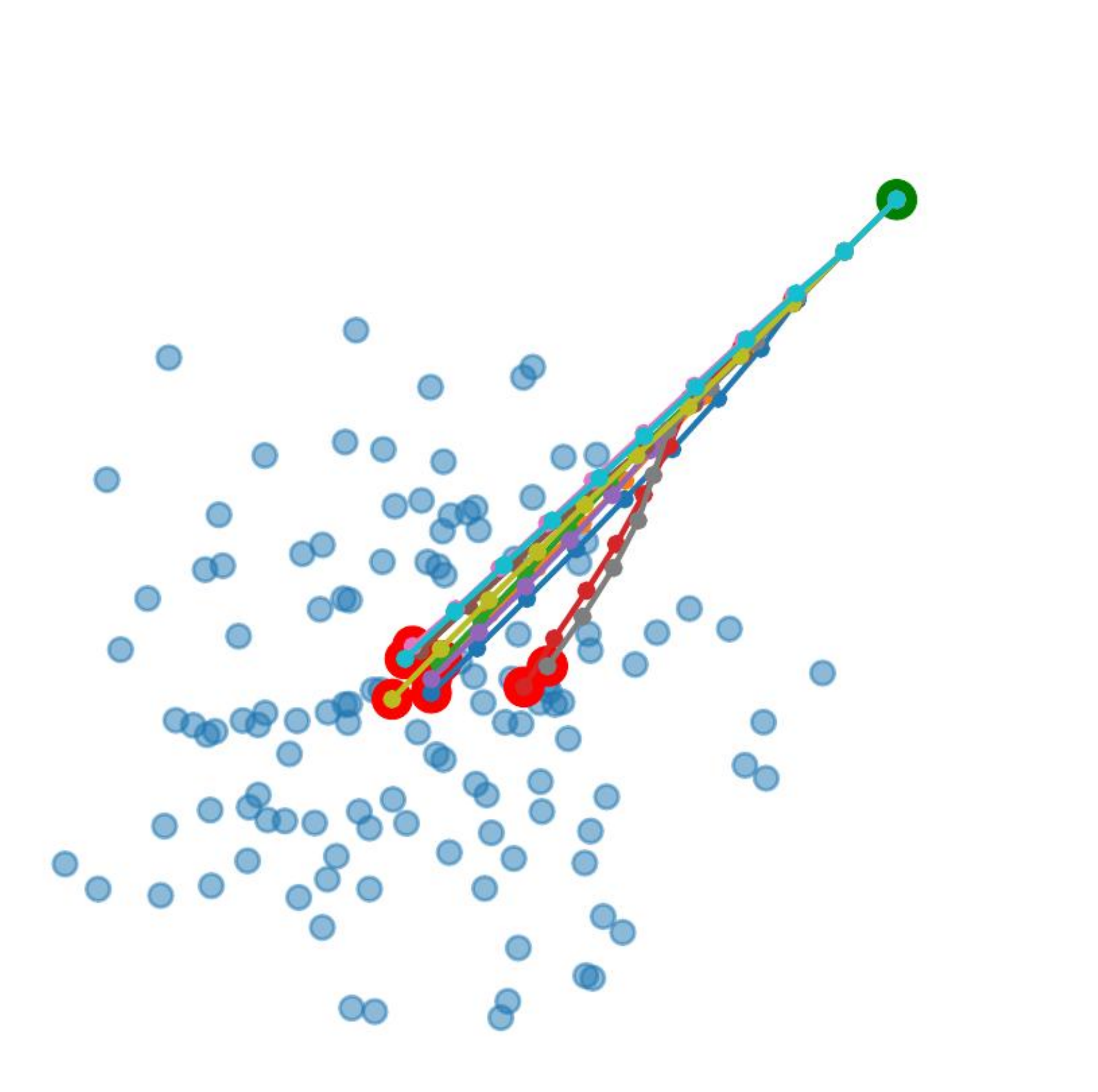}
        \caption{10}
    \end{subfigure}
    \begin{subfigure}[b]{0.09\textwidth}
        \includegraphics[width=\textwidth]{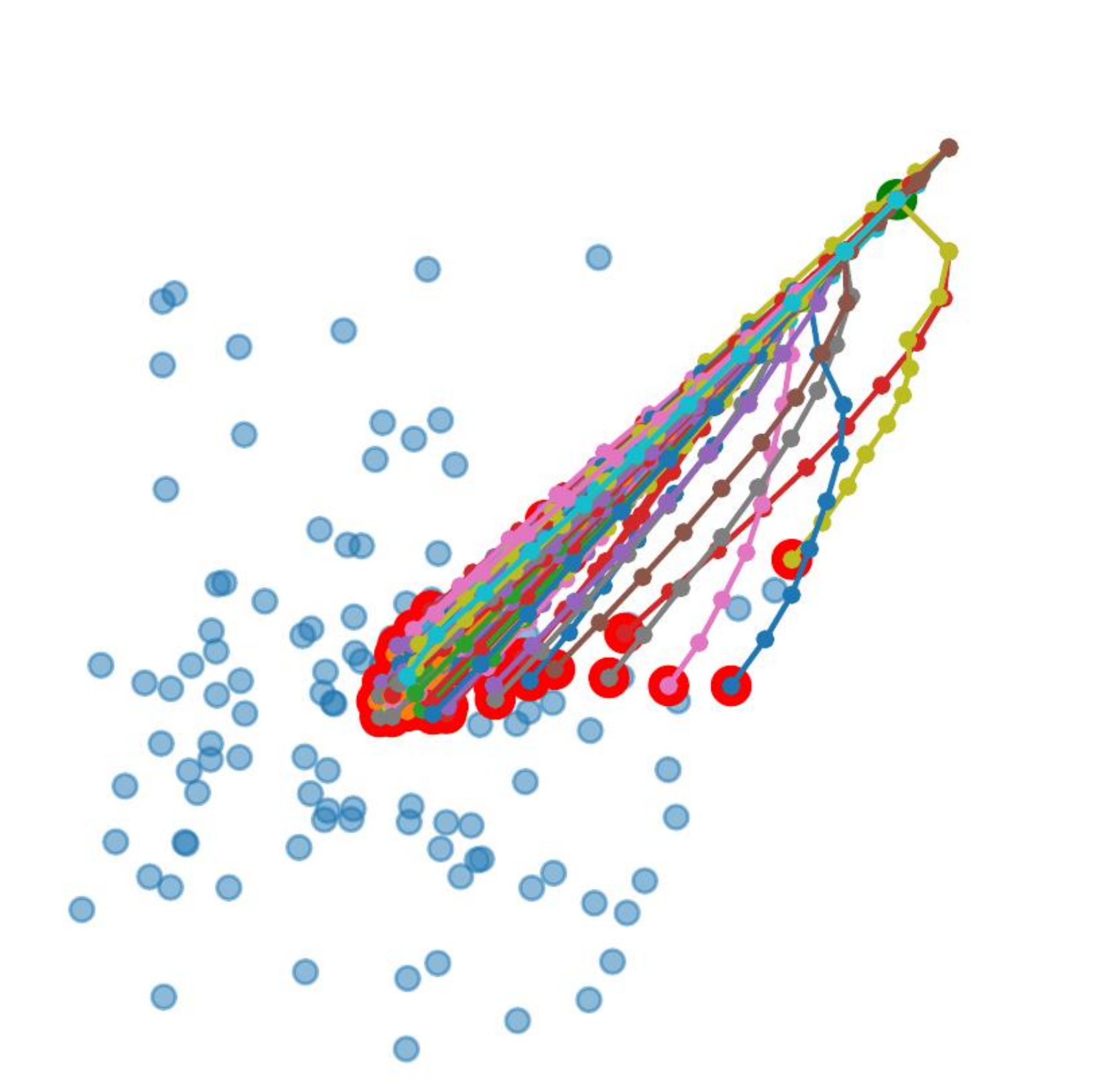}
        \caption{100}
    \end{subfigure}
    \begin{subfigure}[b]{0.09\textwidth}
        \includegraphics[width=\textwidth]{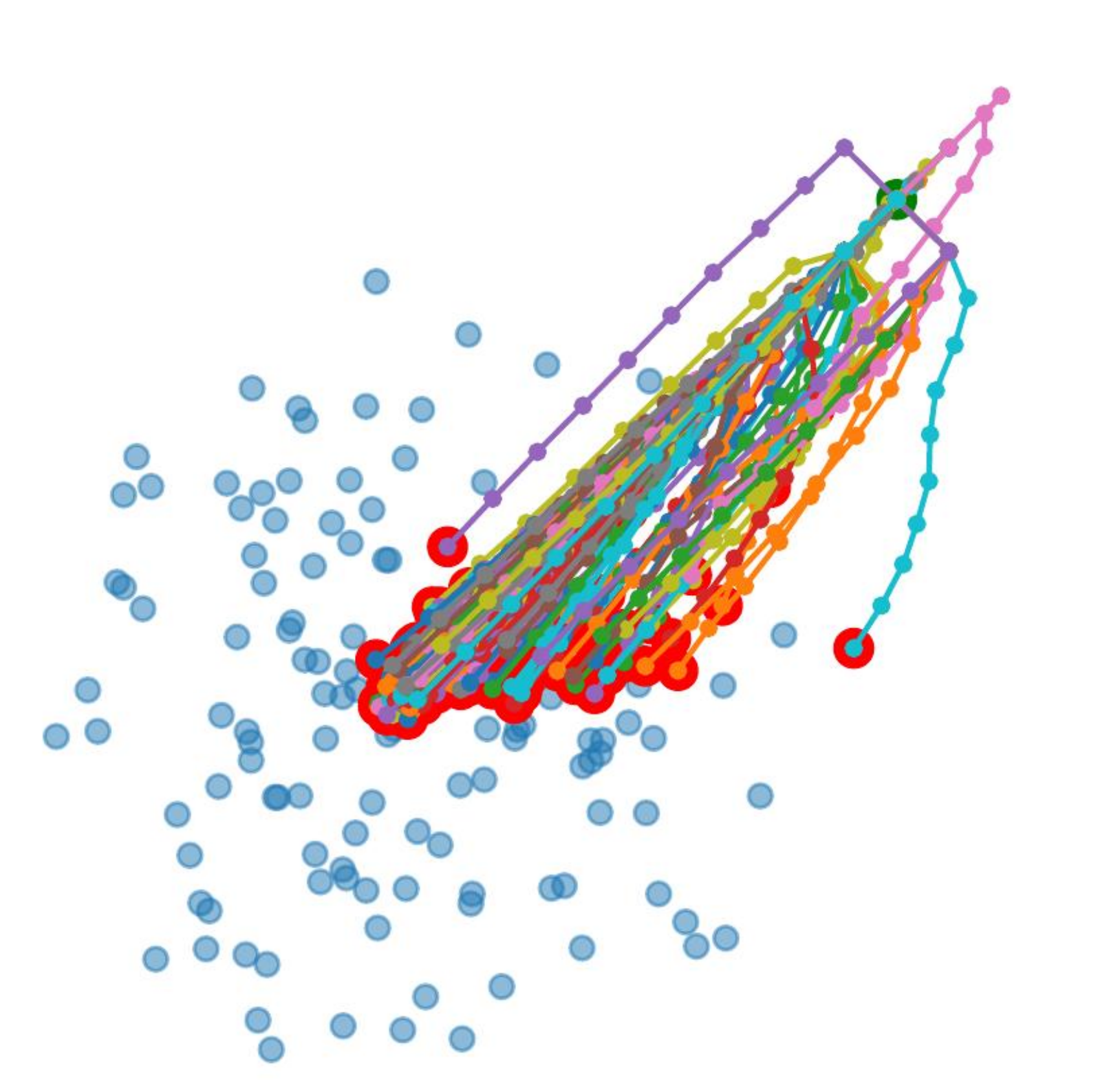}
        \caption{200}
    \end{subfigure}
    \begin{subfigure}[b]{0.09\textwidth}
        \includegraphics[width=\textwidth]{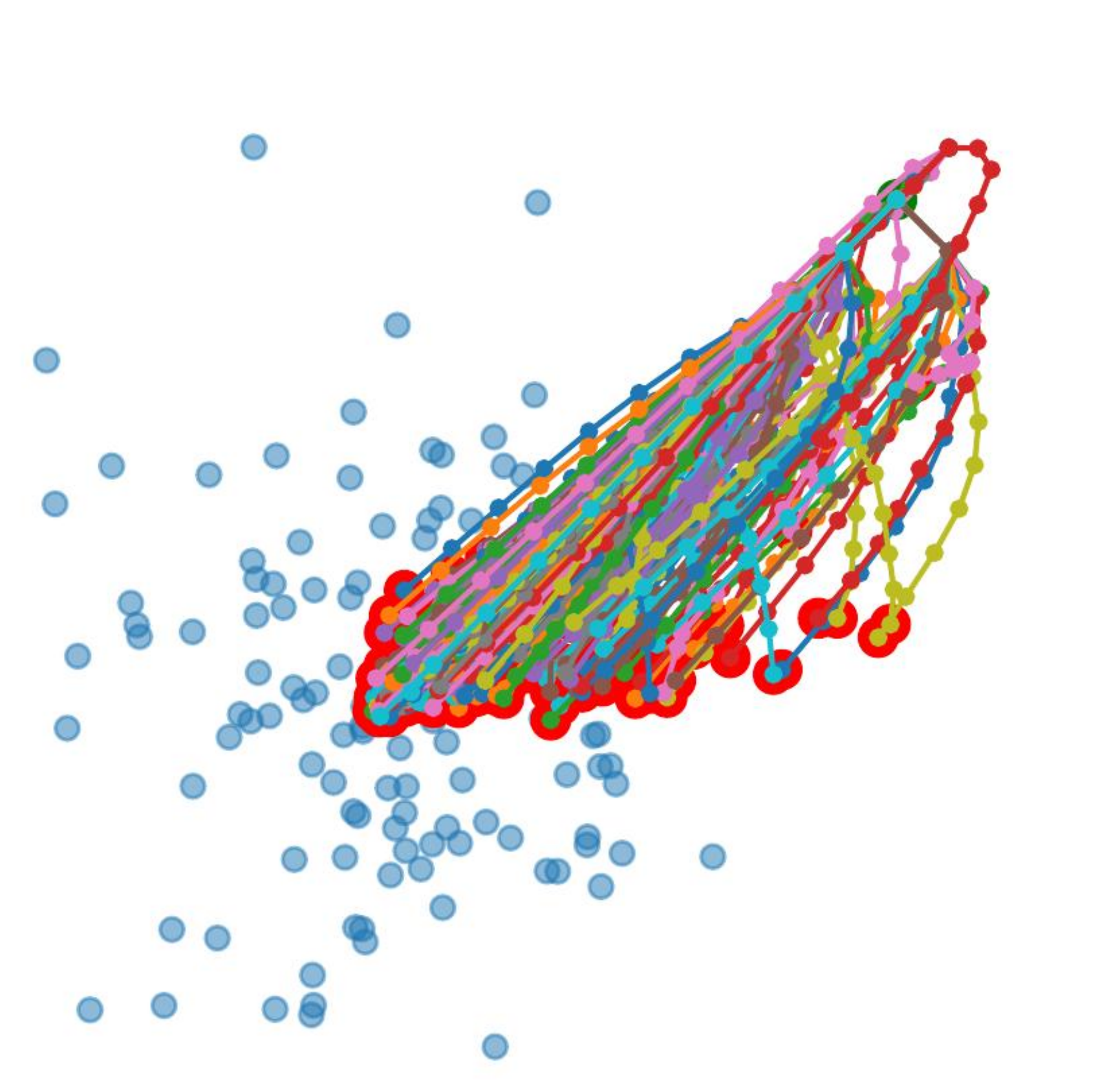}
        \caption{500}
    \end{subfigure}
    \begin{subfigure}[b]{0.09\textwidth}
        \includegraphics[width=\textwidth]{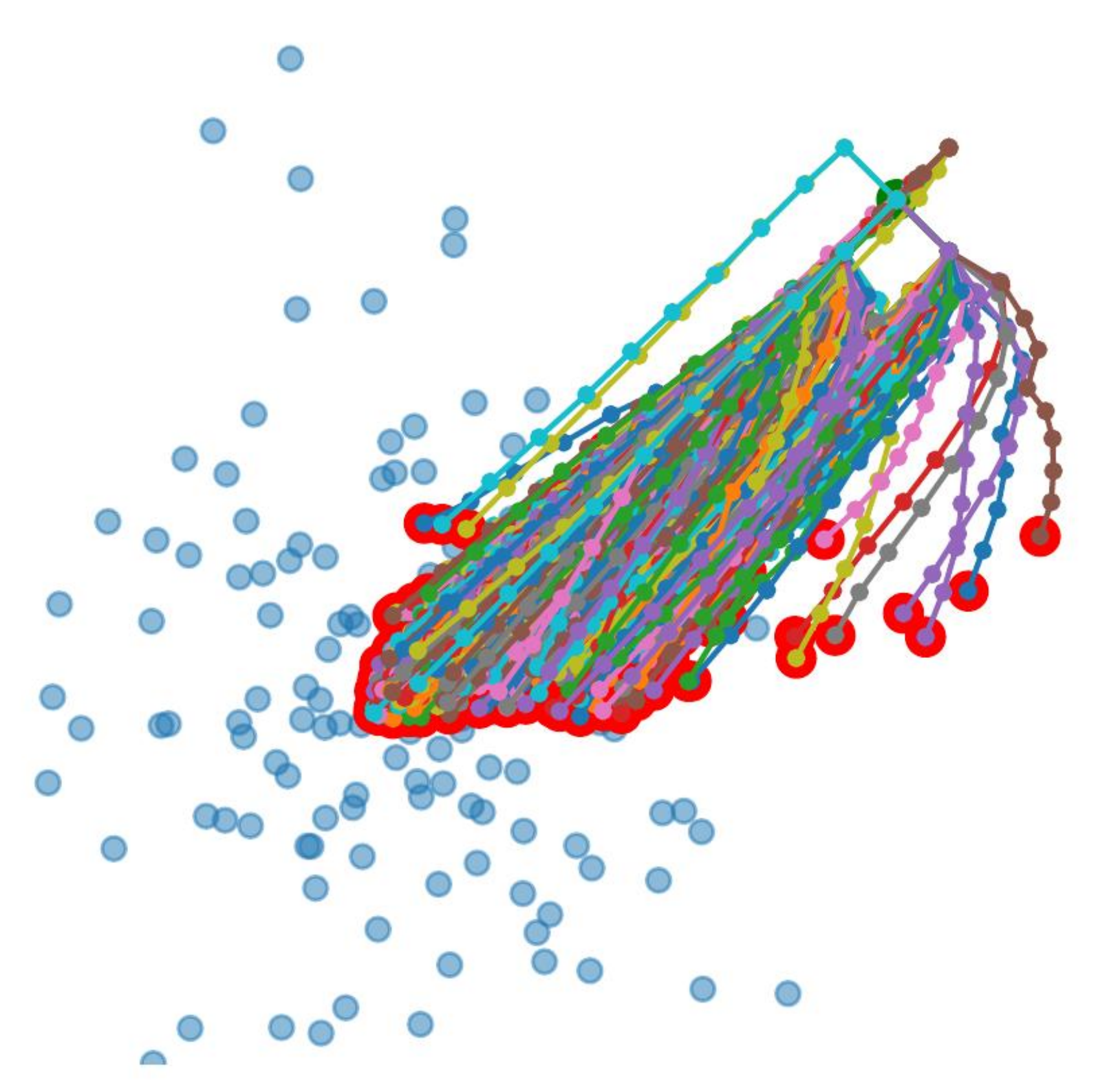}
        \caption{1000}
    \end{subfigure}
    \caption{Optimization behavior of SDS (top row) and our method (bottom row). Each score distillation is performed on a toy diffusion model from a fixed starting point ({\color{teal}{$\bullet$}}), converging to a result ({\color{red}{$\bullet$}}). Each trajectory is initialized with a different random seed. The number of trajectories is from 10 to 1000. More results are in the supplementary material.}
    \label{fig:1ddiff-sample}
    \vspace{-7pt}
\end{figure}
Figure~\ref{fig:1ddiff-sample} illustrates the gradient trajectory from the same initialization, comparing between SDS and our gradients from JSD.  It can be seen that our method has more diverse trajectories, resulting in different modes. 
\subsection{3D Generation Results}\label{sec:exp}

\begin{table*}[!ht]
    \caption{Comparative results for the text-to-3D tasks in T3Bench. The best results are \textbf{bold} while the second best results are \underline{underlined}.}
    \vspace{-5pt}
    \label{tab:t3bench_results}
    \small
    \centering
    \setlength{\tabcolsep}{4.5pt} % Adjust column padding
    \begin{tabular}{lccccccccccc}
        \toprule
         & Conference &\textbf{Time} & \multicolumn{3}{c}{\textbf{Single Object}} & \multicolumn{3}{c}{\textbf{Single Object with Surr}} & \multicolumn{3}{c}{\textbf{Multiple Objects}} \\ 
        \cmidrule{3-12}
        \textbf{Method} & & (mins) & \textbf{Qual.} $\uparrow$ & \textbf{Align.} $\uparrow$ & \textbf{Avg} $\uparrow$
               & \textbf{Qual.} $\uparrow$ & \textbf{Align.} $\uparrow$ & \textbf{Avg} $\uparrow$
               & \textbf{Qual.} $\uparrow$ & \textbf{Align.} $\uparrow$ & \textbf{Avg} $\uparrow$\\ 
        \midrule
        % Dreamfusion \cite{sds} & ICLR 2023 & 30 & 24.9 & 24.0 & 24.4 & 19.3 & 29.8 & 24.6 & 17.3 & 14.8 & 16.1 \\ 
        % Magic3D \cite{magic3d} & CVPR 2023 & 40 & 38.7 & 35.3 & 37.0 & 29.8 & 41.0 & 35.4 & 26.6 & 24.8 & 25.7 \\ 
        LatentNeRF \cite{latentnerf} & CVPR 2023 & 65 & 34.2 & 32.0 & 33.1 & 23.7 & 37.5 & 30.6 & 21.7 & 19.5 & 20.6 \\ 
        % Fantasia3D \cite{fantasia3d} & ICCV 2023 & 45 & 29.2 & 23.5 & 26.4 & 21.9 & 32.0 & 27.0 & 22.7 & 14.3 & 18.5 \\ 
        % SJC \cite{sjc} & CVPR 2023 & 25 & 26.3 & 23.0 & 24.7 & 17.3 & 22.3 & 19.8 & 11.7 & 5.8 & 8.7 \\ 
        VSD \cite{vsd} & NIPS 2023 & 240 & 51.1 & 47.8 & 49.4 & 42.5 & 47.0 & 44.8 & 45.7 & 25.8 & 35.8 \\ 
        MVDream \cite{mvdream} & ICLR 2024 & 30 & 53.2 & 42.3 & 47.8 & 36.3 & 48.5 & 42.4 & 39.0 & 28.5 & 33.8 \\ 
        DreamGaussian \cite{dreamgaussian} & ICLR 2024 & 7 & 19.9 & 19.8 & 19.8 & 10.4 & 17.8 & 14.1 & 12.3 & 9.5 & 10.9 \\
        RichDreamer \cite{richdreamer} & CVPR 2024 & 70 & \underline{57.3} & 40.0 & 48.6 & 43.9 & 42.3 & 43.1 & 34.8 & 22.0 & 28.4 \\ 
        VP3D \cite{vp3d} & CVPR 2024 & - & 54.8 & 52.2 & 53.5 & 45.4 & 50.8 & 48.1 & 49.1 & 31.5 & 40.3 \\ 
        ModeDreamer \cite{isd} & Arxiv 2024 & \underline{40} & 55.4 & 52.6 & 54.0 & \underline{45.7} & \textbf{59.0} & 52.4 & 43.4 & \underline{39.4} & 41.4 \\
        DreamReward$^{*}$~\cite{dreamreward} & ECCV 2024 & \underline{40} & 54.3 & 43.8 & 49.0 & 38.2 & 49.6 & 43.9 & 41.2 & 33.6 & 37.4 \\
        DreamMesh \cite{dreammesh} & ECCV 2024 & 30 & 55.6 & \textbf{53.8} & 54.7 & 43.1 & 54.3 & \underline{48.7} & 47.6 & 30.8 & 39.2 \\ 
        Trellis~\cite{trellis} & CVPR 2025 & $\mathbf{<1}$ & 35.6 & 21.4 & 28.5 & 19.5 & 17.5 & 18.5 & 16.6 & 17.0 & 16.8\\
        CompGS \cite{compgs} & CVPR 2025 & 30 & 55.1 & 52.5 & 53.8 & 43.2 & 46.8 & 45.0 & \textbf{54.2} & 37.9 & \textbf{46.1} \\
        \midrule
        Ours & & 70 & \textbf{58.7} & \underline{53.6} & \textbf{56.1} & \textbf{47.4} & \underline{57.6} & \textbf{52.5} & \underline{51.3} & \textbf{40.2} & \underline{45.7} \\
        \bottomrule
    \end{tabular}
    \vspace{-5pt}
\end{table*}

\hd{Implementation details.} We implement the proposed method on top of the Threestudio \cite{threestudio2023} framework. We perform optimization for 10,000 steps using StableDiffusion\cite{sd}. We set the DDIM inversion \cite{null-text-inversion} steps to $10$. Otherwise, we use a CFG of $13.5$ for all experiments, providing a balance between quality and diversity.

\hd{Benchmark and metrics.} In this paper, we provide results on T3Bench~\cite{t3bench}. T3bench includes 300 prompts, classified into three categories: Single Objects (SO), Single Objects with Surrounding (SOS), and Multiple Objects (MO). To evaluate the fidelity, 3D samples are converted into a mesh form of the level-0 icosahedron and then scored by an ImageReward~\cite{imagereward} model. For asset alignment, an image captioning model (BLIP~\cite{blip2}) is used to obtain captions across multiple views, which GPT4 scores. In the appendix, we also provided benchmark results from GPTEval3D~\cite{gpteval3d}.

\hd{Quantitative results}. Table \ref{tab:t3bench_results} presents the comparisons with SOTA methods. Our method's performance is competitive with DreamMesh~\cite{dreammesh} and CompGS~\cite{compgs}, especially in the SO and SOS categories. For text prompts involving multiple objects, our method achieved the highest alignment assessment score. We also include comparisons with Trellis, a state-of-the-art pretrained 3D generation model~\cite{trellis}. Trellis can rapidly generate 3D models, but tends to underperform score distillation methods. 

\begin{figure*}[!ht]
    \centering
    \includegraphics[width=\linewidth]{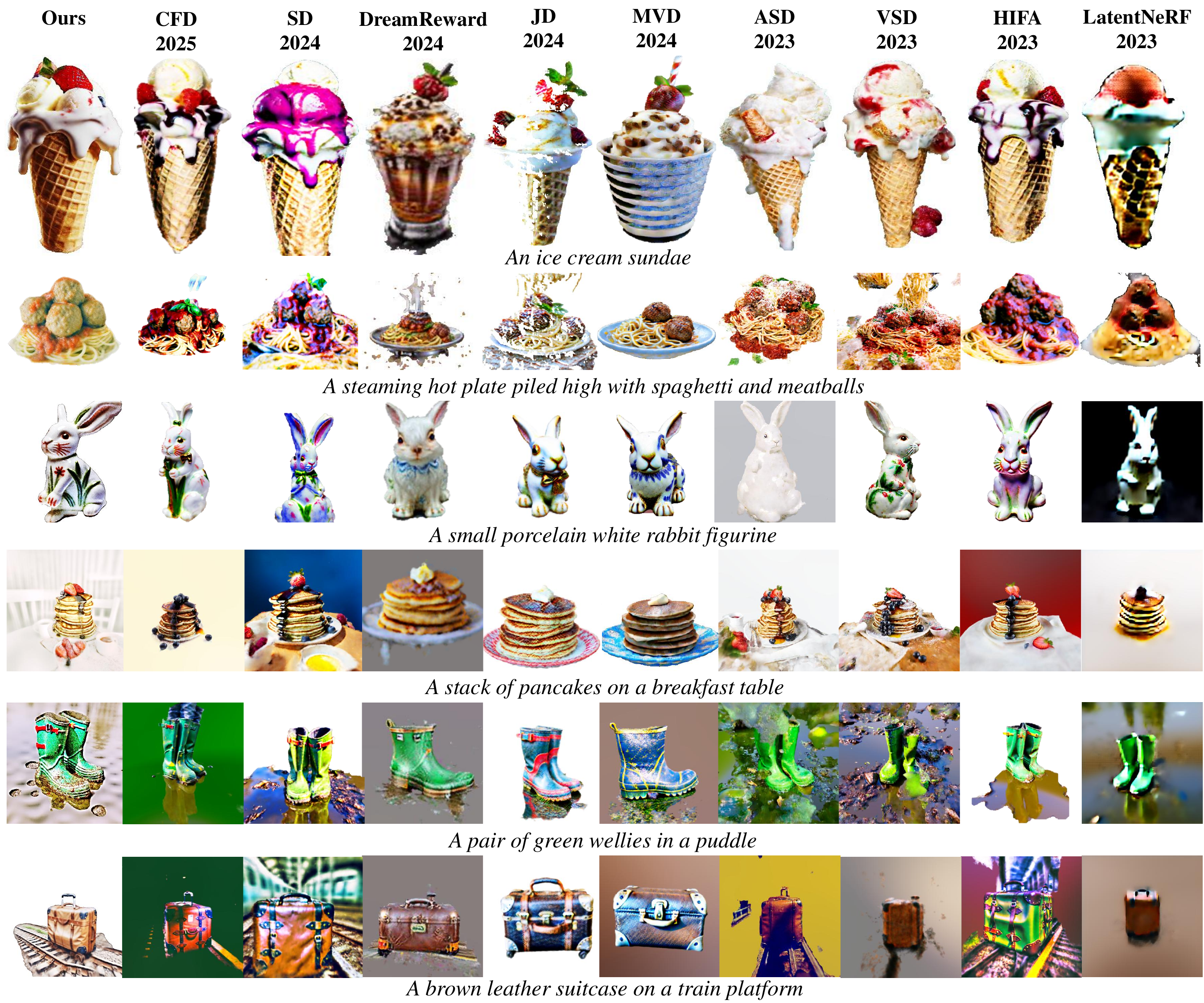}
    \vspace{-5mm}
    \caption{Qualitative comparison with available text-to-3D generation techniques.}
    \label{fig:exp-compare}
    \vspace{-10pt}
\end{figure*}

\hd{Qualitative comparison.} We provide examples to compare our method with SOTA methods, including Consistent Flow Distillation~\cite{cfd} (CFD), ScaleDreamer~\cite{scaledreamer} (SD), DreamReward~\cite{dreamreward}, JointDreamer~\cite{jointdreamer} (JD), MVDream~\cite{mvdream}, ASD~\cite{asd}, ProlificDreamer~\cite{vsd} (VSD), HIFA~\cite{hifa}, and LatentNerf~\cite{latentnerf} (see Figure~\ref{fig:exp-compare}). Overall, our approach can generate higher-quality 3D assets than previous methods. For fair comparison, we used prompts provided by the previous research. In ASD, many artifacts appear around the main objects, which is evidence of the instability of using the Wasserstein metric. For the object with surrounding prompts, ScaleDreamer and ASD are unstable in learning both the foreground and the background. Our approach can learn the main object, the surrounding objects, and the background with good quality. We provide more examples at Appendix.

Our method can produce 3D objects well aligned with the guidance prompt in the alignment assessment. We present various 3D objects generated with long, detailed prompts in Fig.~\ref{fig:teaser} to indicate the strong alignment characteristic of our proposed method. Other methods tend to ignore information in the background, not focus on the surrounding objects, or place those objects in the wrong location.

\begin{figure*}[!ht]
    \centering
    \includegraphics[width=\linewidth]{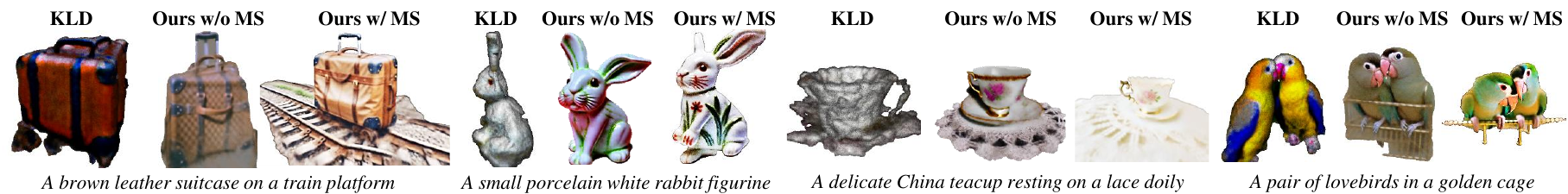}
    \caption{Visual comparison between KLD and our proposed method with and without minority sampling (MS).}
    \label{fig:abs}
    \vspace{-5pt}
\end{figure*}

\begin{figure*}[t]
    \centering
    % \vspace{-10pt}
    \includegraphics[width=\linewidth]{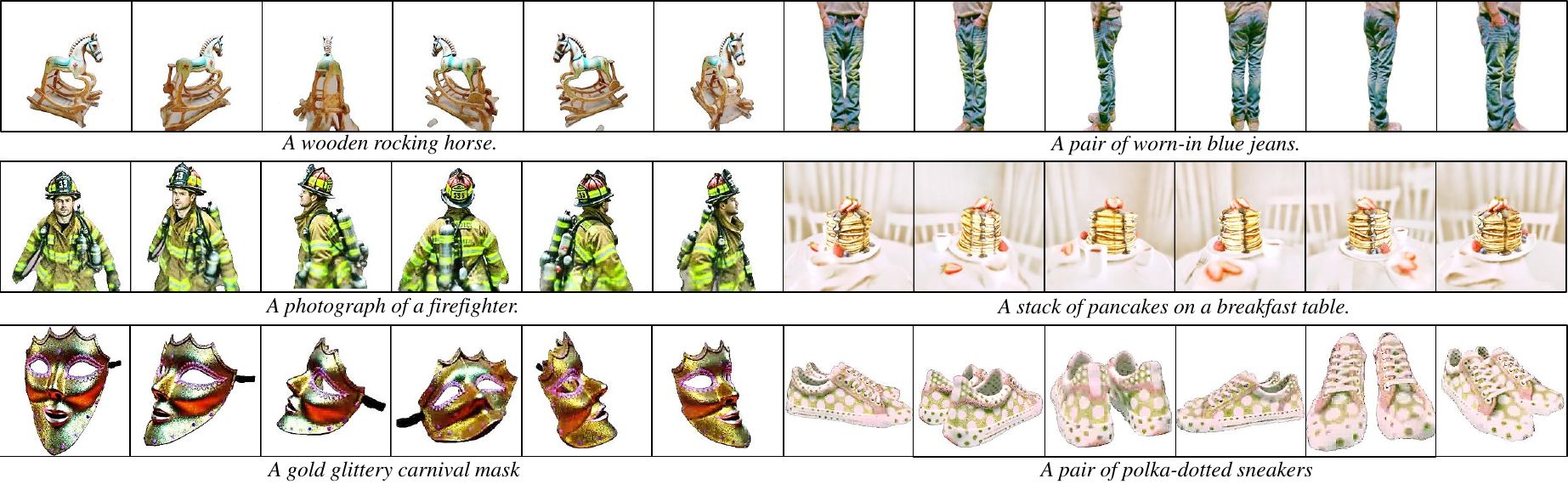}
    \caption{Multi-view analysis. Our method produces objects with consistent views.}
    \vspace{-12pt}
    \label{fig:mv-ana}
\end{figure*}

\noindent\textbf{Janus problem.}
Table \ref{tab:t3bench_results} shows that our method exhibits fewer Janus problems compared to other methods (Fig.~\ref{fig:mv-ana}). However, for objects which has both concave and convex geometry, our method still suffers from the Janus problem, which can be alleviated by using multi-view guidance~\cite{mvdream}. 

\hd{Diversity comparison.} To evaluate diversity in our optimization, we initialize the representation $\theta$ with fixed values and perform score distillation in multiple runs, each using a different random seed. We follow DiverseDream~\cite{diversedream} to compute the Inception Variance (IV) and Cosine Similarity metric. The formula of IV is $\mathrm{IV}(\theta) = \mathcal{H}[\mathbb{E}_{i, c}[p(y | g(\theta_i, c)]]$, where $p(y | x_i = g(\theta_i, c))$ is the pretrained classifier given the rendered images $x_i$ from particles $i$. When the outputs of a pre-trained classification model are uniform, it means that the diversity is high, along with a higher IV. We leverage InceptionV3~\cite{iv3} as a pre-trained classifier and DINO~\cite{dino} as a feature extractor. InceptionV3 and DINO are used to obtain the likelihood and feature vector to compute the cosine similarity matrix. We utilize 120 views of an object in the evaluation process. 

We compare the diversity of our method with DreamFusion~\cite{sds}, ProlificDreamer~\cite{vsd}, and DiverseDream~\cite{diversedream}. The result is shown in Table~\ref{tab:div} using 70 prompts inherited from DiverseDream~\cite{diversedream}. While DiverseDream produces a wide range of 3D models given the same prompt, their object quality varies, including some over-saturated objects. Unlike DiverseDream, our method generates diverse and high-quality 3D objects solely from text prompts, not requiring any reference images or textual inversion. 

\begin{table}[ht]
\small
\centering
% \vspace{-10pt}
\caption{Diversity comparisons between DiverseDream and ours.}
\vspace{-6pt}
\label{tab:div}
\setlength{\tabcolsep}{1pt}
\renewcommand{\arraystretch}{0.8}
\begin{tabular}{@{}lcccc@{}}
\toprule
Method                 & SDS\cite{sds}   & VSD\cite{vsd}   & DiverseDream\cite{diversedream}   & Ours           \\ \midrule
IV$\uparrow$           & 4.577 & 4.586 & 5.075          & \textbf{5.826} \\
Cosine Sim$\downarrow$ & 0.720 & 0.476 & \textbf{0.380} & 0.401          \\ \bottomrule
\end{tabular}
\vspace{-12pt}
\end{table}

\hd{Inversion analysis.} To prove the independence of our proposed method on the inversion technique, we compare the results of our method using DDIM~\cite{ddim} and Renoise~\cite{renoise} inversion (Tab.~\ref{tab:inversion}). As can be seen, our proposed method is not influenced much by the choice of the inversion method, either DDIM or Renoise, where the performance achieved by both inversion techniques are very close. 

\hd{Ablation test with Minority Sampling}
To better understand the effect of minority sampling, we perform an ablation test to compare SDS~\cite{sds}, our proposed method, with and without minority sampling. According to Fig.~\ref{fig:abs}, minority sampling plays a crucial role in our proposed method, which stabilizes the updating gradient (as discussed in Sec. \ref{sec:ana}) and thereby enhances the quality of the output 3D object.

\begin{table}[ht]
\small
\centering
\caption{Quantitative comparison between DDIM and Renoise inversion conducted on T3Bench~\cite{t3bench} benchmark.}
\vspace{-6pt}
\label{tab:inversion}
\renewcommand{\arraystretch}{0.8}
\begin{tabular}{@{}lllll@{}}
\toprule
                    &         & \multicolumn{1}{c}{\textbf{Qual.}} & \multicolumn{1}{c}{\textbf{Align}} & \multicolumn{1}{c}{\textbf{Avg}} \\ \midrule
\multirow{2}{*}{Single Object} & \textit{DDIM}    & \multicolumn{1}{l}{\textbf{58.7}}      & \multicolumn{1}{l}{\textbf{53.6}}      & \multicolumn{1}{l}{56.1}    \\
                    & \textit{Renoise} & \multicolumn{1}{l}{\textbf{58.7}}      & \multicolumn{1}{l}{53.5}      & \multicolumn{1}{l}{\textbf{56.2}}    \\ \midrule
\multirow{2}{*}{Single Object w. Surr} & \textit{DDIM}    & 47.4 & 57.6 & 52.5 \\
                     & \textit{Renoise} & \textbf{47.8} & \textbf{58.2} & \textbf{53.0} \\ \midrule
\multirow{2}{*}{Multiple Objects}  & \textit{DDIM}    & 51.3 & \textbf{40.2} & 45.7 \\
                     & \textit{Renoise} & \textbf{51.6} & \textbf{40.2} & \textbf{45.9} \\ \bottomrule
\end{tabular}
\vspace{-15pt}
\end{table}
\section{Limitation and Conclusion}
\hd{Limitation.} Our method is not without limitations. First, although our approach can generate high-fidelity objects, it still faces common problems in text-to-3D generation, such as Janus~\cite{perneg}, hollow face illusion~\cite{hollow-face-illusion}, and diffusion anomalies~\cite{sdi}, which might be solved by further regularization. Second, the diversity in our pipeline is fully automatic. Extending our method to include more controls would benefit downstream applications like 3D manipulation and editing.

\hd{Conclusion.} In this research, we propose to approximate the Jensen-Shannon Divergence (JSD) to improve the convergence and diversity of text-to-3D generation. Our method reformulates JSD using the theory of GAN training, leading to a minority sampling technique that effectively approximates JSD. 
For future work, we believe that there are more objective functions to explore in addition to JSD for 3D generation, including variants of the Wasserstein distances~\cite{wgan, wgan-gp, imp-wgan-gp}. Extending our method to generate dynamic 3D objects would be an interesting future work as well.
{
    \small
    \bibliographystyle{ieeenat_fullname}
    \bibliography{main}

\begin{thebibliography}{102}
\providecommand{\natexlab}[1]{#1}
\providecommand{\url}[1]{\texttt{#1}}
\expandafter\ifx\csname urlstyle\endcsname\relax
  \providecommand{\doi}[1]{doi: #1}\else
  \providecommand{\doi}{doi: \begingroup \urlstyle{rm}\Url}\fi

\bibitem[Akbari et~al.(2021)Akbari, Awais, Bashar, and Kittler]{stable-loss}
Ali Akbari, Muhammad Awais, Manijeh Bashar, and Josef Kittler.
\newblock How does loss function affect generalization performance of deep learning? application to human age estimation.
\newblock In \emph{Proceedings of the 38th International Conference on Machine Learning}, pages 141--151. PMLR, 2021.

\bibitem[Arjovsky and Bottou(2017{\natexlab{a}})]{gan-mode-collapse-theory}
Martin Arjovsky and Leon Bottou.
\newblock Towards principled methods for training generative adversarial networks.
\newblock In \emph{International Conference on Learning Representations}, 2017{\natexlab{a}}.

\bibitem[Arjovsky and Bottou(2017{\natexlab{b}})]{gan-modecollapse-theory}
Martin Arjovsky and Leon Bottou.
\newblock Towards principled methods for training generative adversarial networks.
\newblock In \emph{International Conference on Learning Representations}, 2017{\natexlab{b}}.

\bibitem[Arjovsky et~al.(2017)Arjovsky, Chintala, and Bottou]{wgan}
Martin Arjovsky, Soumith Chintala, and L{\'e}on Bottou.
\newblock {W}asserstein generative adversarial networks.
\newblock In \emph{Proceedings of the 34th International Conference on Machine Learning}, pages 214--223. PMLR, 2017.

\bibitem[Armandpour et~al.(2021)Armandpour, Sadeghian, Li, and Zhou]{pgan}
Mohammadreza Armandpour, Ali Sadeghian, Chunyuan Li, and Mingyuan Zhou.
\newblock Partition-guided gans.
\newblock In \emph{Proceedings of the IEEE/CVF Conference on Computer Vision and Pattern Recognition (CVPR)}, pages 5099--5109, 2021.

\bibitem[Armandpour et~al.(2024)Armandpour, Sadeghian, Zheng, Sadeghian, and Zhou]{perneg}
Mohammadreza Armandpour, Ali Sadeghian, Huangjie Zheng, Amir Sadeghian, and Mingyuan Zhou.
\newblock Re-imagine the negative prompt algorithm for 2d/3d diffusion, 2024.

\bibitem[Bang and Shim(2021)]{mg-gan}
Duhyeon Bang and Hyunjung Shim.
\newblock Mggan: Solving mode collapse using manifold-guided training.
\newblock In \emph{Proceedings of the IEEE/CVF International Conference on Computer Vision (ICCV) Workshops}, pages 2347--2356, 2021.

\bibitem[Bishop(2006)]{bishop}
Christopher~M. Bishop.
\newblock \emph{Pattern Recognition and Machine Learning (Information Science and Statistics)}.
\newblock Springer-Verlag, Berlin, Heidelberg, 2006.

\bibitem[Caron et~al.(2021)Caron, Touvron, Misra, J\'egou, Mairal, Bojanowski, and Joulin]{dino}
Mathilde Caron, Hugo Touvron, Ishan Misra, Herv\'e J\'egou, Julien Mairal, Piotr Bojanowski, and Armand Joulin.
\newblock Emerging properties in self-supervised vision transformers.
\newblock In \emph{Proceedings of the International Conference on Computer Vision (ICCV)}, 2021.

\bibitem[Chen et~al.(2024{\natexlab{a}})Chen, Yang, Yang, Feng, Fu, Foo, Lin, and Liu]{sculpt3d}
Cheng Chen, Xiaofeng Yang, Fan Yang, Chengzeng Feng, Zhoujie Fu, Chuan-Sheng Foo, Guosheng Lin, and Fayao Liu.
\newblock Sculpt3d: Multi-view consistent text-to-3d generation with sparse 3d prior.
\newblock In \emph{Proceedings of the IEEE/CVF Conference on Computer Vision and Pattern Recognition (CVPR)}, pages 10228--10237, 2024{\natexlab{a}}.

\bibitem[Chen et~al.(2024{\natexlab{b}})Chen, Dong, Shao, Hao, Yang, Su, and Zhu]{diff-certifiable-zero-shot}
Huanran Chen, Yinpeng Dong, Shitong Shao, Zhongkai Hao, Xiao Yang, Hang Su, and Jun Zhu.
\newblock Diffusion models are certifiably robust classifiers.
\newblock In \emph{The Thirty-eighth Annual Conference on Neural Information Processing Systems}, 2024{\natexlab{b}}.

\bibitem[Chen et~al.(2023)Chen, Chen, Jiao, and Jia]{fantasia3d}
Rui Chen, Yongwei Chen, Ningxin Jiao, and Kui Jia.
\newblock Fantasia3d: Disentangling geometry and appearance for high-quality text-to-3d content creation.
\newblock In \emph{Proceedings of the IEEE/CVF International Conference on Computer Vision (ICCV)}, 2023.

\bibitem[Chen et~al.(2024{\natexlab{c}})Chen, Pant, Yang, Yao, and Meit]{vp3d}
Yang Chen, Yingwei Pant, Haibo Yang, Ting Yao, and Tao Meit.
\newblock Vp3d: Unleashing 2d visual prompt for text-to-3d generation.
\newblock In \emph{2024 IEEE/CVF Conference on Computer Vision and Pattern Recognition (CVPR)}, pages 4896--4905, 2024{\natexlab{c}}.

\bibitem[Choi and Han(2022)]{mclgan}
Jinyoung Choi and Bohyung Han.
\newblock Mcl-gan: Generative adversarial networks with multiple specialized discriminators.
\newblock In \emph{Advances in Neural Information Processing Systems}, pages 29597--29609. Curran Associates, Inc., 2022.

\bibitem[Clark and Jaini(2023)]{diff-zero-shot}
Kevin Clark and Priyank Jaini.
\newblock Text-to-image diffusion models are zero shot classifiers.
\newblock In \emph{Thirty-seventh Conference on Neural Information Processing Systems}, 2023.

\bibitem[Deasy et~al.(2020)Deasy, Simidjievski, and Li\'{o}]{jsdg-star}
Jacob Deasy, Nikola Simidjievski, and Pietro Li\'{o}.
\newblock Constraining variational inference with geometric jensen-shannon divergence.
\newblock In \emph{Advances in Neural Information Processing Systems}, pages 10647--10658. Curran Associates, Inc., 2020.

\bibitem[Dendorfer et~al.(2021)Dendorfer, Elflein, and Leal-Taixé]{mggan}
Patrick Dendorfer, Sven Elflein, and Laura Leal-Taixé.
\newblock Mg-gan: A multi-generator model preventing out-of-distribution samples in pedestrian trajectory prediction.
\newblock In \emph{2021 IEEE/CVF International Conference on Computer Vision (ICCV)}, pages 13138--13147, 2021.

\bibitem[Endres and Schindelin(2003)]{jsd-triangle}
D.M. Endres and J.E. Schindelin.
\newblock A new metric for probability distributions.
\newblock \emph{IEEE Transactions on Information Theory}, 49\penalty0 (7):\penalty0 1858--1860, 2003.

\bibitem[Englesson and Azizpour(2021)]{jsdg-noisy}
Erik Englesson and Hossein Azizpour.
\newblock Generalized jensen-shannon divergence loss for learning with noisy labels.
\newblock In \emph{Advances in Neural Information Processing Systems}, pages 30284--30297. Curran Associates, Inc., 2021.

\bibitem[Garibi et~al.(2025)Garibi, Patashnik, Voynov, Averbuch-Elor, and Cohen-Or]{renoise}
Daniel Garibi, Or Patashnik, Andrey Voynov, Hadar Averbuch-Elor, and Daniel Cohen-Or.
\newblock Renoise: Real image inversion through iterative noising.
\newblock In \emph{Computer Vision -- ECCV 2024}, 2025.

\bibitem[Ge et~al.(2025)Ge, Xu, Ji, Peng, Tomizuka, Luo, Ding, Jampani, and Zhan]{compgs}
Chongjian Ge, Chenfeng Xu, Yuanfeng Ji, Chensheng Peng, Masayoshi Tomizuka, Ping Luo, Mingyu Ding, Varun Jampani, and Wei Zhan.
\newblock Compgs: Unleashing 2d compositionality for compositional text-to-3d via dynamically optimizing 3d gaussians.
\newblock In \emph{Proceedings of the IEEE/CVF Conference on Computer Vision and Pattern Recognition (CVPR)}, pages 18509--18520, 2025.

\bibitem[Ghosh et~al.(2018)Ghosh, Kulharia, Namboodiri, Torr, and Dokania]{madgan}
Arnab Ghosh, Viveka Kulharia, Vinay Namboodiri, Philip~H.S. Torr, and Puneet~K. Dokania.
\newblock Multi-agent diverse generative adversarial networks.
\newblock In \emph{2018 IEEE/CVF Conference on Computer Vision and Pattern Recognition}, pages 8513--8521, 2018.

\bibitem[Goodfellow et~al.(2014)Goodfellow, Pouget-Abadie, Mirza, Xu, Warde-Farley, Ozair, Courville, and Bengio]{gan}
Ian Goodfellow, Jean Pouget-Abadie, Mehdi Mirza, Bing Xu, David Warde-Farley, Sherjil Ozair, Aaron Courville, and Yoshua Bengio.
\newblock Generative adversarial nets.
\newblock In \emph{Advances in Neural Information Processing Systems}. Curran Associates, Inc., 2014.

\bibitem[Gulrajani et~al.(2017)Gulrajani, Ahmed, Arjovsky, Dumoulin, and Courville]{wgan-gp}
Ishaan Gulrajani, Faruk Ahmed, Martin Arjovsky, Vincent Dumoulin, and Aaron~C Courville.
\newblock Improved training of wasserstein gans.
\newblock In \emph{Advances in Neural Information Processing Systems}. Curran Associates, Inc., 2017.

\bibitem[Guo et~al.(2023)Guo, Liu, Shao, Laforte, Voleti, Luo, Chen, Zou, Wang, Cao, and Zhang]{threestudio2023}
Yuan-Chen Guo, Ying-Tian Liu, Ruizhi Shao, Christian Laforte, Vikram Voleti, Guan Luo, Chia-Hao Chen, Zi-Xin Zou, Chen Wang, Yan-Pei Cao, and Song-Hai Zhang.
\newblock threestudio: A unified framework for 3d content generation.
\newblock \url{https://github.com/threestudio-project/threestudio}, 2023.

\bibitem[He et~al.(2023)He, Bai, Lin, Zhao, Hu, Sheng, Yi, Li, and Liu]{t3bench}
Yuze He, Yushi Bai, Matthieu Lin, Wang Zhao, Yubin Hu, Jenny Sheng, Ran Yi, Juanzi Li, and Yong-Jin Liu.
\newblock T$^3$bench: Benchmarking current progress in text-to-3d generation, 2023.

\bibitem[Hill and Johnston(2007)]{hollow-face-illusion}
Harold Hill and Alan Johnston.
\newblock The hollow-face illusion: Object-specific knowledge, general assumptions or properties of the stimulus?
\newblock \emph{Perception}, 36\penalty0 (2):\penalty0 199--223, 2007.
\newblock PMID: 17402664.

\bibitem[Ho and Salimans(2021)]{ho2021classifierfree}
Jonathan Ho and Tim Salimans.
\newblock Classifier-free diffusion guidance.
\newblock In \emph{NeurIPS 2021 Workshop on Deep Generative Models and Downstream Applications}, 2021.

\bibitem[Ho et~al.(2020)Ho, Jain, and Abbeel]{ddpm}
Jonathan Ho, Ajay Jain, and Pieter Abbeel.
\newblock Denoising diffusion probabilistic models.
\newblock In \emph{Advances in Neural Information Processing Systems}, pages 6840--6851. Curran Associates, Inc., 2020.

\bibitem[Hoang et~al.(2018)Hoang, Nguyen, Le, and Phung]{mgan}
Quan Hoang, Tu~Dinh Nguyen, Trung Le, and Dinh Phung.
\newblock {MGAN}: Training generative adversarial nets with multiple generators.
\newblock In \emph{International Conference on Learning Representations}, 2018.

\bibitem[Hu et~al.(2022)Hu, yelong shen, Wallis, Allen-Zhu, Li, Wang, Wang, and Chen]{lora}
Edward~J Hu, yelong shen, Phillip Wallis, Zeyuan Allen-Zhu, Yuanzhi Li, Shean Wang, Lu Wang, and Weizhu Chen.
\newblock Lo{RA}: Low-rank adaptation of large language models.
\newblock In \emph{International Conference on Learning Representations}, 2022.

\bibitem[Jain et~al.(2022)Jain, Mildenhall, Barron, Abbeel, and Poole]{dreamfield}
Ajay Jain, Ben Mildenhall, Jonathan~T. Barron, Pieter Abbeel, and Ben Poole.
\newblock Zero-shot text-guided object generation with dream fields.
\newblock In \emph{2022 IEEE/CVF Conference on Computer Vision and Pattern Recognition (CVPR)}, pages 857--866, 2022.

\bibitem[Jiang et~al.(2025)Jiang, Zeng, Hu, Xu, Zhang, Xu, and Yeung]{jointdreamer}
Chenhan Jiang, Yihan Zeng, Tianyang Hu, Songcun Xu, Wei Zhang, Hang Xu, and Dit-Yan Yeung.
\newblock Jointdreamer: Ensuring geometry consistency and text congruence in text-to-3d generation via joint score distillation.
\newblock In \emph{Computer Vision -- ECCV 2024}, pages 439--456, Cham, 2025.

\bibitem[Jun and Nichol(2023)]{shap-e}
Heewoo Jun and Alex Nichol.
\newblock Shap-e: Generating conditional 3d implicit functions, 2023.

\bibitem[Kerbl et~al.(2023)Kerbl, Kopanas, Leimkuehler, and Drettakis]{gs}
Bernhard Kerbl, Georgios Kopanas, Thomas Leimkuehler, and George Drettakis.
\newblock 3d gaussian splatting for real-time radiance field rendering.
\newblock \emph{ACM Trans. Graph.}, 42\penalty0 (4), 2023.

\bibitem[Kim et~al.(2025)Kim, Park, and Ye]{dream-sampler}
Jeongsol Kim, Geon~Yeong Park, and Jong~Chul Ye.
\newblock Dreamsampler: Unifying diffusion sampling and score distillation for image manipulation.
\newblock In \emph{Computer Vision -- ECCV 2024}, pages 398--414, Cham, 2025. Springer Nature Switzerland.

\bibitem[Knapp(2005)]{basic-real-analysis}
Anthony~W Knapp.
\newblock \emph{Basic Real Analysis}.
\newblock Birkhauser Boston, Secaucus, NJ, 2005 edition, 2005.

\bibitem[Lee et~al.(2021)Lee, Lee, Park, and Lee]{bounded-smooth-loss}
Sungyoon Lee, Woojin Lee, Jinseong Park, and Jaewook Lee.
\newblock Towards better understanding of training certifiably robust models against adversarial examples.
\newblock In \emph{Advances in Neural Information Processing Systems}, pages 953--964. Curran Associates, Inc., 2021.

\bibitem[Li et~al.(2025{\natexlab{a}})Li, Zhang, Yang, and Wang]{coser}
Bonan Li, Zicheng Zhang, Xingyi Yang, and Xinchao Wang.
\newblock Coser: Towards consistent dense multiview text-to-image generator for 3d creation.
\newblock In \emph{Proceedings of the IEEE/CVF Conference on Computer Vision and Pattern Recognition (CVPR)}, pages 2880--2890, 2025{\natexlab{a}}.

\bibitem[Li et~al.(2025{\natexlab{b}})Li, Liu, Sznaier, and Camps]{half_gs}
Haolin Li, Jinyang Liu, Mario Sznaier, and Octavia Camps.
\newblock 3d-hgs: 3d half-gaussian splatting.
\newblock In \emph{Proceedings of the IEEE/CVF Conference on Computer Vision and Pattern Recognition (CVPR)}, pages 10996--11005, 2025{\natexlab{b}}.

\bibitem[Li et~al.(2023)Li, Li, Savarese, and Hoi]{blip2}
Junnan Li, Dongxu Li, Silvio Savarese, and Steven Hoi.
\newblock Blip-2: bootstrapping language-image pre-training with frozen image encoders and large language models.
\newblock In \emph{Proceedings of the 40th International Conference on Machine Learning}. JMLR.org, 2023.

\bibitem[Li et~al.(2024{\natexlab{a}})Li, Tan, Zhang, Xu, Luan, Xu, Hong, Sunkavalli, Shakhnarovich, and Bi]{instant3d}
Jiahao Li, Hao Tan, Kai Zhang, Zexiang Xu, Fujun Luan, Yinghao Xu, Yicong Hong, Kalyan Sunkavalli, Greg Shakhnarovich, and Sai Bi.
\newblock Instant3d: Fast text-to-3d with sparse-view generation and large reconstruction model.
\newblock In \emph{The Twelfth International Conference on Learning Representations}, 2024{\natexlab{a}}.

\bibitem[Li et~al.(2024{\natexlab{b}})Li, Tan, Zhang, Xu, Luan, Xu, Hong, Sunkavalli, Shakhnarovich, and Bi]{li2024instantd}
Jiahao Li, Hao Tan, Kai Zhang, Zexiang Xu, Fujun Luan, Yinghao Xu, Yicong Hong, Kalyan Sunkavalli, Greg Shakhnarovich, and Sai Bi.
\newblock Instant3d: Fast text-to-3d with sparse-view generation and large reconstruction model.
\newblock In \emph{The Twelfth International Conference on Learning Representations}, 2024{\natexlab{b}}.

\bibitem[Liang et~al.(2024)Liang, Yang, Lin, Li, Xu, and Chen]{luciddreamer}
Yixun Liang, Xin Yang, Jiantao Lin, Haodong Li, Xiaogang Xu, and Yingcong Chen.
\newblock { LucidDreamer: Towards High-Fidelity Text-to-3D Generation via Interval Score Matching }.
\newblock In \emph{2024 IEEE/CVF Conference on Computer Vision and Pattern Recognition (CVPR)}, pages 6517--6526, Los Alamitos, CA, USA, 2024. IEEE Computer Society.

\bibitem[Lin et~al.(2023)Lin, Gao, Tang, Takikawa, Zeng, Huang, Kreis, Fidler, Liu, and Lin]{magic3d}
Chen-Hsuan Lin, Jun Gao, Luming Tang, Towaki Takikawa, Xiaohui Zeng, Xun Huang, Karsten Kreis, Sanja Fidler, Ming-Yu Liu, and Tsung-Yi Lin.
\newblock Magic3d: High-resolution text-to-3d content creation.
\newblock In \emph{IEEE Conference on Computer Vision and Pattern Recognition ({CVPR})}, 2023.

\bibitem[Lin(1991)]{jsd}
J. Lin.
\newblock Divergence measures based on the shannon entropy.
\newblock \emph{IEEE Transactions on Information Theory}, 37\penalty0 (1):\penalty0 145--151, 1991.

\bibitem[Liu et~al.(2023{\natexlab{a}})Liu, Li, Wu, Liang, Huang, Li, Ghanem, and Zheng]{omeegan}
Haozhe Liu, Bing Li, Haoqian Wu, Hanbang Liang, Yawen Huang, Yuexiang Li, Bernard Ghanem, and Yefeng Zheng.
\newblock Combating mode collapse via offline manifold entropy estimation.
\newblock \emph{Proceedings of the AAAI Conference on Artificial Intelligence}, 37\penalty0 (7):\penalty0 8834--8842, 2023{\natexlab{a}}.

\bibitem[Liu et~al.(2023{\natexlab{b}})Liu, Xu, Jin, Chen, T, Xu, and Su]{one2345}
Minghua Liu, Chao Xu, Haian Jin, Linghao Chen, Mukund~Varma T, Zexiang Xu, and Hao Su.
\newblock One-2-3-45: Any single image to 3d mesh in 45 seconds without per-shape optimization.
\newblock In \emph{Thirty-seventh Conference on Neural Information Processing Systems}, 2023{\natexlab{b}}.

\bibitem[Lorraine et~al.(2023)Lorraine, Xie, Zeng, Lin, Takikawa, Sharp, Lin, Liu, Fidler, and Lucas]{att3d}
Jonathan Lorraine, Kevin Xie, Xiaohui Zeng, Chen-Hsuan Lin, Towaki Takikawa, Nicholas Sharp, Tsung-Yi Lin, Ming-Yu Liu, Sanja Fidler, and James Lucas.
\newblock Att3d: Amortized text-to-3d object synthesis.
\newblock \emph{The International Conference on Computer Vision (ICCV)}, 2023.

\bibitem[Lukoianov et~al.(2024)Lukoianov, de~Oc{\'a}riz~Borde, Greenewald, Guizilini, Bagautdinov, Sitzmann, and Solomon]{sdi}
Artem Lukoianov, Haitz~S{\'a}ez de Oc{\'a}riz~Borde, Kristjan Greenewald, Vitor~Campagnolo Guizilini, Timur Bagautdinov, Vincent Sitzmann, and Justin Solomon.
\newblock Score distillation via reparametrized {DDIM}.
\newblock In \emph{The Thirty-eighth Annual Conference on Neural Information Processing Systems}, 2024.

\bibitem[Ma et~al.(2025)Ma, Wei, Zhang, Zhu, Lei, and Zhang]{scaledreamer}
Zhiyuan Ma, Yuxiang Wei, Yabin Zhang, Xiangyu Zhu, Zhen Lei, and Lei Zhang.
\newblock Scaledreamer: Scalable text-to-3d synthesis with asynchronous score distillation.
\newblock In \emph{Computer Vision -- ECCV 2024}, pages 1--19, Cham, 2025.

\bibitem[Mescheder et~al.(2018)Mescheder, Geiger, and Nowozin]{gan-converge}
Lars Mescheder, Andreas Geiger, and Sebastian Nowozin.
\newblock Which training methods for {GAN}s do actually converge?
\newblock In \emph{Proceedings of the 35th International Conference on Machine Learning}, pages 3481--3490. PMLR, 2018.

\bibitem[Metzer et~al.(2023)Metzer, Richardson, Patashnik, Giryes, and Cohen-Or]{latentnerf}
Gal Metzer, Elad Richardson, Or Patashnik, Raja Giryes, and Daniel Cohen-Or.
\newblock Latent-nerf for shape-guided generation of 3d shapes and textures.
\newblock In \emph{Proceedings of the IEEE/CVF Conference on Computer Vision and Pattern Recognition (CVPR)}, pages 12663--12673, 2023.

\bibitem[Mildenhall et~al.(2020)Mildenhall, Srinivasan, Tancik, Barron, Ramamoorthi, and Ng]{nerf}
Ben Mildenhall, Pratul~P. Srinivasan, Matthew Tancik, Jonathan~T. Barron, Ravi Ramamoorthi, and Ren Ng.
\newblock Nerf: Representing scenes as neural radiance fields for view synthesis.
\newblock In \emph{Computer Vision -- ECCV 2020}, pages 405--421, Cham, 2020.

\bibitem[Mokady et~al.(2023)Mokady, Hertz, Aberman, Pritch, and Cohen-Or]{null-text-inversion}
Ron Mokady, Amir Hertz, Kfir Aberman, Yael Pritch, and Daniel Cohen-Or.
\newblock Null-text inversion for editing real images using guided diffusion models.
\newblock In \emph{2023 IEEE/CVF Conference on Computer Vision and Pattern Recognition (CVPR)}, pages 6038--6047, 2023.

\bibitem[Ni et~al.(2022)Ni, Koniusz, Hartley, and Nock]{mlbgan}
Yao Ni, Piotr Koniusz, Richard Hartley, and Richard Nock.
\newblock Manifold learning benefits gans.
\newblock In \emph{Proceedings of the IEEE/CVF Conference on Computer Vision and Pattern Recognition (CVPR)}, pages 11265--11274, 2022.

\bibitem[Nichol et~al.(2022)Nichol, Jun, Dhariwal, Mishkin, and Chen]{point-e}
Alex Nichol, Heewoo Jun, Prafulla Dhariwal, Pamela Mishkin, and Mark Chen.
\newblock Point-e: A system for generating 3d point clouds from complex prompts, 2022.

\bibitem[Nielsen(2013)]{jd}
Frank Nielsen.
\newblock Jeffreys centroids: A closed-form expression for positive histograms and a guaranteed tight approximation for frequency histograms.
\newblock \emph{IEEE Signal Processing Letters}, 20\penalty0 (7):\penalty0 657--660, 2013.

\bibitem[Nielsen(2019)]{jsds}
Frank Nielsen.
\newblock On the jensen–shannon symmetrization of distances relying on abstract means.
\newblock \emph{Entropy}, 21\penalty0 (5):\penalty0 485, 2019.

\bibitem[Nielsen and Garcia(2011)]{jsdg}
Frank Nielsen and Vincent Garcia.
\newblock Statistical exponential families: A digest with flash cards, 2011.

\bibitem[Nowozin et~al.(2016)Nowozin, Cseke, and Tomioka]{fgan}
Sebastian Nowozin, Botond Cseke, and Ryota Tomioka.
\newblock f-gan: Training generative neural samplers using variational divergence minimization.
\newblock In \emph{Advances in Neural Information Processing Systems}. Curran Associates, Inc., 2016.

\bibitem[Park et~al.(2018)Park, Yoo, Bahng, Choo, and Park]{megan}
David~Keetae Park, Seungjoo Yoo, Hyojin Bahng, Jaegul Choo, and Noseong Park.
\newblock Megan: Mixture of experts of generative adversarial networks for multimodal image generation.
\newblock In \emph{Proceedings of the Twenty-Seventh International Joint Conference on Artificial Intelligence, {IJCAI-18}}, pages 878--884. International Joint Conferences on Artificial Intelligence Organization, 2018.

\bibitem[Poole et~al.(2023)Poole, Jain, Barron, and Mildenhall]{sds}
Ben Poole, Ajay Jain, Jonathan~T. Barron, and Ben Mildenhall.
\newblock Dreamfusion: Text-to-3d using 2d diffusion.
\newblock In \emph{The Eleventh International Conference on Learning Representations}, 2023.

\bibitem[Qiu et~al.(2024)Qiu, Chen, Gu, Zuo, Xu, Wu, Yuan, Dong, Bo, and Han]{richdreamer}
Lingteng Qiu, Guanying Chen, Xiaodong Gu, Qi Zuo, Mutian Xu, Yushuang Wu, Weihao Yuan, Zilong Dong, Liefeng Bo, and Xiaoguang Han.
\newblock Richdreamer: A generalizable normal-depth diffusion model for detail richness in text-to-3d.
\newblock In \emph{Proceedings of the IEEE/CVF Conference on Computer Vision and Pattern Recognition}, pages 9914--9925, 2024.

\bibitem[Rombach et~al.(2022)Rombach, Blattmann, Lorenz, Esser, and Ommer]{sd}
Robin Rombach, Andreas Blattmann, Dominik Lorenz, Patrick Esser, and Björn Ommer.
\newblock High-resolution image synthesis with latent diffusion models, 2022.

\bibitem[Shi et~al.(2024)Shi, Wang, Ye, Mai, Li, and Yang]{mvdream}
Yichun Shi, Peng Wang, Jianglong Ye, Long Mai, Kejie Li, and Xiao Yang.
\newblock {MVD}ream: Multi-view diffusion for 3d generation.
\newblock In \emph{The Twelfth International Conference on Learning Representations}, 2024.

\bibitem[Sohl-Dickstein et~al.(2015)Sohl-Dickstein, Weiss, Maheswaranathan, and Ganguli]{diff}
Jascha Sohl-Dickstein, Eric~A. Weiss, Niru Maheswaranathan, and Surya Ganguli.
\newblock Deep unsupervised learning using nonequilibrium thermodynamics.
\newblock In \emph{Proceedings of the 32nd International Conference on International Conference on Machine Learning - Volume 37}, page 2256–2265. JMLR.org, 2015.

\bibitem[Song et~al.(2021)Song, Meng, and Ermon]{ddim}
Jiaming Song, Chenlin Meng, and Stefano Ermon.
\newblock Denoising diffusion implicit models.
\newblock In \emph{International Conference on Learning Representations}, 2021.

\bibitem[Song and Ermon(2019)]{sbgm}
Yang Song and Stefano Ermon.
\newblock Generative modeling by estimating gradients of the data distribution.
\newblock In \emph{Advances in Neural Information Processing Systems}. Curran Associates, Inc., 2019.

\bibitem[Szegedy et~al.(2016)Szegedy, Vanhoucke, Ioffe, Shlens, and Wojna]{iv3}
Christian Szegedy, Vincent Vanhoucke, Sergey Ioffe, Jon Shlens, and Zbigniew Wojna.
\newblock Rethinking the inception architecture for computer vision.
\newblock In \emph{2016 IEEE Conference on Computer Vision and Pattern Recognition (CVPR)}, pages 2818--2826, 2016.

\bibitem[Tang et~al.(2024{\natexlab{a}})Tang, Wang, Wu, and Zhang]{tang2024ssd}
Boshi Tang, Jianan Wang, Zhiyong Wu, and Lei Zhang.
\newblock Stable score distillation for high-quality 3d generation, 2024{\natexlab{a}}.

\bibitem[Tang et~al.(2024{\natexlab{b}})Tang, Ren, Zhou, Liu, and Zeng]{dreamgaussian}
Jiaxiang Tang, Jiawei Ren, Hang Zhou, Ziwei Liu, and Gang Zeng.
\newblock Dreamgaussian: Generative gaussian splatting for efficient 3d content creation.
\newblock In \emph{The Twelfth International Conference on Learning Representations}, 2024{\natexlab{b}}.

\bibitem[Tang et~al.(2025)Tang, Chen, Chen, Wang, Zeng, and Liu]{lgm}
Jiaxiang Tang, Zhaoxi Chen, Xiaokang Chen, Tengfei Wang, Gang Zeng, and Ziwei Liu.
\newblock Lgm: Large multi-view gaussian model for high-resolution 3d content creation.
\newblock In \emph{Computer Vision -- ECCV 2024}, pages 1--18, Cham, 2025. Springer Nature Switzerland.

\bibitem[Thiagarajan and Ghosh(2025)]{jsdg-kl-unstable}
Ponkrshnan Thiagarajan and Susanta Ghosh.
\newblock Jensen–shannon divergence based novel loss functions for bayesian neural networks.
\newblock \emph{Neurocomputing}, 618:\penalty0 129115, 2025.

\bibitem[Thomas M.~Cover(2005)]{kld}
Joy A.~Thomas Thomas M.~Cover.
\newblock \emph{Entropy, Relative Entropy, and Mutual Information}, chapter~2, pages 13--55.
\newblock John Wiley and Sons, Ltd, 2005.

\bibitem[Tian et~al.(2024)Tian, Aggarwal, Colaco, Kira, and Gonzalez-Franco]{DAS}
Junjiao Tian, Lavisha Aggarwal, Andrea Colaco, Zsolt Kira, and Mar Gonzalez-Franco.
\newblock { Diffuse, Attend, and Segment: Unsupervised Zero-Shot Segmentation using Stable Diffusion }.
\newblock In \emph{2024 IEEE/CVF Conference on Computer Vision and Pattern Recognition (CVPR)}, pages 3554--3563. IEEE Computer Society, 2024.

\bibitem[Tran et~al.(2024)Tran, Luu, Nguyen, Nguyen, and Hua]{isd}
Uy~Dieu Tran, Minh Luu, Phong~Ha Nguyen, Khoi Nguyen, and Binh-Son Hua.
\newblock Modedreamer: Mode guiding score distillation for text-to-3d generation using reference image prompts, 2024.

\bibitem[Tran et~al.(2025)Tran, Luu, Nguyen, Nguyen, and Hua]{diversedream}
Uy~Dieu Tran, Minh Luu, Phong~Ha Nguyen, Khoi Nguyen, and Binh-Son Hua.
\newblock Diverse text-to-3d synthesis with augmented text embedding.
\newblock In \emph{European Conference on Computer Vision}, pages 217--235. Springer, 2025.

\bibitem[Um and Ye(2025)]{self-guide}
Soobin Um and Jong~Chul Ye.
\newblock Self-guided generation of minority samples using diffusion models.
\newblock In \emph{Computer Vision -- ECCV 2024}, pages 414--430, Cham, 2025. Springer Nature Switzerland.

\bibitem[Um et~al.(2024)Um, Lee, and Ye]{dont-play-fav}
Soobin Um, Suhyeon Lee, and Jong~Chul Ye.
\newblock Don't play favorites: Minority guidance for diffusion models.
\newblock In \emph{ICLR}, 2024.

\bibitem[Wan et~al.(2024)Wan, Paschalidou, Huang, Liu, Shen, Xiang, Liao, and Guibas]{cad}
Ziyu Wan, Despoina Paschalidou, Ian Huang, Hongyu Liu, Bokui Shen, Xiaoyu Xiang, Jing Liao, and Leonidas Guibas.
\newblock Cad : Photorealistic 3d generation via adversarial distillation.
\newblock In \emph{2024 IEEE/CVF Conference on Computer Vision and Pattern Recognition (CVPR)}, pages 10194--10207, 2024.

\bibitem[Wang et~al.(2018)Wang, Liu, and Liu]{fd-vi}
Dilin Wang, Hao Liu, and Qiang Liu.
\newblock Variational inference with tail-adaptive f-divergence.
\newblock In \emph{Advances in Neural Information Processing Systems}. Curran Associates, Inc., 2018.

\bibitem[Wang et~al.(2023{\natexlab{a}})Wang, Du, Li, Yeh, and Shakhnarovich]{sjc}
Haochen Wang, Xiaodan Du, Jiahao Li, Raymond~A. Yeh, and Greg Shakhnarovich.
\newblock Score jacobian chaining: Lifting pretrained 2d diffusion models for 3d generation.
\newblock In \emph{Proceedings of the IEEE/CVF Conference on Computer Vision and Pattern Recognition (CVPR)}, pages 12619--12629, 2023{\natexlab{a}}.

\bibitem[Wang et~al.(2023{\natexlab{b}})Wang, Fan, Xu, Wang, Mohan, Iandola, Ranjan, Li, Liu, Wang, et~al.]{wang2023steindreamer}
Peihao Wang, Zhiwen Fan, Dejia Xu, Dilin Wang, Sreyas Mohan, Forrest Iandola, Rakesh Ranjan, Yilei Li, Qiang Liu, Zhangyang Wang, et~al.
\newblock Steindreamer: Variance reduction for text-to-3d score distillation via stein identity.
\newblock \emph{arXiv preprint arXiv:2401.00604}, 2023{\natexlab{b}}.

\bibitem[Wang et~al.(2024)Wang, Xu, Fan, Wang, Mohan, Iandola, Ranjan, Li, Liu, Wang, and Chandra]{anonymous2023taming}
Peihao Wang, Dejia Xu, Zhiwen Fan, Dilin Wang, Sreyas Mohan, Forrest Iandola, Rakesh Ranjan, Yilei Li, Qiang Liu, Zhangyang Wang, and Vikas Chandra.
\newblock Taming mode collapse in score distillation for text-to-3d generation.
\newblock In \emph{Proceedings of the IEEE/CVF Conference on Computer Vision and Pattern Recognition (CVPR)}, pages 9037--9047, 2024.

\bibitem[Wang et~al.(2023{\natexlab{c}})Wang, Lu, Wang, Bao, LI, Su, and Zhu]{vsd}
Zhengyi Wang, Cheng Lu, Yikai Wang, Fan Bao, Chongxuan LI, Hang Su, and Jun Zhu.
\newblock Prolificdreamer: High-fidelity and diverse text-to-3d generation with variational score distillation.
\newblock In \emph{Advances in Neural Information Processing Systems}, pages 8406--8441. Curran Associates, Inc., 2023{\natexlab{c}}.

\bibitem[Wang et~al.(2023{\natexlab{d}})Wang, Zheng, He, Chen, and Zhou]{diff-gan}
Zhendong Wang, Huangjie Zheng, Pengcheng He, Weizhu Chen, and Mingyuan Zhou.
\newblock Diffusion-{GAN}: Training {GAN}s with diffusion.
\newblock In \emph{The Eleventh International Conference on Learning Representations}, 2023{\natexlab{d}}.

\bibitem[Wei et~al.(2024)Wei, Zhou, Sun, and Zhang]{asd}
Min Wei, Jingkai Zhou, Junyao Sun, and Xuesong Zhang.
\newblock Adversarial score distillation: When score distillation meets gan.
\newblock In \emph{Proceedings of the IEEE/CVF Conference on Computer Vision and Pattern Recognition (CVPR)}, pages 8131--8141, 2024.

\bibitem[Wei et~al.(2018)Wei, Liu, Wang, and Gong]{imp-wgan-gp}
Xiang Wei, Zixia Liu, Liqiang Wang, and Boqing Gong.
\newblock Improving the improved training of wasserstein {GAN}s.
\newblock In \emph{International Conference on Learning Representations}, 2018.

\bibitem[Wu et~al.(2019)Wu, Huang, Acharya, Li, Thoma, Paudel, and Gool]{swgan}
Jiqing Wu, Zhiwu Huang, Dinesh Acharya, Wen Li, Janine Thoma, Danda~Pani Paudel, and Luc~Van Gool.
\newblock Sliced wasserstein generative models.
\newblock In \emph{Proceedings of the IEEE/CVF Conference on Computer Vision and Pattern Recognition (CVPR)}, 2019.

\bibitem[Wu et~al.(2024)Wu, Yang, Li, Zhang, Liu, Guibas, Lin, and Wetzstein]{gpteval3d}
Tong Wu, Guandao Yang, Zhibing Li, Kai Zhang, Ziwei Liu, Leonidas Guibas, Dahua Lin, and Gordon Wetzstein.
\newblock Gpt-4v(ision) is a human-aligned evaluator for text-to-3d generation.
\newblock In \emph{Proceedings of the IEEE/CVF Conference on Computer Vision and Pattern Recognition (CVPR)}, pages 22227--22238, 2024.

\bibitem[Xia et~al.(2024)Xia, Shen, Yang, Yi, Wang, and Liu]{smart}
Mengfei Xia, Yujun Shen, Ceyuan Yang, Ran Yi, Wenping Wang, and Yong-Jin Liu.
\newblock {SM}a{R}t: Improving {GAN}s with score matching regularity.
\newblock In \emph{Proceedings of the 41st International Conference on Machine Learning}, pages 54133--54155. PMLR, 2024.

\bibitem[Xiang et~al.(2024)Xiang, Lv, Xu, Deng, Wang, Zhang, Chen, Tong, and Yang]{trellis}
Jianfeng Xiang, Zelong Lv, Sicheng Xu, Yu Deng, Ruicheng Wang, Bowen Zhang, Dong Chen, Xin Tong, and Jiaolong Yang.
\newblock Structured 3d latents for scalable and versatile 3d generation.
\newblock \emph{arXiv preprint arXiv:2412.01506}, 2024.

\bibitem[Xie et~al.(2024)Xie, Lorraine, Cao, Gao, Lucas, Torralba, Fidler, and Zeng]{latte3d}
Kevin Xie, Jonathan Lorraine, Tianshi Cao, Jun Gao, James Lucas, Antonio Torralba, Sanja Fidler, and Xiaohui Zeng.
\newblock Latte3d: Large-scale amortized text-to-enhanced3d synthesis.
\newblock \emph{The 18th European Conference on Computer Vision (ECCV)}, 2024.

\bibitem[Xu et~al.(2023)Xu, Liu, Wu, Tong, Li, Ding, Tang, and Dong]{imagereward}
Jiazheng Xu, Xiao Liu, Yuchen Wu, Yuxuan Tong, Qinkai Li, Ming Ding, Jie Tang, and Yuxiao Dong.
\newblock Imagereward: learning and evaluating human preferences for text-to-image generation.
\newblock In \emph{Proceedings of the 37th International Conference on Neural Information Processing Systems}, pages 15903--15935, 2023.

\bibitem[Yan et~al.(2025)Yan, Chen, and Wang]{cfd}
Runjie Yan, Yinbo Chen, and Xiaolong Wang.
\newblock Consistent flow distillation for text-to-3d generation.
\newblock In \emph{The Thirteenth International Conference on Learning Representations}, 2025.

\bibitem[Yang et~al.(2025)Yang, Chen, Pan, Yao, Chen, Wu, Jiang, and Mei]{dreammesh}
Haibo Yang, Yang Chen, Yingwei Pan, Ting Yao, Zhineng Chen, Zuxuan Wu, Yu-Gang Jiang, and Tao Mei.
\newblock Dreammesh: Jointly manipulating and texturing triangle meshes for text-to-3d generation.
\newblock In \emph{Computer Vision -- ECCV 2024}, pages 162--178, Cham, 2025. Springer Nature Switzerland.

\bibitem[Ye et~al.(2024)Ye, Liu, Li, Wang, Wang, Wang, Duan, and Zhu]{dreamreward}
Junliang Ye, Fangfu Liu, Qixiu Li, Zhengyi Wang, Yikai Wang, Xinzhou Wang, Yueqi Duan, and Jun Zhu.
\newblock Dreamreward: Text-to-3d generation with human preference.
\newblock In \emph{European Conference on Computer Vision}, pages 259--276. Springer, 2024.

\bibitem[Yong et~al.(2024)Yong, Xie, Stepputtis, and Sycara]{glnerf}
Silong Yong, Yaqi Xie, Simon Stepputtis, and Katia Sycara.
\newblock Gl-nerf: Gauss-laguerre quadrature enables training-free nerf acceleration.
\newblock In \emph{Advances in Neural Information Processing Systems}, pages 120418--120442. Curran Associates, Inc., 2024.

\bibitem[Zhang et~al.(2023)Zhang, Rao, and Agrawala]{controlnet}
Lvmin Zhang, Anyi Rao, and Maneesh Agrawala.
\newblock Adding conditional control to text-to-image diffusion models.
\newblock In \emph{2023 IEEE/CVF International Conference on Computer Vision (ICCV)}, pages 3813--3824, 2023.

\bibitem[Zheng et~al.(2025)Zheng, Lin, Liu, Xu, Nie, and He]{recdreamer}
Chenxi Zheng, Yihong Lin, Bangzhen Liu, Xuemiao Xu, Yongwei Nie, and Shengfeng He.
\newblock Recdreamer: Consistent text-to-3d generation via uniform score distillation.
\newblock In \emph{The Thirteenth International Conference on Learning Representations}, 2025.

\bibitem[Zhu et~al.(2024)Zhu, Zhuang, and Koyejo]{hifa}
Junzhe Zhu, Peiye Zhuang, and Sanmi Koyejo.
\newblock {HIFA}: High-fidelity text-to-3d generation with advanced diffusion guidance.
\newblock In \emph{The Twelfth International Conference on Learning Representations}, 2024.

\end{thebibliography}
}
\begin{abstract}
    In this supplementary document, we provide an additional discussion to explain how our method compares to Adversarial Score Distillation (ASD)~\cite{asd} (Section~\ref{apx:proof-asd-jd}). 
    We then provide detailed derivations of the JSD objective and its approximation, taking inspiration from GAN theory (Section~\ref{apx:proof-jsd-logodd} and Section~\ref{apx:loss-derivation}). We provide quantitative results of the GPTEval3D~\cite{gpteval3d} benchmark in Section~\ref{apx:gpteval3d}.
    Finally, we provide additional empirical analysis on the toy dataset (Section~\ref{apx:empirical-study}), and more qualitative results (Section~\ref{apx:more_qualitative}). 
    
\end{abstract}

\section{Comparison with Adversarial Score Distillation (ASD)}\label{apx:proof-asd-jd}
Our method is relevant to ASD~\cite{asd} in that both methods are built upon theories of GAN~\cite{gan,wgan}. ASD explains ProlificDreamer~\cite{vsd} using the adversarial training framework. ASD defines an optimizable discriminator as follows:
\begin{align}
    \mathcal{D}_{\rm asd}(\hat{x}_t; y, \phi) = \log\frac{p_\psi(y|\hat{x}_t)}{p(\phi|\hat{x}_t)},
\end{align}
where $\hat{x}_t = \alpha_t \hat{x}_0 + \sigma_t \epsilon$, $y$ and $\phi$ are real and fake prompts, respectively. 
One drawback of such a definition is that the log-likelihood $p(\phi|\hat{x}_t)$ is intractable and must be approximated by training the external LoRA~\cite{lora}, making ASD optimization process similar to the alternating optimization scheme used in variational score distillation (VSD)~\cite{vsd}. 

While our log-odds classifier has a similar form to the discriminator in ASD, the fundamental difference between our method and ASD lies in the assumption of the discriminator. 
ASD follows the traditional adversarial training scheme with a trainable discriminator. 
Contrastively, our method assumes an \emph{optimal} discriminator so that our optimization only requires training the generator, making adversarial training no longer necessary. 

\begin{figure}[!tb]
    \centering
    \includegraphics[width=\linewidth]{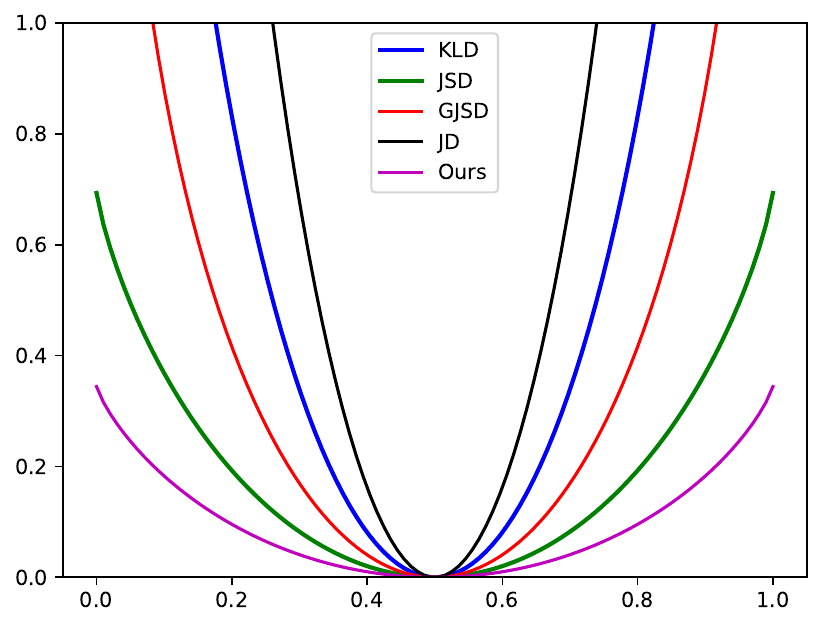}
    \caption{Comparison on different divergences: KLD, JD, JSD, Geometric JSD (GJSD), and our approximated JSD}
    \label{fig:divergence}
\end{figure}

In Fig.~\ref{fig:divergence}, for a better comparison among the divergences, we plot different divergences on two probability distributions: $p = (a, 1-a)$ and $q = (1-a, a)$, where $a$ was linearly spaced between $0$ and $1$ with $100$ points.
As can be seen, JSD and our approximated JSD are the lower bound of other divergences. 

\section{JSD-based Objective}
\label{apx:proof-jsd-logodd}

We provide a detailed derivation of the JSD objective based on the theory of generative adversarial networks (GANs). 

\begin{proof}
    Let the discriminator be $D(\hat{x}_t; y)$. Following GAN~\cite{gan,gan-modecollapse-theory} min-max optimization, the overall GAN criterion is derived as follows:
    \begin{align}\label{eq:min-max}
        &V(\mathcal{G}(\theta), \mathcal{D}) = \int_t\int_{\epsilon}\Big[p_\psi(\hat{x}_t | y)\log D(\hat{x}_t; y) \notag\\
        &+ q(\hat{x}_t|\hat{x}_0)\log(1 - D(\hat{x}_t; y))\Big]{\rm d}\mu(t)\eta(\epsilon),\notag\\
        % &= \int_t\int_{\epsilon}\Big[p_\psi (\hat{x}_t|y) \frac{p_\psi(\hat{x}_t|y)}{1 - p_\psi(\hat{x}_t|y)} \notag\\
        % &- q(\hat{x}_t|\hat{x}_0)\log\frac{p_\psi(\hat{x}_t|y)}{1 - p_\psi(\hat{x}_t|y)}\Big]{\rm d}\mu(t)\eta(\epsilon)\notag\\
    \end{align}
    where $\epsilon\sim\mathcal{N}(0, \mathbb{I})$. The optimal solution for the discriminator is then:
    \begin{align}
        \mathcal{D}^*(\hat{x}_t; y) = \frac{p_\psi(\hat{x}_t | y)}{p_\psi(\hat{x}_t | y) + q(\hat{x}_t|\hat{x}_0)},
    \end{align}
    Substituting the optimal discriminator back to Eq. \eqref{eq:min-max}, we have
    \begin{align}
        &V(\mathcal{G}(\theta), \mathcal{D}^*) = \int_t\int_{\epsilon}\Big[p_\psi (\hat{x}_t|y)\log \frac{p_\psi(\hat{x}_t|y)}{p_\psi(\hat{x}_t|y) + q(\hat{x}_t|\hat{x}_0)}\notag\\ 
        &+ q(\hat{x}_t|\hat{x}_0)\log \frac{q(\hat{x}_t|\hat{x}_0)}{p_\psi(\hat{x}_t|y) + q(\hat{x}_t|\hat{x}_0)}\Big]{\rm d}\mu(\epsilon)\eta(t) \notag\\
        &= \int_t\int_{\epsilon}\Big[p_\psi (\hat{x}_t|y)\log \frac{2p_\psi(\hat{x}_t|y)}{p_\psi(\hat{x}_t|y) + q(\hat{x}_t|\hat{x}_0)}\notag\\ 
        &+ q(\hat{x}_t|\hat{x}_0)\log \frac{2q(\hat{x}_t|\hat{x}_0)}{p_\psi(\hat{x}_t|y) + q(\hat{x}_t|\hat{x}_0)} - \log(4)\Big]{\rm d}\mu(\epsilon)\eta(t) \notag\\
        &= \JSD(q(\hat{x}_t|\hat{x}_0)\|p_\psi(\hat{x}_t|y)) - \int_t\int_{\epsilon}\log(4){\rm d}\mu(\epsilon)\eta(t),
        %&\leq \JSD(q(\hat{x}_t|\hat{x}_0)\|p_\psi(\hat{x}_t|y))
        \label{eq:jsd_optimal}
    \end{align}
    where the term $\int_t\int_{ \epsilon \sim\mathcal{N}(0, \mathbb{I})} \log(4){\rm d}\mu(\epsilon)\eta(t)$ is positive and constant. Therefore, the gradient is equivalent to the gradient of JSD.
    \begin{align}
    \nabla_\theta V(\mathcal{G}(\theta), \mathcal{D}^*) = \nabla_\theta \JSD(q(\hat{x}_t|\hat{x}_0) \|p_\psi(\hat{x}_t|y)). 
    \end{align}
    %To conclude by using the alternative form of discriminator, we can conduct a loss function that is a strict lower-bound of JSD when the constant term is considerable or $t\rightarrow\infty$.
    This means that when the discriminator is optimal, optimizing the generator objective is equivalent to optimizing a JSD objective. 
\end{proof}

\section{Approximating JSD}\label{apx:loss-derivation}
By assuming that our discriminator has the form: 
\begin{align}
\mathcal{D}(\hat{x}_t; y) = \frac{p_\psi(y|\hat{x}_t)}{1 - p_\psi(y|\hat{x}_t)}.
\end{align}
We define our objective for the generator as:
\begin{align}
    \mathcal{L}_{\mathcal{G}} &= \mathbb{E}_{t, \epsilon}\Big[ -\log \mathcal{D}(\hat{x}_t ; y) \Big] \notag\\
    &= \mathbb{E}_{t, \epsilon}\Big[-\log\frac{p_\psi(y|\hat{x}_t)}{1 - p_\psi(y|\hat{x}_t)}\Big] \notag \\
    &= \mathbb{E}_{t, \epsilon}\Big[\log(1 - p_\psi(y|\hat{x}_t)) - \log p_\psi(y|\hat{x}_t)\Big].
\end{align}
The derivative of this objective is as follows:
\begin{align}
    &\nabla_\theta\mathcal{L}_{G} = \nabla_\theta\mathbb{E}_{t, \epsilon}\Big[\log(1 - p_\psi(y|\hat{x}_t)) - \log p_\psi(y|\hat{x}_t)\Big] \notag\\
    &= \mathbb{E}_{t, \epsilon}\Big[\nabla_\theta\log(1 - p_\psi(y|\hat{x}_t)) - \nabla_\theta\log p_\psi(y|\hat{x}_t)\Big].
\end{align}
We then independently derive each gradient term. The right term can be factorized using the Bayesian Theorem as follows:
\begin{align}
    &\nabla_\theta\log p_\psi(y|\hat{x}_t) = \nabla_\theta\log \frac{p_\psi(\hat{x}_t|y)p_\psi(y)}{p_\psi(\hat{x}_t)} \notag\\
    &\propto \nabla_\theta\log p_\psi(\hat{x}_t|y) - \nabla_\theta\log p_\psi(\hat{x}_t|\oslash).
\end{align}
The left term can be derived via multiclass generalization of the logistic sigmoid~\cite{bishop} as follows:
\begin{align}
    &\nabla_\theta\log(1 - p_\psi(y|\hat{x}_t)) \notag\\
    &= \nabla_\theta\log p_\psi(\phi|\hat{x}_t) \notag\\
    &\propto \nabla_\theta\log p_\psi(\hat{x}_t|\phi) - \nabla_\theta\log p_\psi(\hat{x}_t|\oslash) \notag\\
    &\approx \nabla_\theta\log p_\psi(\bar{x}_t|y) - \nabla_\theta\log p_\psi(\bar{x}_t|\oslash), 
\end{align}
where $p(\phi| \hat{x}_t) \approx 1 - p_\psi(y|\hat{x}_t)$, assuming the existence of a prompt $\phi$ related to $y$, where $p(\phi| \hat{x}_t)$ indicates a low-density sample. The sample $\bar{x}_t$ is obtained the minority sampling scheme as in the main paper. 

Combining both terms, the gradient becomes:
\begin{align}
    &\nabla_\theta\mathcal{L} = \mathbb{E}_{t, \epsilon}\Big[\nabla_\theta\log p_\psi(\hat{x}_t|\oslash) - \nabla_\theta\log p_\psi(\hat{x}_t|y) \notag\\
    &+ \nabla_\theta\log p_\psi(\bar{x}_t|y) - \nabla_\theta\log p_\psi(\bar{x}_t|\oslash)\Big] \notag\\
    &= \mathbb{E}_{t, \epsilon}\Big[w(t)\frac{\alpha_t}{\sigma_t}(\hat{\epsilon}_\psi(\hat{x}_t, y) - \hat{\epsilon}_\psi(\bar{x}_t, y))\frac{\partial\hat{x}_0}{\partial\theta}\Big],
\end{align}
where $\hat{\epsilon}_\psi(\hat{x}_t, y) = \epsilon_\psi(\hat{x}_t, \oslash) + s(\epsilon_\psi(\hat{x}_t, y) - \epsilon_\psi(\hat{x}_t, \oslash))$ and $\hat{\epsilon}_\psi(\bar{x}_t, y) = \epsilon_\psi(\bar{x}_t, \oslash) + s(\epsilon_\psi(\bar{x}_t, y) - \epsilon_\psi(\bar{x}_t, \oslash))$, respectively. 

\section{Evaluation with T3Bench \& GPTEval3D}\label{apx:gpteval3d}

We provide a full comparison table for T3Bench~\cite{t3bench}, as shown in Table~\ref{tab:t3bench_results_apx}.

\begin{table*}[!ht]
    \caption{Comparative results for the text-to-3D tasks in T3Bench. The best results are \textbf{bold} while the second best results are \underline{underlined}.}
    \vspace{-5pt}
    \label{tab:t3bench_results_apx}
    \small
    \centering
    \setlength{\tabcolsep}{4.5pt} % Adjust column padding
    \begin{tabular}{lccccccccccc}
        \toprule
         & Conference &\textbf{Time} & \multicolumn{3}{c}{\textbf{Single Object}} & \multicolumn{3}{c}{\textbf{Single Object with Surr}} & \multicolumn{3}{c}{\textbf{Multiple Objects}} \\ 
        \cmidrule{3-12}
        \textbf{Method} & & (mins) & \textbf{Qual.} $\uparrow$ & \textbf{Align.} $\uparrow$ & \textbf{Avg} $\uparrow$
               & \textbf{Qual.} $\uparrow$ & \textbf{Align.} $\uparrow$ & \textbf{Avg} $\uparrow$
               & \textbf{Qual.} $\uparrow$ & \textbf{Align.} $\uparrow$ & \textbf{Avg} $\uparrow$\\ 
        \midrule
        Dreamfusion \cite{sds} & ICLR 2023 & 30 & 24.9 & 24.0 & 24.4 & 19.3 & 29.8 & 24.6 & 17.3 & 14.8 & 16.1 \\ 
        Magic3D \cite{magic3d} & CVPR 2023 & 40 & 38.7 & 35.3 & 37.0 & 29.8 & 41.0 & 35.4 & 26.6 & 24.8 & 25.7 \\ 
        LatentNeRF \cite{latentnerf} & CVPR 2023 & 65 & 34.2 & 32.0 & 33.1 & 23.7 & 37.5 & 30.6 & 21.7 & 19.5 & 20.6 \\ 
        Fantasia3D \cite{fantasia3d} & ICCV 2023 & 45 & 29.2 & 23.5 & 26.4 & 21.9 & 32.0 & 27.0 & 22.7 & 14.3 & 18.5 \\ 
        SJC \cite{sjc} & CVPR 2023 & 25 & 26.3 & 23.0 & 24.7 & 17.3 & 22.3 & 19.8 & 11.7 & 5.8 & 8.7 \\ 
        VSD \cite{vsd} & NIPS 2023 & 240 & 51.1 & 47.8 & 49.4 & 42.5 & 47.0 & 44.8 & 45.7 & 25.8 & 35.8 \\ 
        MVDream \cite{mvdream} & ICLR 2024 & 30 & 53.2 & 42.3 & 47.8 & 36.3 & 48.5 & 42.4 & 39.0 & 28.5 & 33.8 \\ 
        DreamGaussian \cite{dreamgaussian} & ICLR 2024 & 7 & 19.9 & 19.8 & 19.8 & 10.4 & 17.8 & 14.1 & 12.3 & 9.5 & 10.9 \\
        RichDreamer \cite{richdreamer} & CVPR 2024 & 70 & \underline{57.3} & 40.0 & 48.6 & 43.9 & 42.3 & 43.1 & 34.8 & 22.0 & 28.4 \\ 
        VP3D \cite{vp3d} & CVPR 2024 & - & 54.8 & 52.2 & 53.5 & 45.4 & 50.8 & 48.1 & 49.1 & 31.5 & 40.3 \\ 
        ModeDreamer \cite{isd} & Arxiv 2024 & \underline{40} & 55.4 & 52.6 & 54.0 & \underline{45.7} & \textbf{59.0} & 52.4 & 43.4 & \underline{39.4} & 41.4 \\
        DreamReward$^{*}$~\cite{dreamreward} & ECCV 2024 & \underline{40} & 54.3 & 43.8 & 49.0 & 38.2 & 49.6 & 43.9 & 41.2 & 33.6 & 37.4 \\
        DreamMesh \cite{dreammesh} & ECCV 2024 & 30 & 55.6 & \textbf{53.8} & 54.7 & 43.1 & 54.3 & \underline{48.7} & 47.6 & 30.8 & 39.2 \\ 
        Trellis~\cite{trellis} & CVPR 2025 & $\mathbf{<1}$ & 35.6 & 21.4 & 28.5 & 19.5 & 17.5 & 18.5 & 16.6 & 17.0 & 16.8\\
        CompGS \cite{compgs} & CVPR 2025 & 30 & 55.1 & 52.5 & 53.8 & 43.2 & 46.8 & 45.0 & \textbf{54.2} & 37.9 & \textbf{46.1} \\
        \midrule
        Ours & & 70 & \textbf{58.7} & \underline{53.6} & \textbf{56.1} & \textbf{47.4} & \underline{57.6} & \textbf{52.5} & \underline{51.3} & \textbf{40.2} & \underline{45.7} \\
        \bottomrule
    \end{tabular}
    \vspace{-5pt}
\end{table*}

We further present additional quantitative results on the GPTEval3D~\cite{gpteval3d} benchmark (refers to Tab.~\ref{tab:gpt_eval3d}). We conducted the experiments with 110 prompts provided by the benchmark on six aspects: Text-asset alignment, 3D plausibility, Texture details, Geometry details, Texture-geometry coherence, and overall.

\begin{table*}[!ht]
    \centering
    \setlength{\tabcolsep}{5pt}
    \begin{tabular}{lp{2cm}p{1.6cm}p{2.5cm}p{1.4cm}p{1.6cm}p{2cm}}
        \toprule
        \textbf{Methods} & \textbf{Text-Asset Alignment} $\uparrow$ & \textbf{3D  Plausibility} $\uparrow$ & \textbf{Text-Geometry Alignment} $\uparrow$ & \textbf{Texture Details} $\uparrow$ & \textbf{Geometry Details} $\uparrow$ & \textbf{Overall} $\uparrow$ \\
        \midrule
        Dreamfusion~\cite{sds}      & 1000 & 1000 & 1000 & 1000 & 1000 & 1000 \\
        Magic3D~\cite{magic3d}          & 1152 & 1001 & 1084 & 1178 & 1100 & 1100 \\
        SJC~\cite{sjc}              & 1130 & 995 & 1034 & 1080 & 1043 & 1056 \\
        LatentNeRF~\cite{latentnerf}       & 1222 & 1145 & 1157 & 1180 & 1161 & 1173 \\
        Fantasia3D~\cite{fantasia3d}       & 1068 & 892 & 1006 & 1109 & 1027 & 1021 \\
        VSD~\cite{vsd}  & 1262 & 1059 & 1152 & 1246 & 1181 & 1180 \\
        MVDream~\cite{mvdream}          & 1271 & 1147 & 1251 & 1325 & 1255 & 1250 \\
        RichDreamer~\cite{richdreamer}      & 1295 & 1225 & 1260 & 1356 & 1251 & 1277 \\
        DreamReward$^*$~\cite{dreamreward} & \underline{1567} & \textbf{1305} & \underline{1282} & \textbf{1492} & \textbf{1320} & \underline{1393} \\
        Instant3D~\cite{instant3d}        & 1200 & 1088 & 1153 & 1152 & 1181 & 1155 \\
        DreamGaussian~\cite{dreamgaussian}    & 1101 & 954 & 1159 & 1126 & 1131 & 1094 \\
        One2345~\cite{one2345}          & 872 & 829 & 850 & 911 & 860 & 864 \\
        Shap-E~\cite{shap-e}         & 843 & 842 & 846 & 784 & 846 & 836 \\
        Point-E~\cite{point-e}         & 725 & 690 & 689 & 716 & 746 & 713 \\
        \midrule
         Ours & \textbf{1572} & \underline{1291} & \textbf{1371} & \underline{1488} & \underline{1286} & \textbf{1401} \\
        \bottomrule
    \end{tabular}
    \caption{Comparision with text-to-3D methods using GPTEval3D benchmark. Results of DreamReward~\cite{dreamreward} were obtained using DreamReward’s publicly released pretrained models. While the performance is consistent, it is slightly lower than the figures reported in their paper, possibly due to differences in evaluation setup, environment versions, or implementation details not fully disclosed. We provided results of our proposed method by optimizing both NeRF~\cite{nerf} and Gaussian Splatting~\cite{gs} (GS).}
    \label{tab:gpt_eval3d}
\end{table*}

\section{Empirical Analysis with Toy Dataset}\label{apx:empirical-study}

\begin{figure}[!ht]
    \centering
    \includegraphics[width=0.8\linewidth]{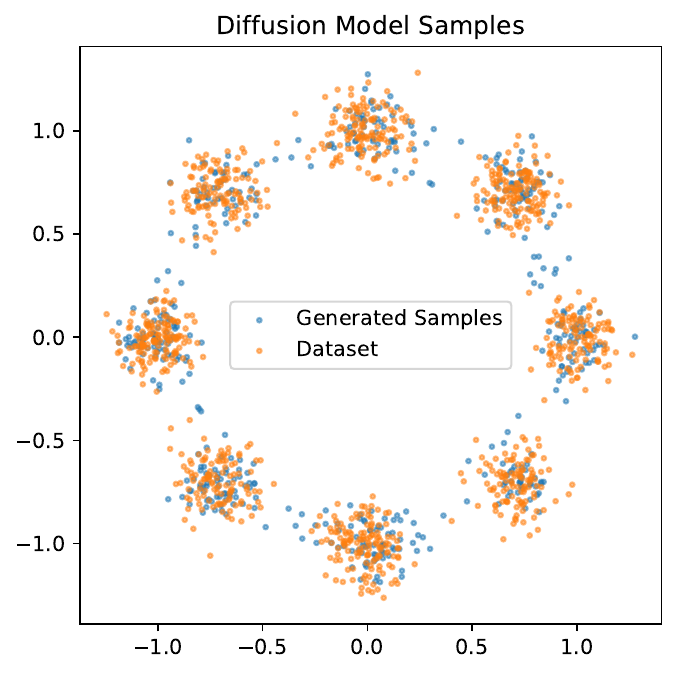}
    \caption{Visualization of generated and ground truth data points. }
    \label{fig:diff-samples}
\end{figure}

\hd{Experimental settings.} In this experiment, we create a synthetic dataset with 8 modes using a mixture of 8 two-dimensional Gaussian distributions. Each mode is centered at one of the following coordinates: $(1, 0)$, $(-1, 0)$, $(0, 1)$, $(0, -1)$, $(\frac{1}{\sqrt{2}}, \frac{1}{\sqrt{2}})$, $(-\frac{1}{\sqrt{2}}, \frac{1}{\sqrt{2}})$, $(\frac{1}{\sqrt{2}}, -\frac{1}{\sqrt{2}})$, and $(-\frac{1}{\sqrt{2}}, -\frac{1}{\sqrt{2}})$ (refers to Fig.~\ref{fig:diff-samples}). A standard deviation of $0.1$ is used for all modes.

We use a diffusion model with the following architecture:
\begin{itemize}
    \item \textbf{Time Embedding:} A fully connected network with two layers, activated by ReLU, which embeds the time step into a higher-dimensional space.
    \item \textbf{Main Network:} The network takes the concatenation of the data point and its time embedding as input. It comprises multiple fully connected layers with $128$ hidden units, ReLU activations, and LayerNorm for normalization.
\end{itemize}

A noise scheduler with $1000$ timesteps is used to add noise progressively to the data. The scheduler linearly interpolates $\beta = 1 - \alpha_t$ from $1 \times 10^{-4}$ to $0.02$ and computes cumulative products to obtain $\alpha_t$ and $\sigma_t$ values.

For training, the model is optimized using the Adam optimizer with a learning rate of $1 \times 10^{-3}$. A CosineAnnealingLR learning rate scheduler is employed, and the Mean Squared Error (MSE) loss function is used to compare the predicted noise with the actual noise. The model is trained for $1000$ epochs with a batch size of $128$.

\begin{figure}[!ht]
    \centering
    % First row of subfigures (no captions)
    \begin{subfigure}[b]{0.09\textwidth}
        \includegraphics[width=\textwidth]{imgs/analysis/distillation/sds/noaxis_trials_10_start__-1.0_1.0_.pdf}
    \end{subfigure}
    \begin{subfigure}[b]{0.09\textwidth}
        \includegraphics[width=\textwidth]{imgs/analysis/distillation/sds/noaxis_trials_100_start__-1.0_1.0_.pdf}
    \end{subfigure}
    \begin{subfigure}[b]{0.09\textwidth}
        \includegraphics[width=\textwidth]{imgs/analysis/distillation/sds/noaxis_trials_200_start__-1.0_1.0_.pdf}
    \end{subfigure}
    \begin{subfigure}[b]{0.09\textwidth}
        \includegraphics[width=\textwidth]{imgs/analysis/distillation/sds/noaxis_trials_500_start__-1.0_1.0_.pdf}
    \end{subfigure}
    \begin{subfigure}[b]{0.09\textwidth}
        \includegraphics[width=\textwidth]{imgs/analysis/distillation/sds/noaxis_trials_1000_start__-1.0_1.0_.pdf}
    \end{subfigure}
    \begin{subfigure}[b]{0.09\textwidth}
        \includegraphics[width=\textwidth]{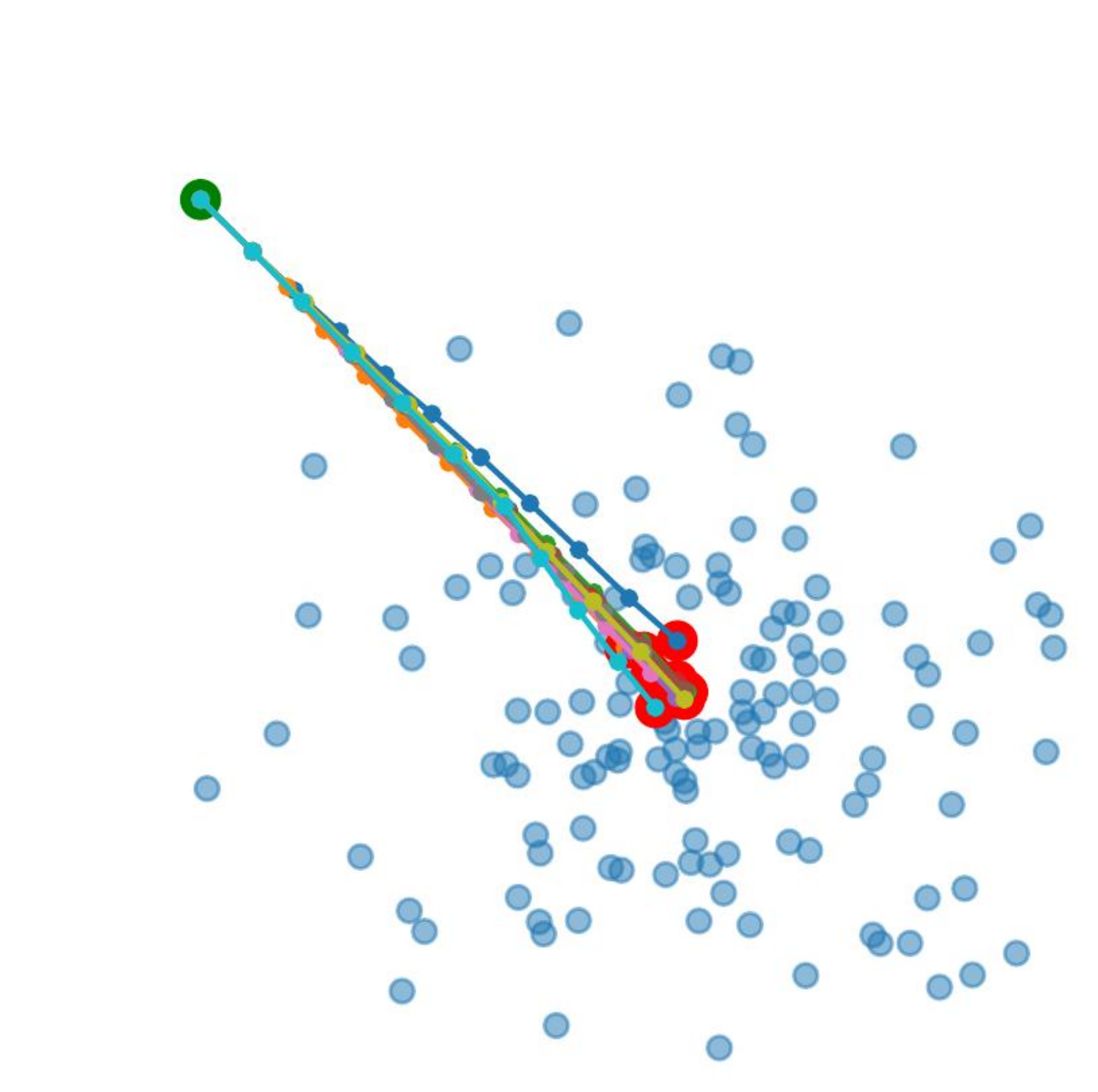}
        \caption{10}
    \end{subfigure}
    \begin{subfigure}[b]{0.09\textwidth}
        \includegraphics[width=\textwidth]{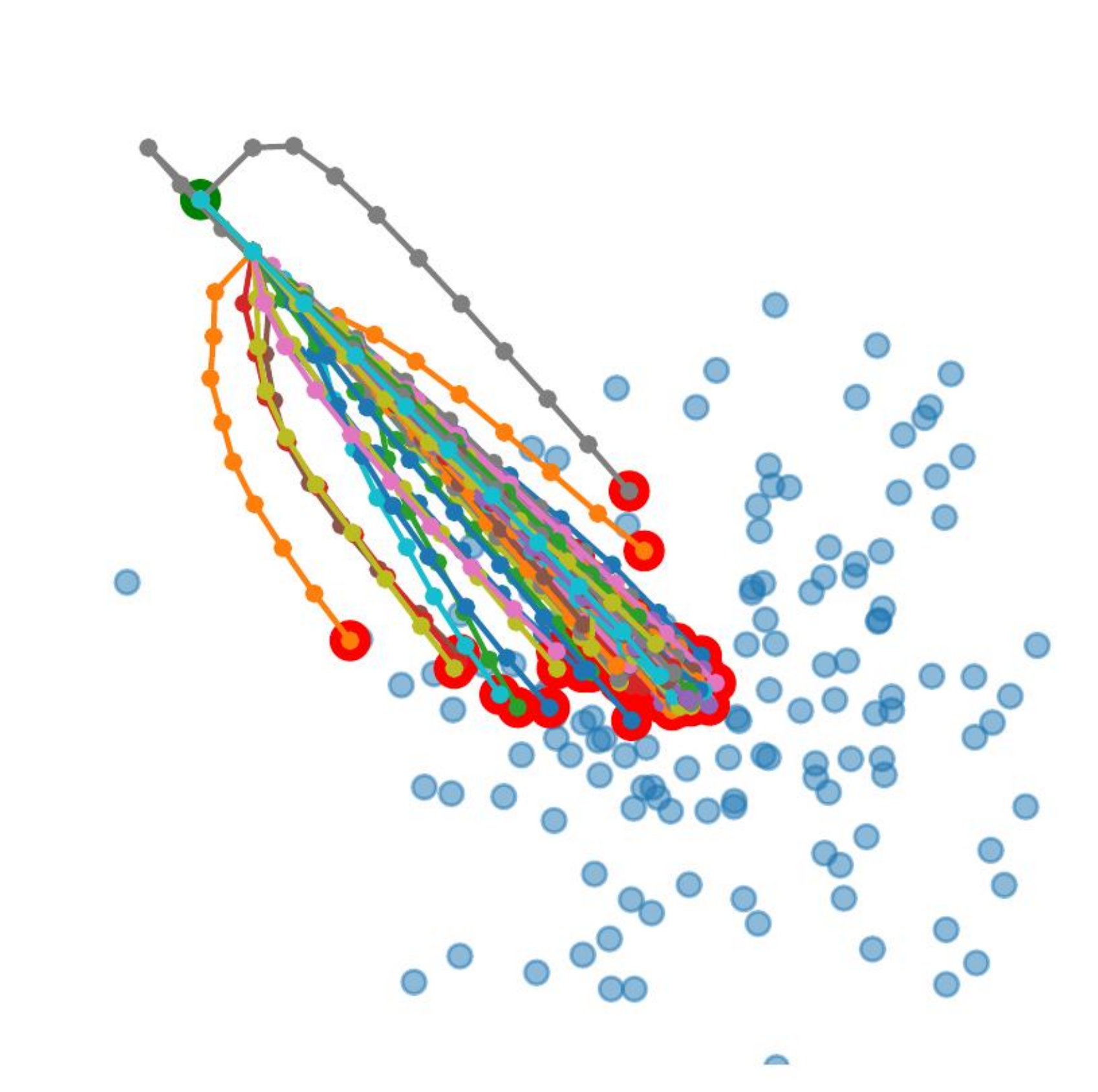}
        \caption{100}
    \end{subfigure}
    \begin{subfigure}[b]{0.09\textwidth}
        \includegraphics[width=\textwidth]{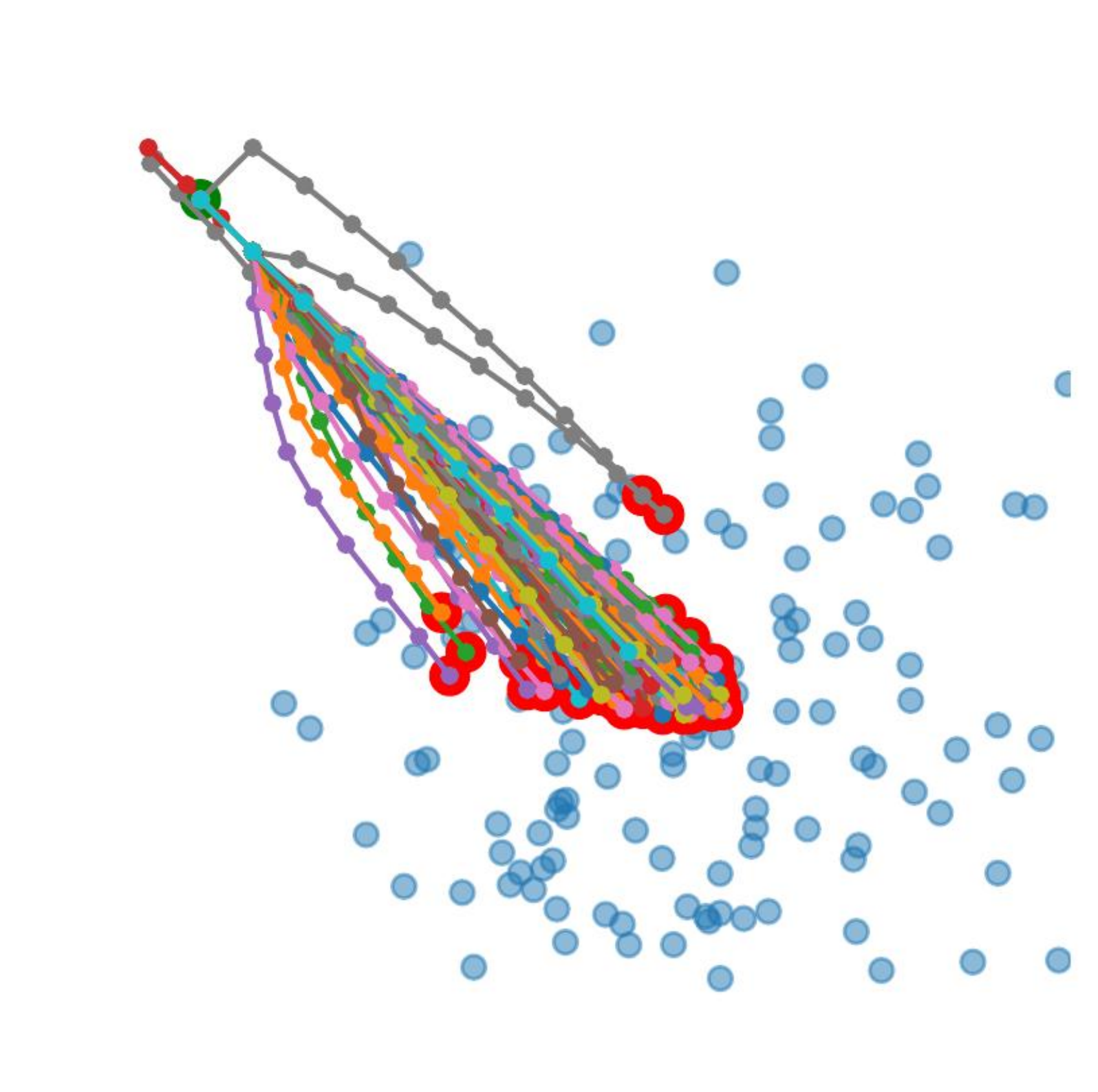}
        \caption{200}
    \end{subfigure}
    \begin{subfigure}[b]{0.09\textwidth}
        \includegraphics[width=\textwidth]{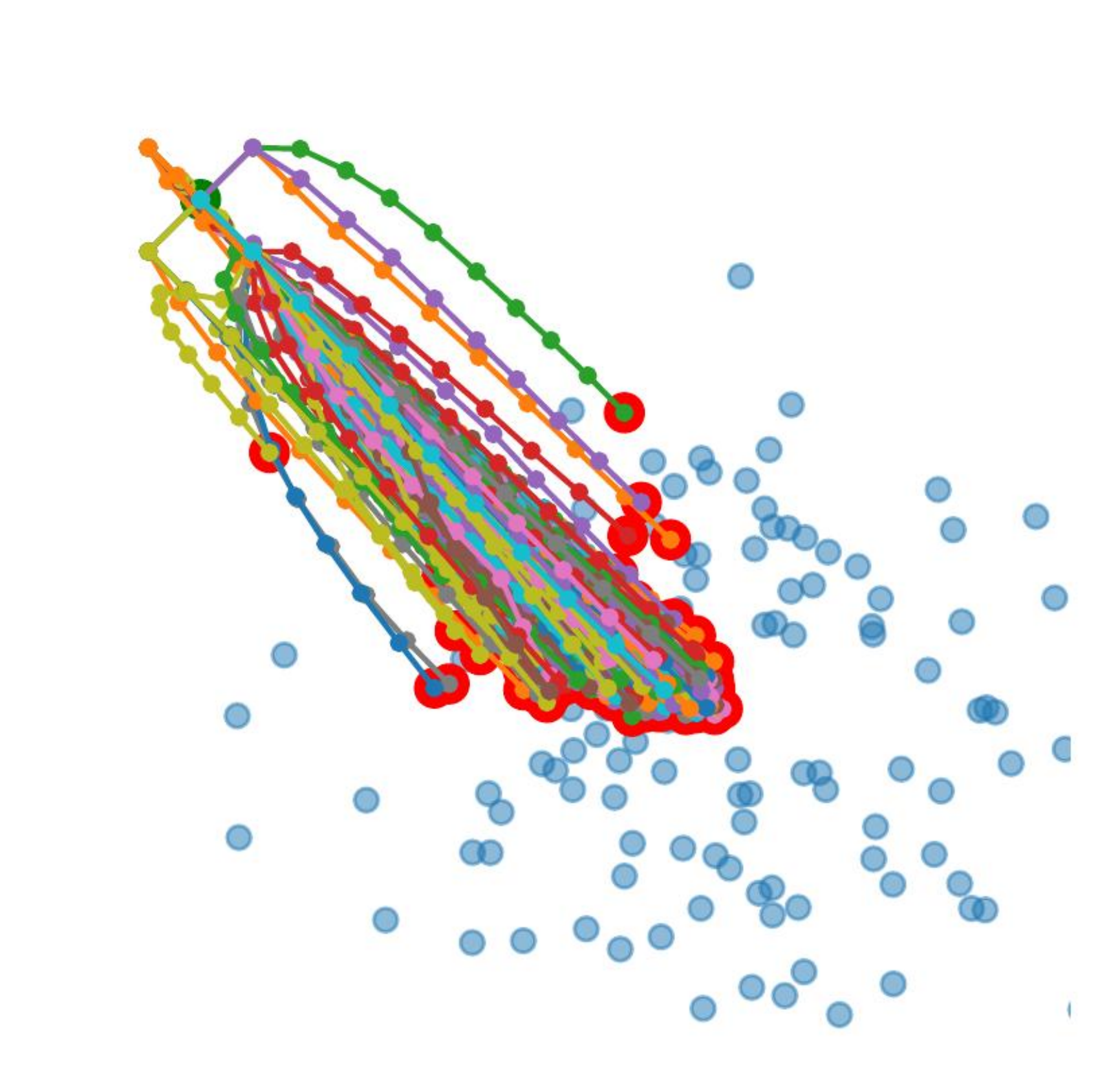}
        \caption{500}
    \end{subfigure}
    \begin{subfigure}[b]{0.09\textwidth}
        \includegraphics[width=\textwidth]{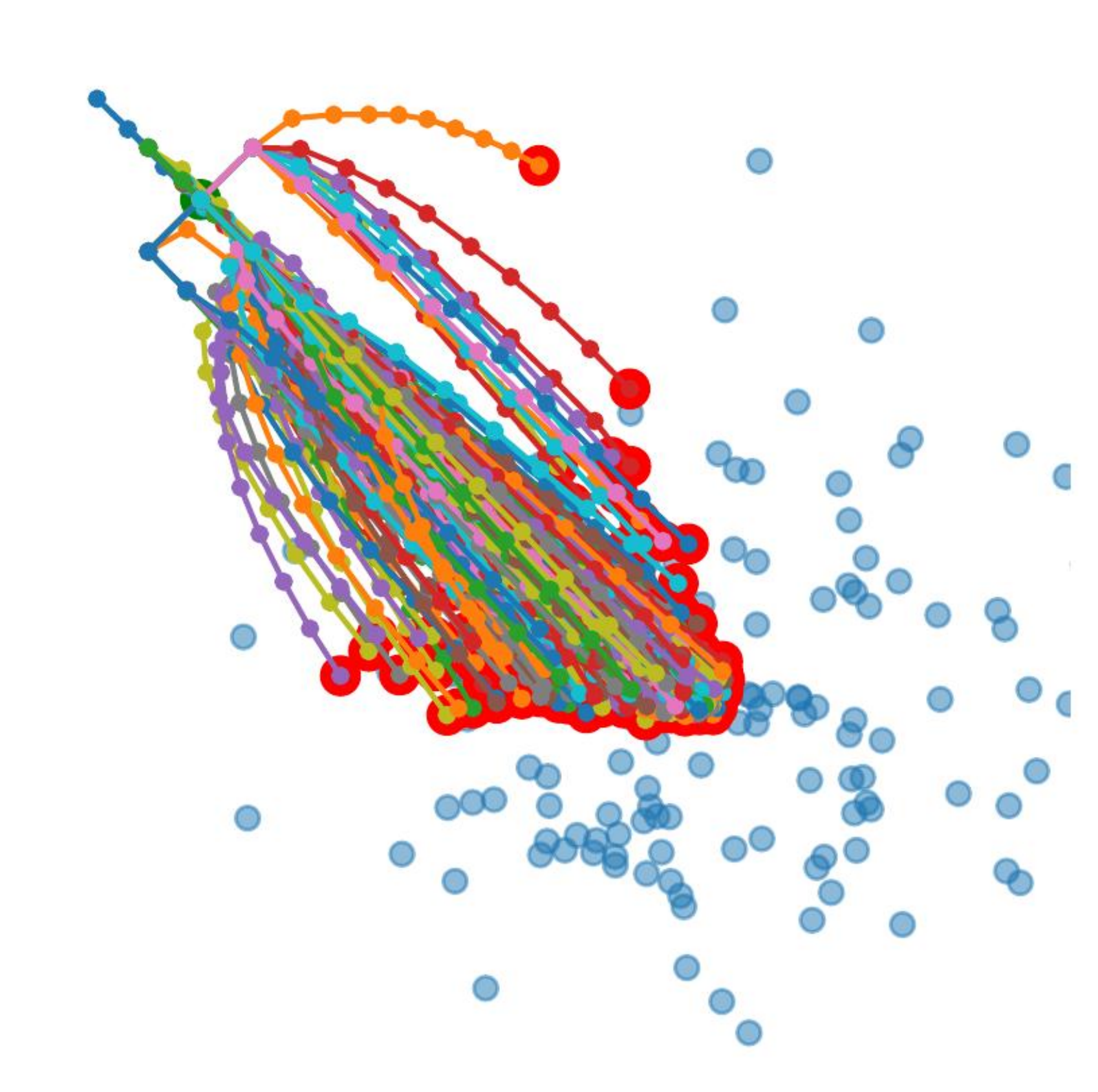}
        \caption{1000}
    \end{subfigure}
    \caption{Initialization point: $(-1, 1)$. Top row: SDS, bottom row: proposed method. {\color{teal}{$\bullet$}}: Starting point. {\color{red}{$\bullet$}}: Ending point.}
    \label{fig:1ddiff-sample-stl}
\end{figure}

\begin{figure}[!ht]
    \centering
    % First row of subfigures (no captions)
    \begin{subfigure}[b]{0.09\textwidth}
        \includegraphics[width=\textwidth]{imgs/analysis/distillation/sds/noaxis_trials_10_start__1.0_-1.0_.pdf}
    \end{subfigure}
    \begin{subfigure}[b]{0.09\textwidth}
        \includegraphics[width=\textwidth]{imgs/analysis/distillation/sds/noaxis_trials_100_start__1.0_-1.0_.pdf}
    \end{subfigure}
    \begin{subfigure}[b]{0.09\textwidth}
        \includegraphics[width=\textwidth]{imgs/analysis/distillation/sds/noaxis_trials_200_start__1.0_-1.0_.pdf}
    \end{subfigure}
    \begin{subfigure}[b]{0.09\textwidth}
        \includegraphics[width=\textwidth]{imgs/analysis/distillation/sds/noaxis_trials_500_start__1.0_-1.0_.pdf}
    \end{subfigure}
    \begin{subfigure}[b]{0.09\textwidth}
        \includegraphics[width=\textwidth]{imgs/analysis/distillation/sds/noaxis_trials_1000_start__1.0_-1.0_.pdf}
    \end{subfigure}
    \begin{subfigure}[b]{0.09\textwidth}
        \includegraphics[width=\textwidth]{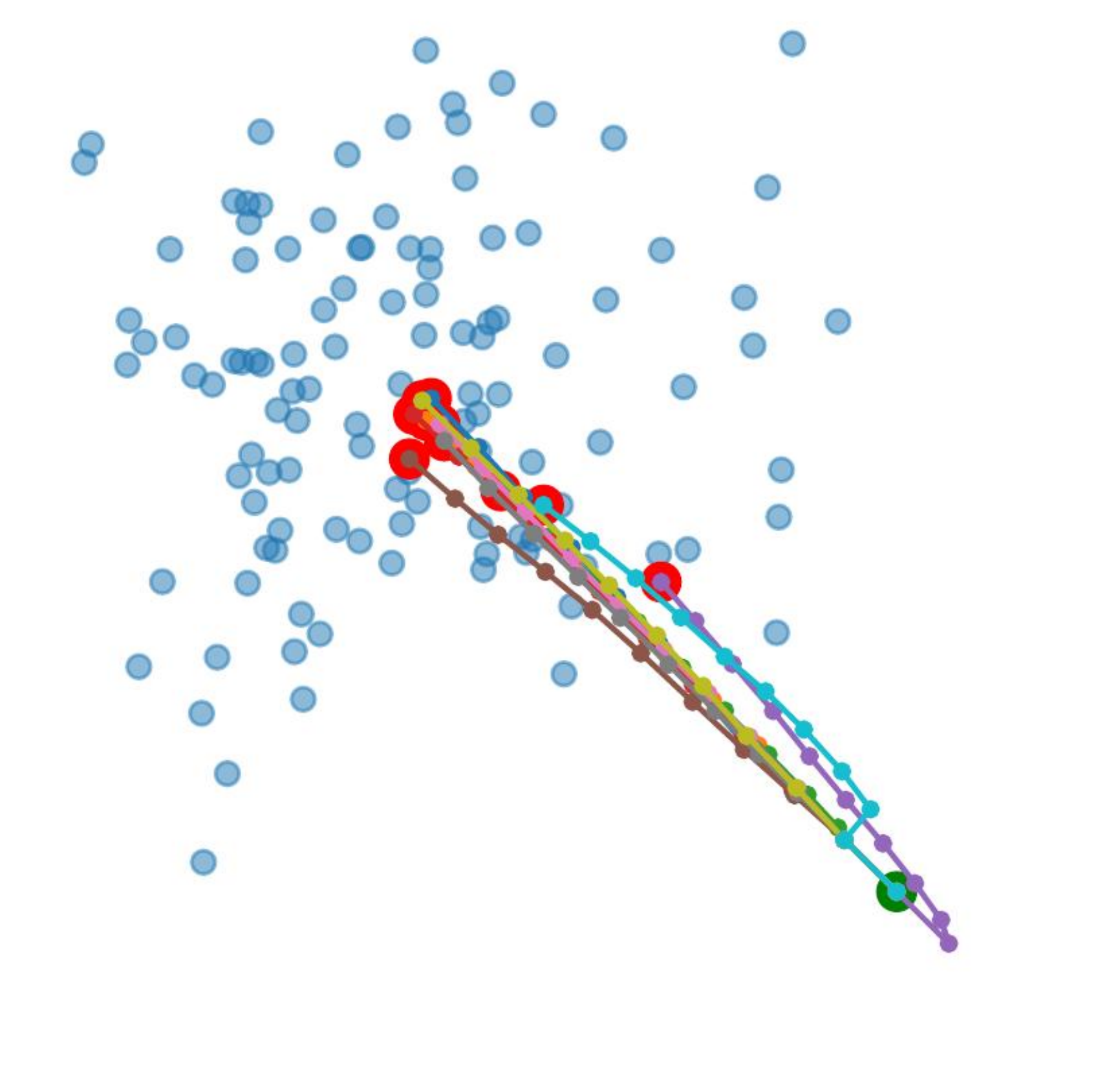}
        \caption{10}
    \end{subfigure}
    \begin{subfigure}[b]{0.09\textwidth}
        \includegraphics[width=\textwidth]{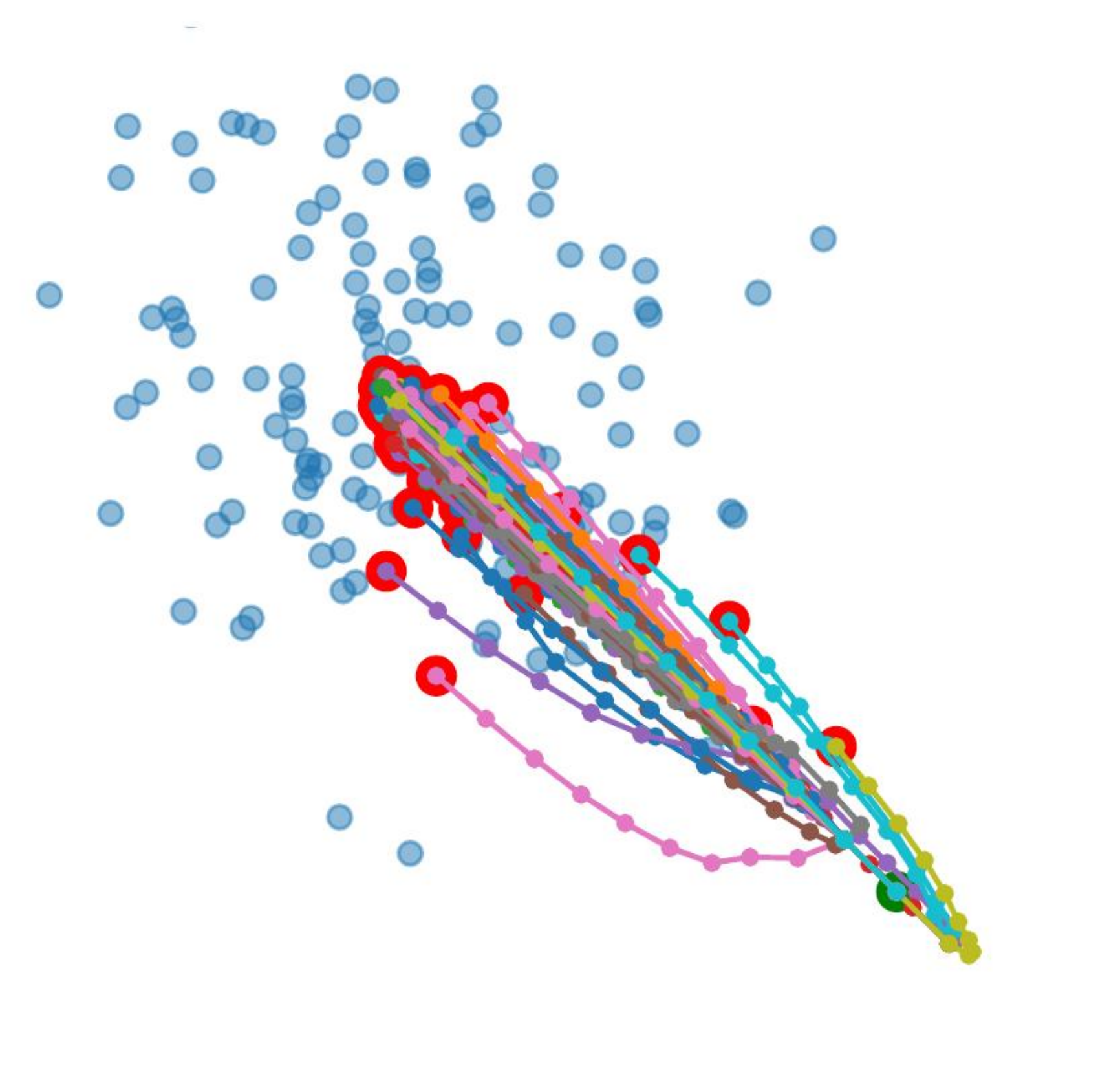}
        \caption{100}
    \end{subfigure}
    \begin{subfigure}[b]{0.09\textwidth}
        \includegraphics[width=\textwidth]{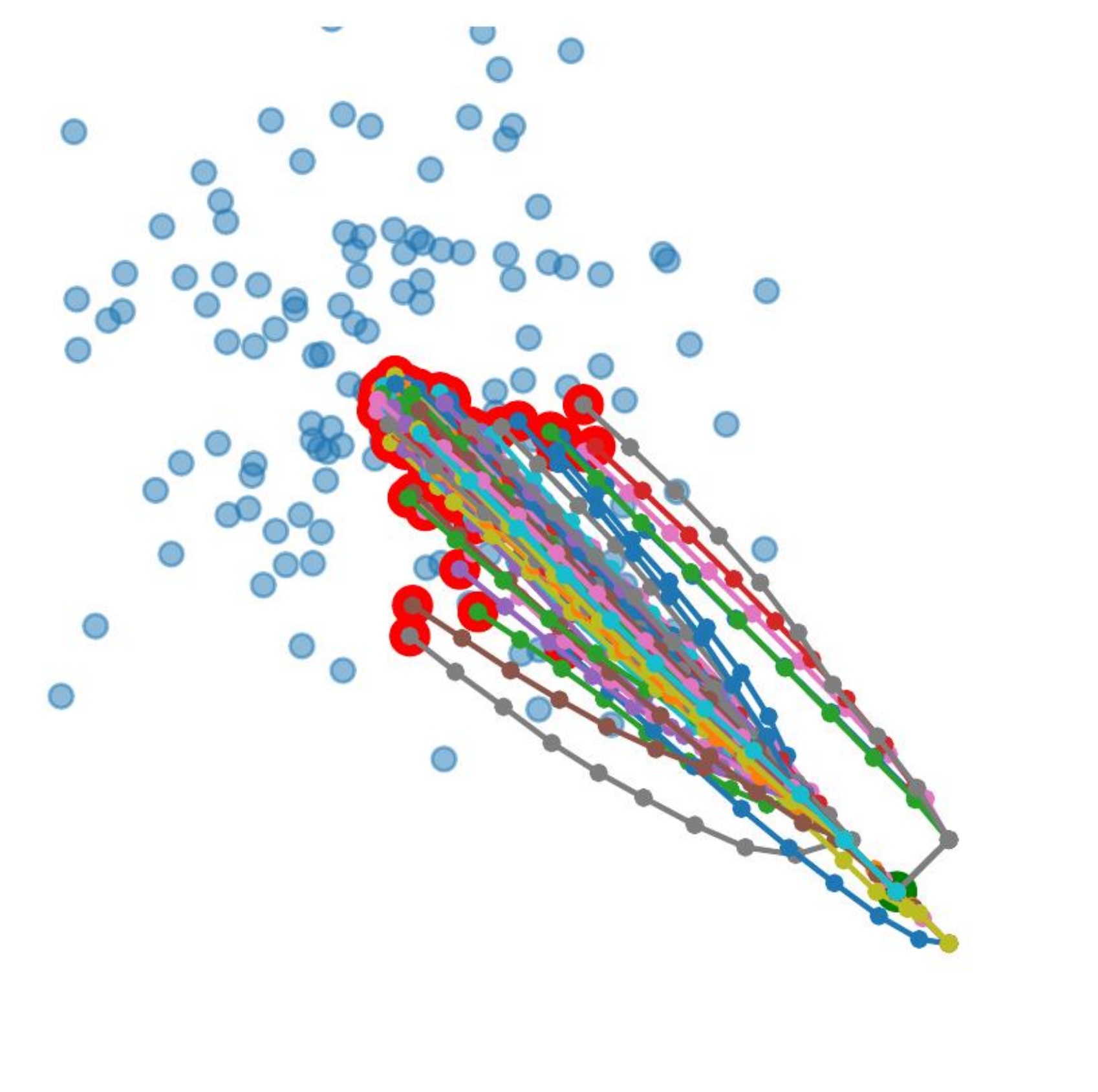}
        \caption{200}
    \end{subfigure}
    \begin{subfigure}[b]{0.09\textwidth}
        \includegraphics[width=\textwidth]{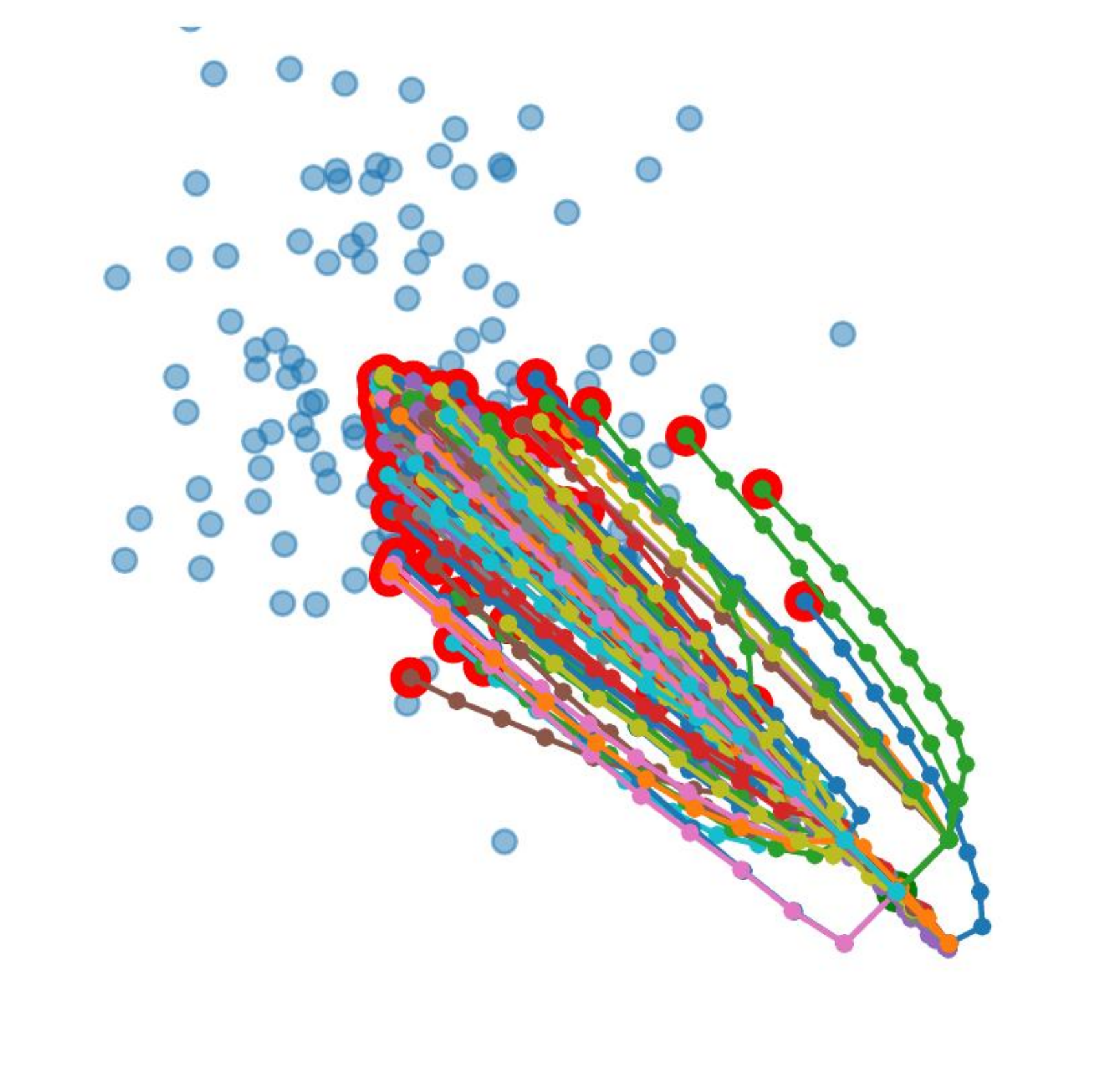}
        \caption{500}
    \end{subfigure}
    \begin{subfigure}[b]{0.09\textwidth}
        \includegraphics[width=\textwidth]{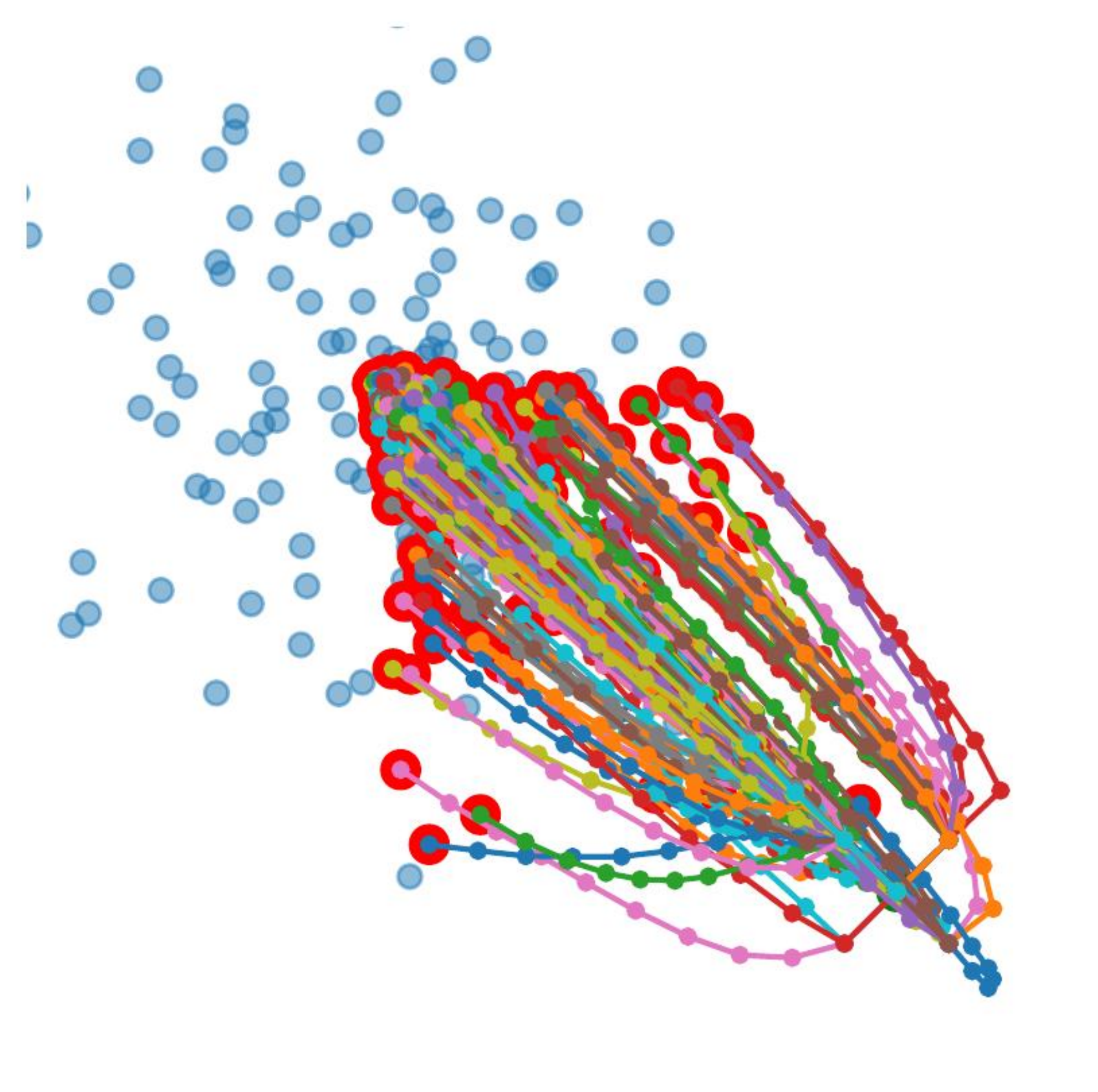}
        \caption{1000}
    \end{subfigure}
    \caption{Initialization point: $(1, -1)$. Top row: SDS, bottom row: proposed method. {\color{teal}{$\bullet$}}: Starting point. {\color{red}{$\bullet$}}: Ending point.}
    \label{fig:1ddiff-sample-sdr}
\end{figure}

\begin{figure}[!ht]
    \centering
    % First row of subfigures (no captions)
    \begin{subfigure}[b]{0.09\textwidth}
        \includegraphics[width=\textwidth]{imgs/analysis/distillation/sds/noaxis_trials_10_start__-1.0_-1.0_.pdf}
    \end{subfigure}
    \begin{subfigure}[b]{0.09\textwidth}
        \includegraphics[width=\textwidth]{imgs/analysis/distillation/sds/noaxis_trials_100_start__-1.0_-1.0_.pdf}
    \end{subfigure}
    \begin{subfigure}[b]{0.09\textwidth}
        \includegraphics[width=\textwidth]{imgs/analysis/distillation/sds/noaxis_trials_200_start__-1.0_-1.0_.pdf}
    \end{subfigure}
    \begin{subfigure}[b]{0.09\textwidth}
        \includegraphics[width=\textwidth]{imgs/analysis/distillation/sds/noaxis_trials_500_start__-1.0_-1.0_.pdf}
    \end{subfigure}
    \begin{subfigure}[b]{0.09\textwidth}
        \includegraphics[width=\textwidth]{imgs/analysis/distillation/sds/noaxis_trials_1000_start__-1.0_-1.0_.pdf}
    \end{subfigure}
    \begin{subfigure}[b]{0.09\textwidth}
        \includegraphics[width=\textwidth]{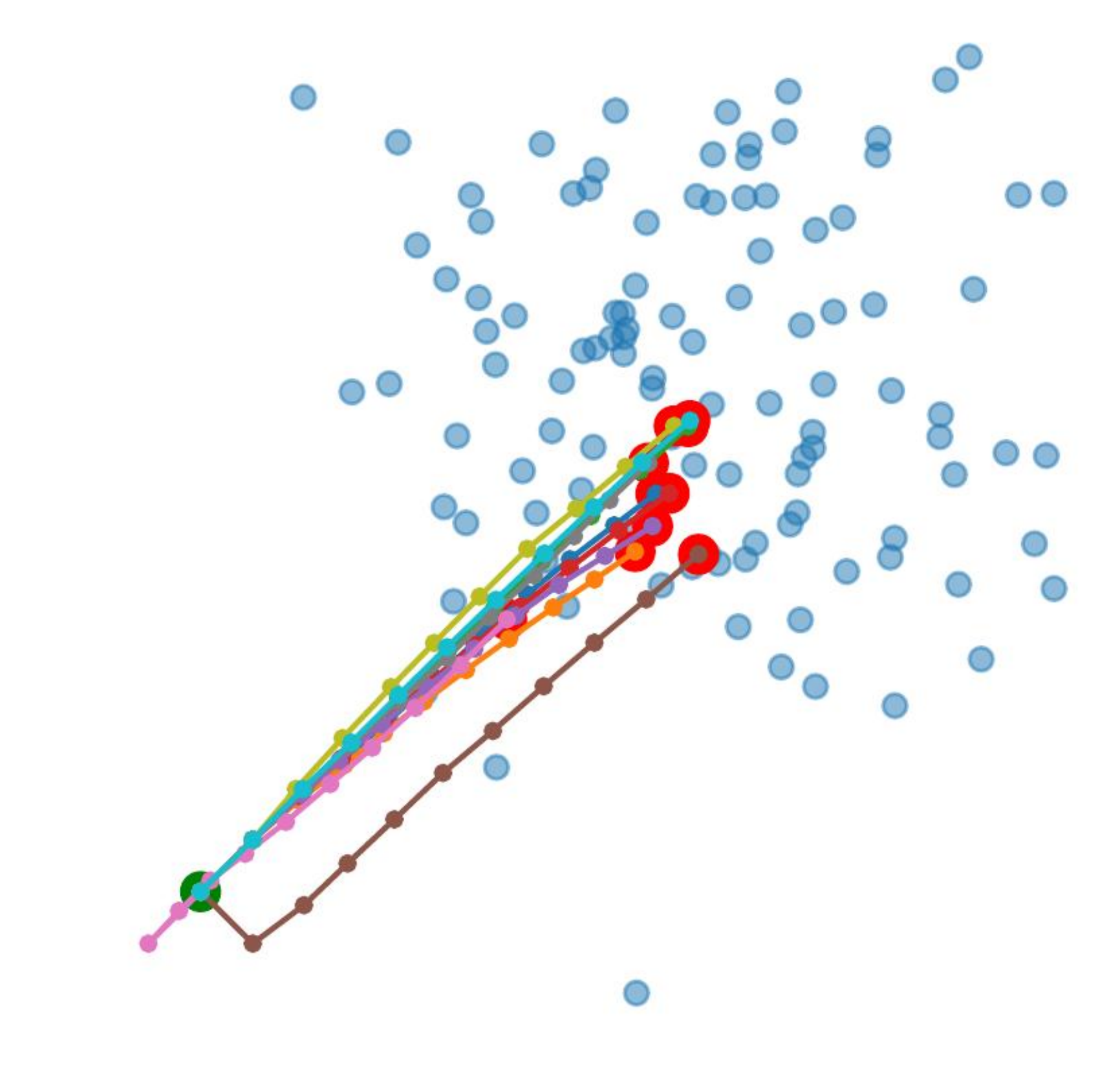}
        \caption{10}
    \end{subfigure}
    \begin{subfigure}[b]{0.09\textwidth}
        \includegraphics[width=\textwidth]{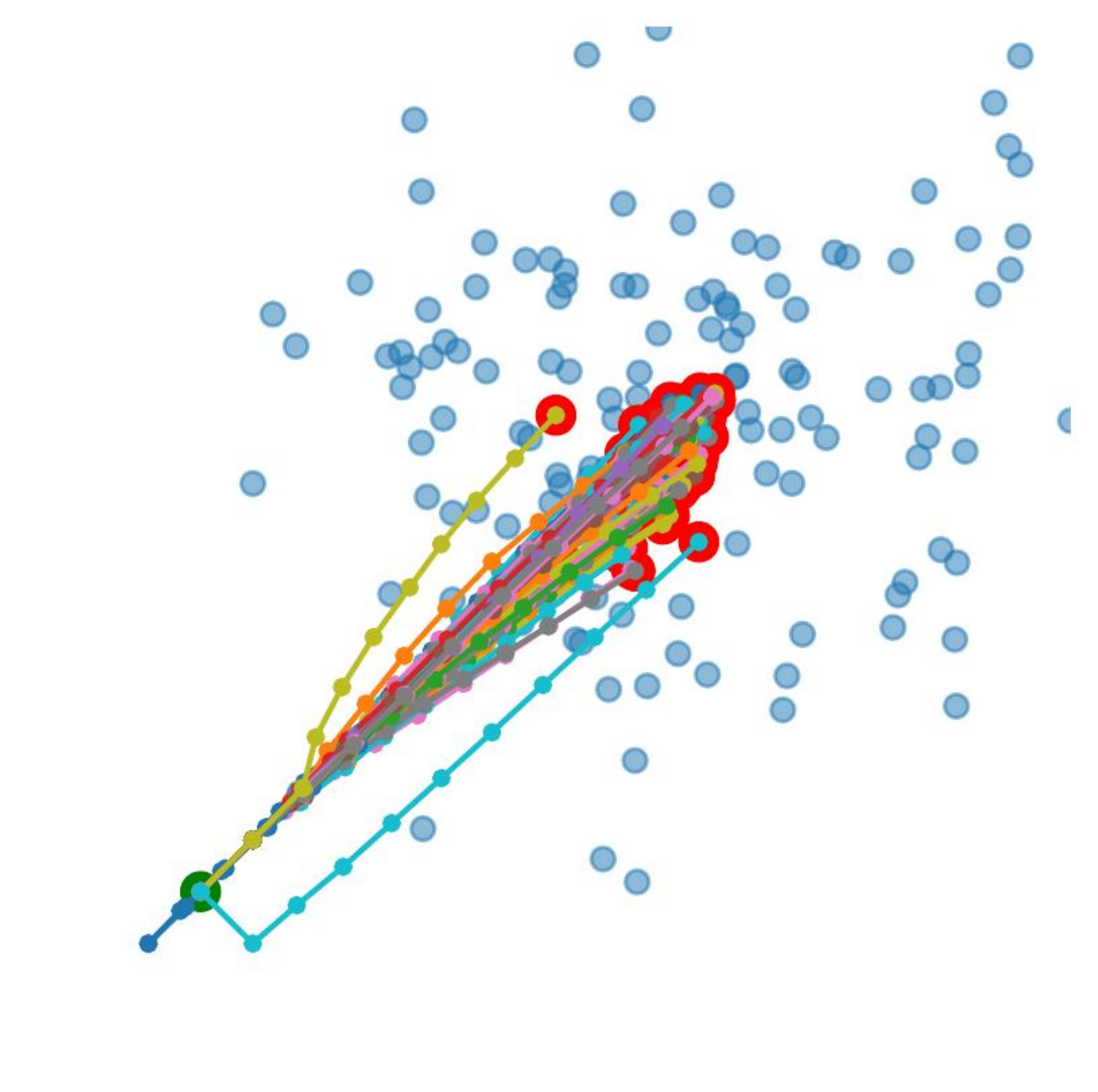}
        \caption{100}
    \end{subfigure}
    \begin{subfigure}[b]{0.09\textwidth}
        \includegraphics[width=\textwidth]{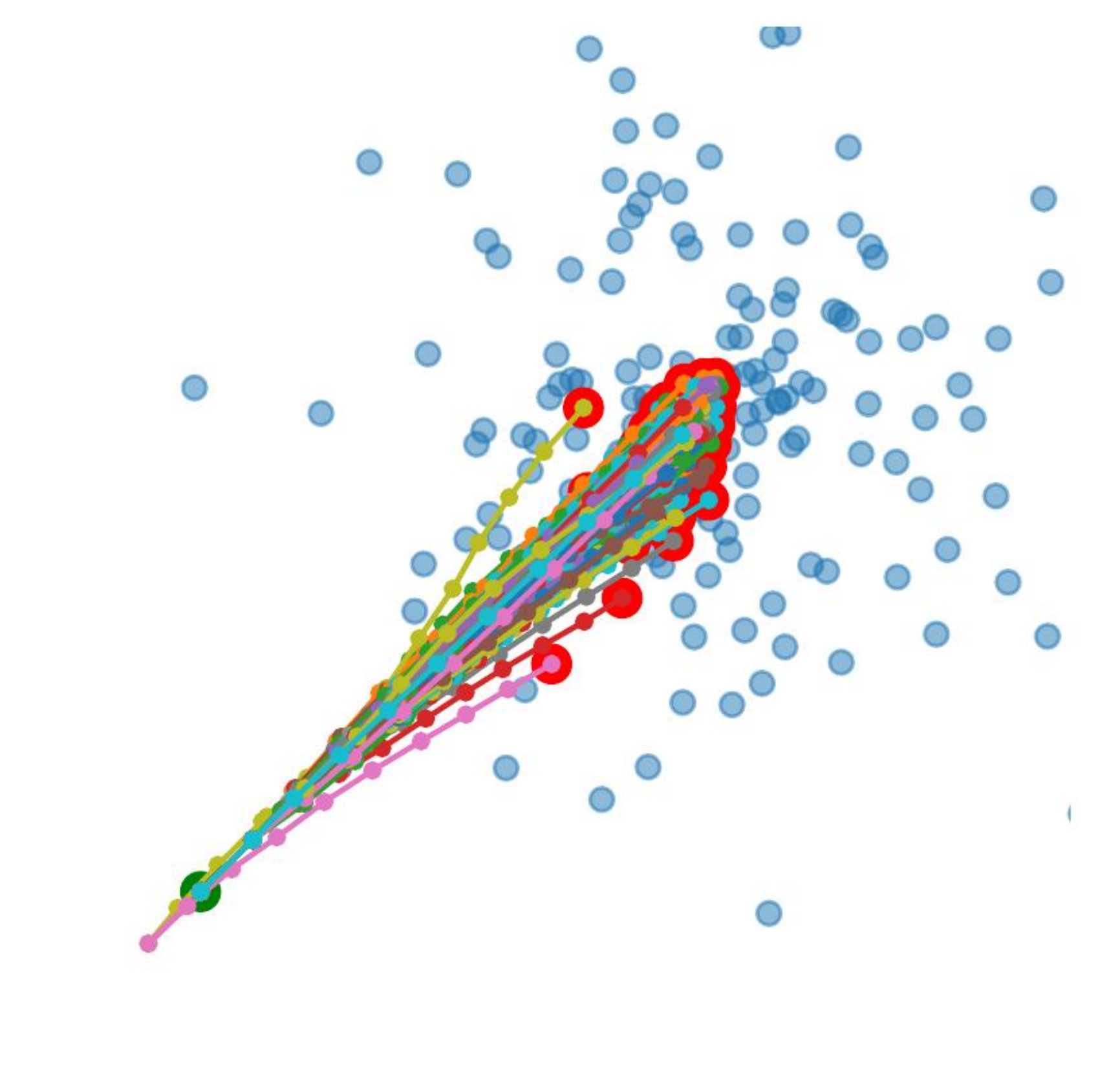}
        \caption{200}
    \end{subfigure}
    \begin{subfigure}[b]{0.09\textwidth}
        \includegraphics[width=\textwidth]{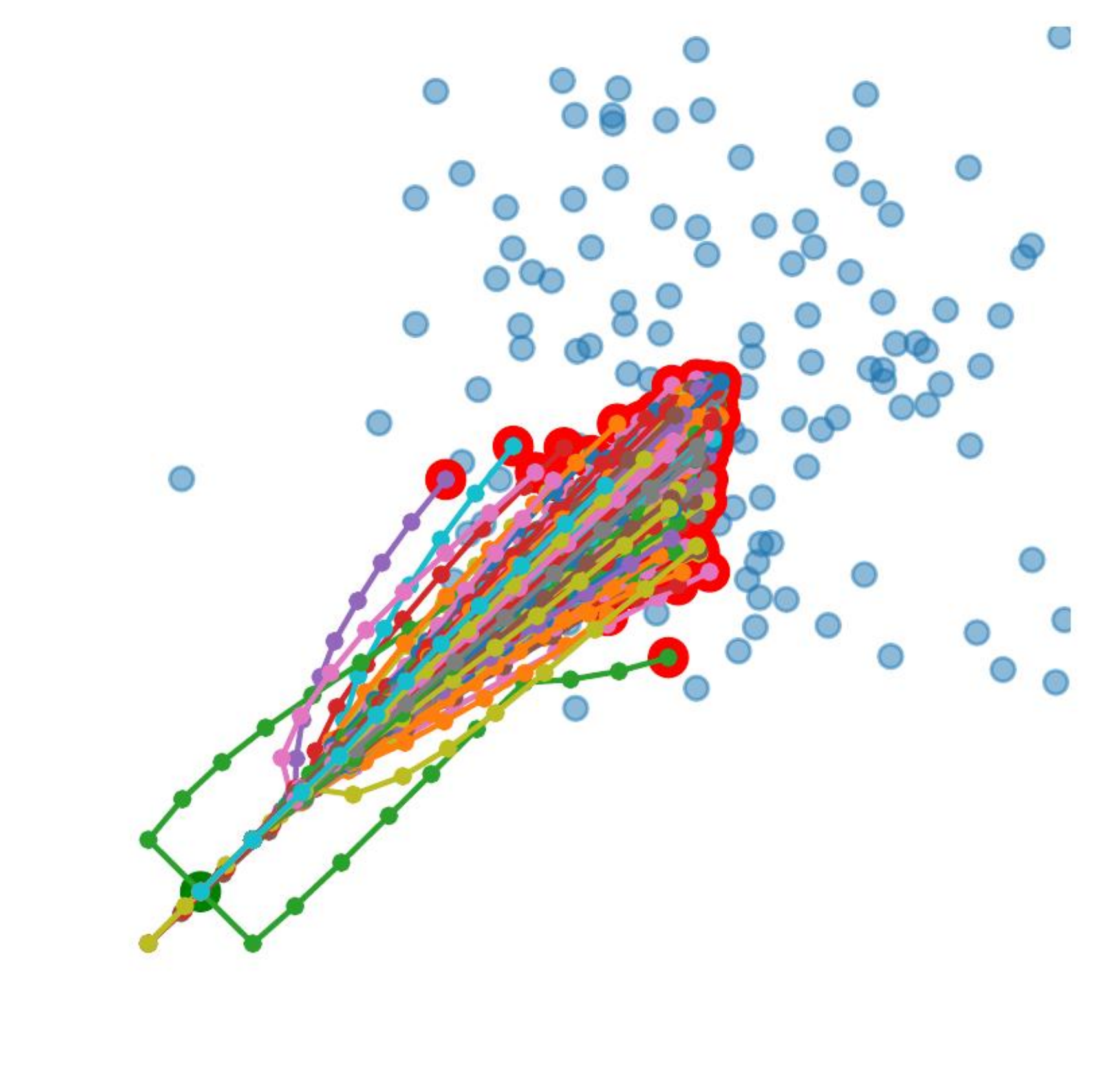}
        \caption{500}
    \end{subfigure}
    \begin{subfigure}[b]{0.09\textwidth}
        \includegraphics[width=\textwidth]{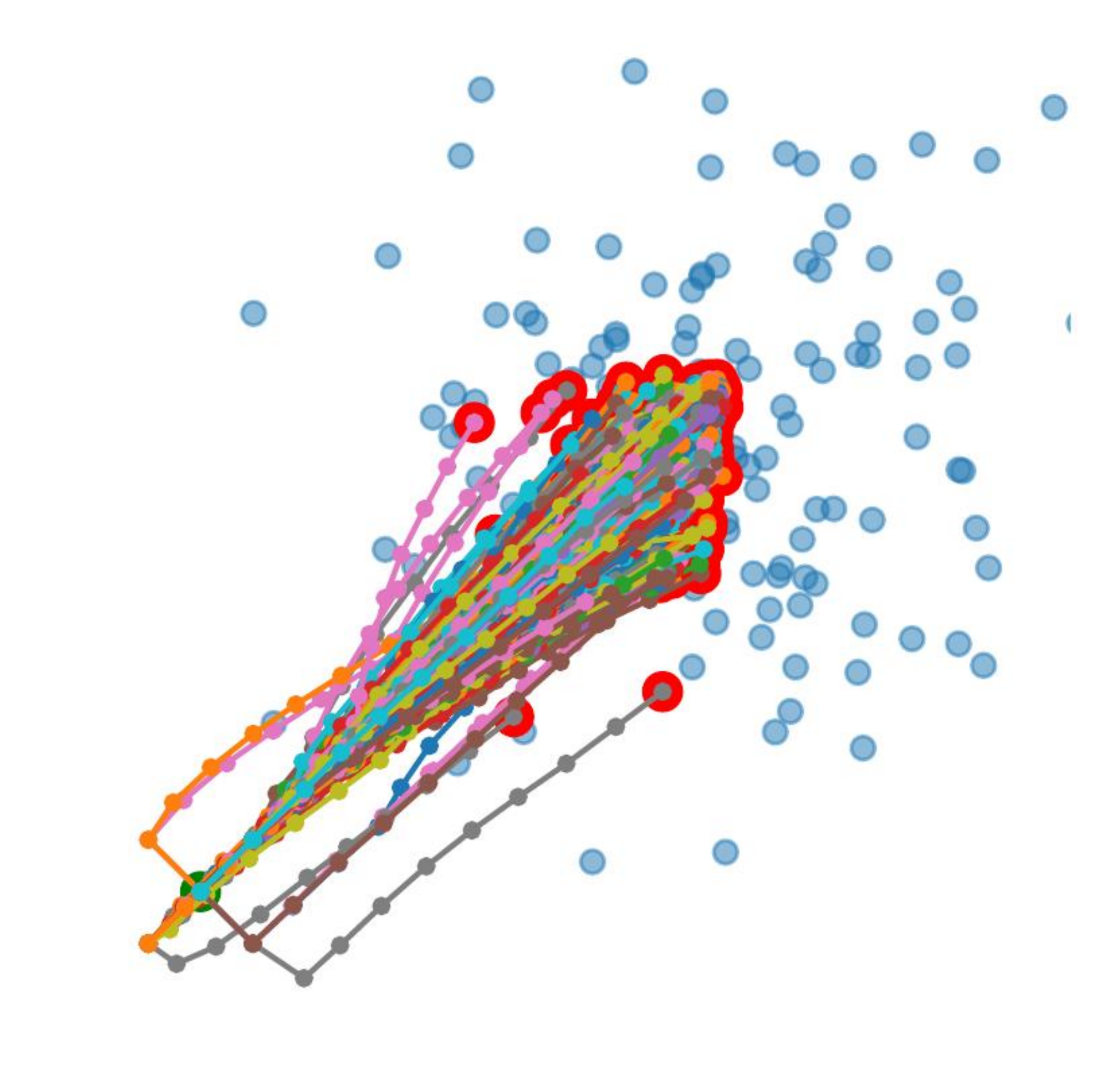}
        \caption{1000}
    \end{subfigure}
    \caption{Initialization point: $(-1, -1)$. Top row: SDS, bottom row: proposed method. {\color{teal}{$\bullet$}}: Starting point. {\color{red}{$\bullet$}}: Ending point.}
    \label{fig:1ddiff-sample-sdl}
\end{figure}

\hd{Optimization convergence.} We evaluated the model's performance in score distillation with different numbers of seeds ($10, 100, 200, 500, 1000$). For all experiments, the learning rate was set to $0.03$, and the optimization was performed using the Adam optimizer with $10$ steps per seed value. The Mean Squared Error (MSE) loss function was used to compute the loss. During optimization, we record the particle position for each iteration. The learning progress of SDS and the proposed method are visualized in Fig.~\ref{fig:1ddiff-sample-stl},~\ref{fig:1ddiff-sample-sdr}, and~\ref{fig:1ddiff-sample-sdl}. 
Each figure illustrates an optimization setting with a different starting point and a destination cluster, respectively.  

\begin{figure}[!ht]
    \centering
    \includegraphics[width=\linewidth]{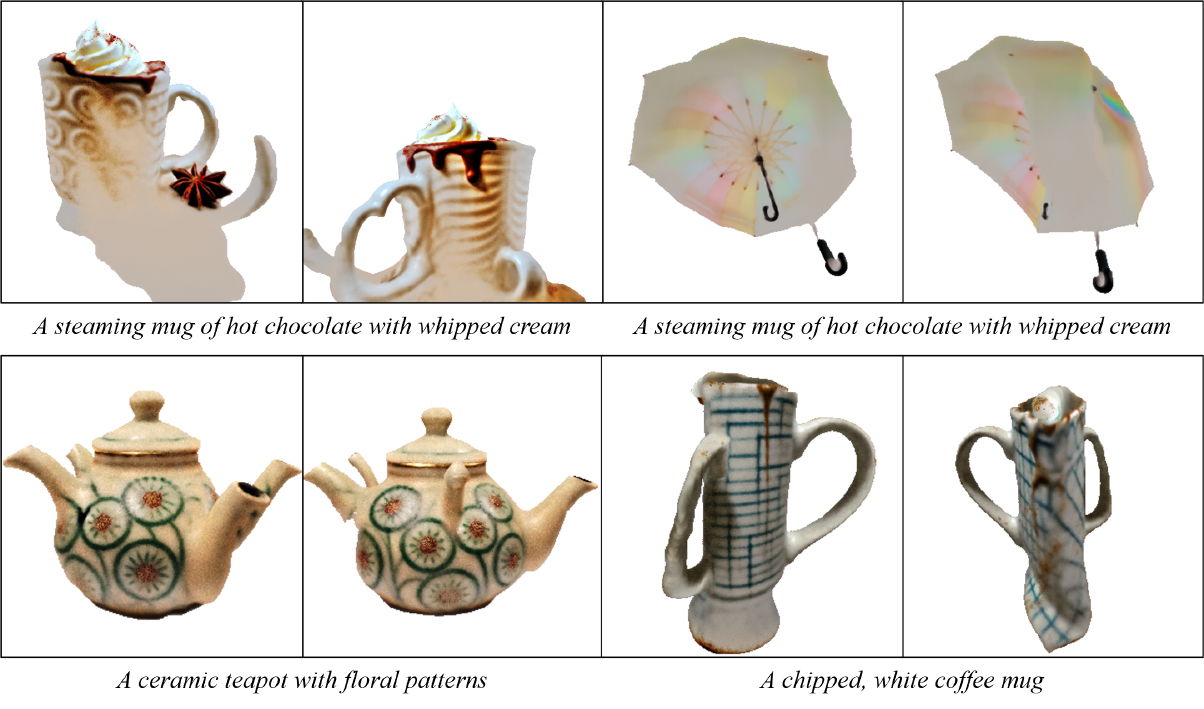}
    \caption{Failed cases.}
    \label{fig:janus}
\end{figure}

\begin{figure*}[!ht]
    \centering
    % Left Side (10 Subfigures)
    \begin{subfigure}[t]{0.49\textwidth}
        \centering
        \begin{adjustbox}{minipage=\linewidth,scale=1}
        \setlength{\tabcolsep}{1pt}
        \begin{tabular}{cc}
            \begin{subfigure}{0.43\linewidth}
                \centering
                \includegraphics[width=\linewidth]{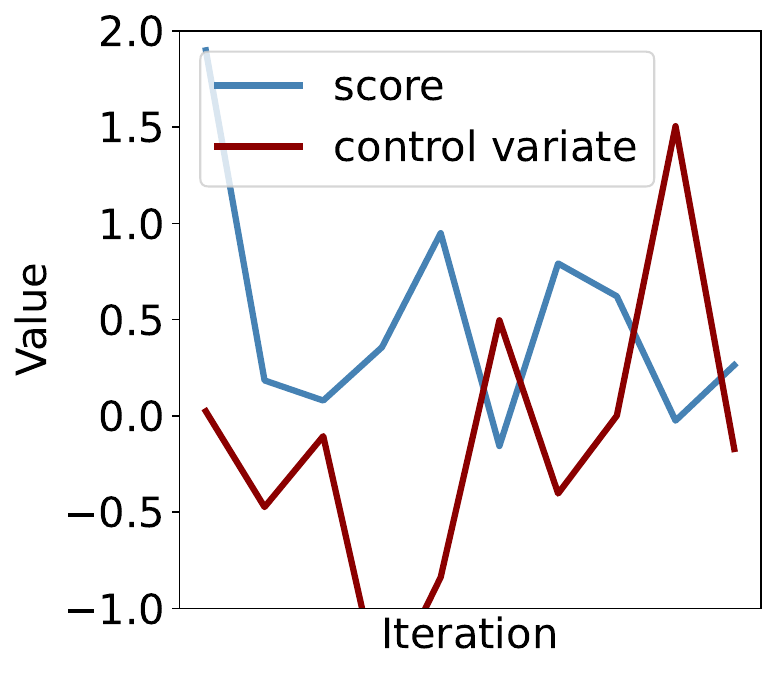}
                \caption{Seed 0}
            \end{subfigure} &
            \begin{subfigure}{0.43\linewidth}
                \centering
                \includegraphics[width=\linewidth]{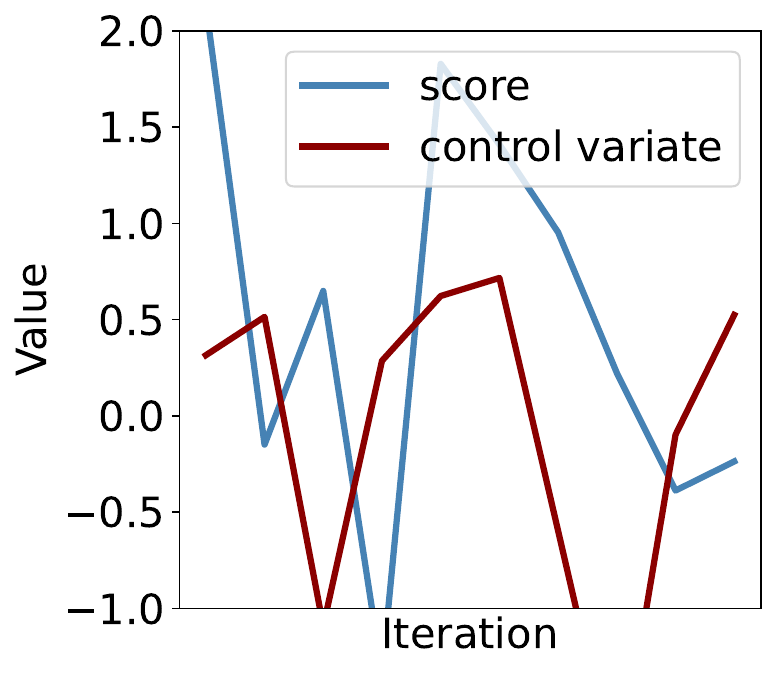}
                \caption{Seed 1}
            \end{subfigure} \\
            \begin{subfigure}{0.43\linewidth}
                \centering
                \includegraphics[width=\linewidth]{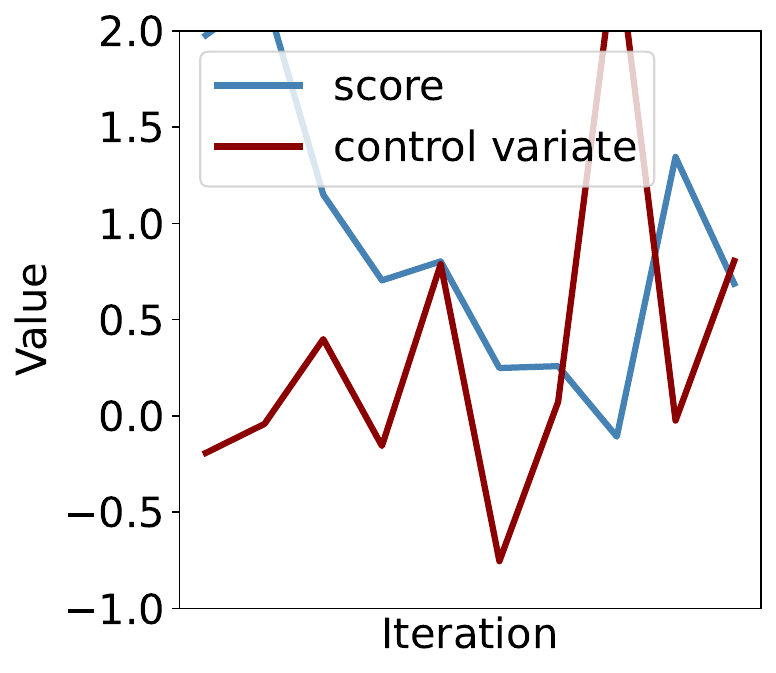}
                \caption{Seed 2}
            \end{subfigure} &
            \begin{subfigure}{0.43\linewidth}
                \centering
                \includegraphics[width=\linewidth]{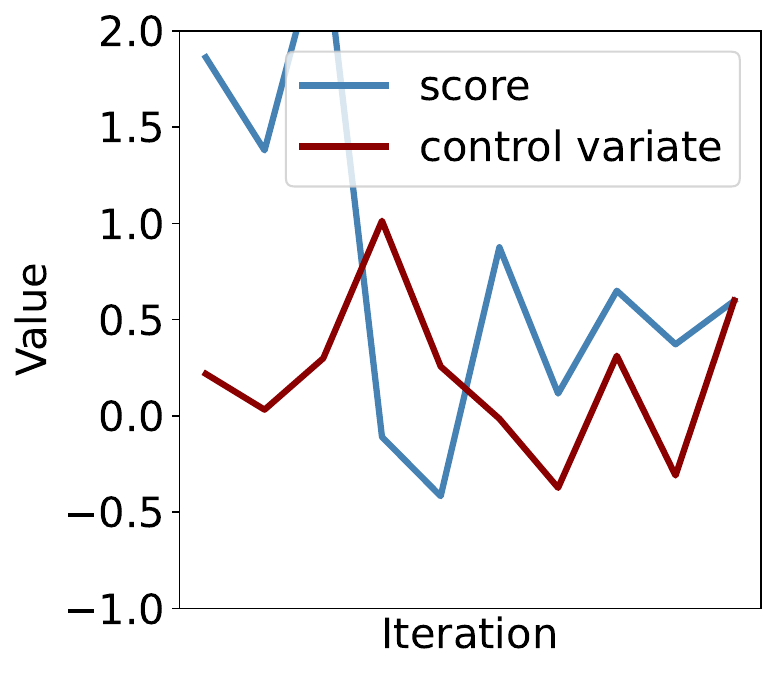}
                \caption{Seed 3}
            \end{subfigure} \\
            \begin{subfigure}{0.43\linewidth}
                \centering
                \includegraphics[width=\linewidth]{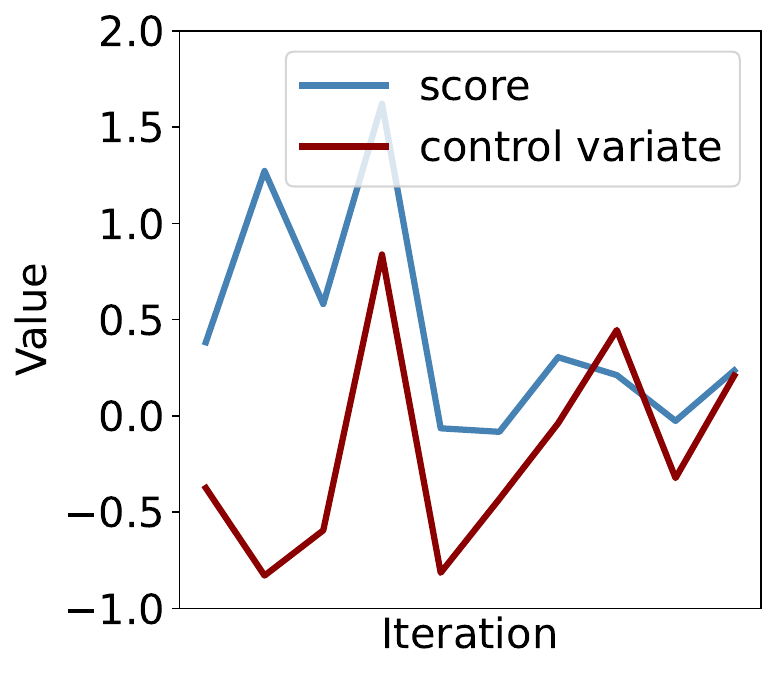}
                \caption{Seed 4}
            \end{subfigure} &
            \begin{subfigure}{0.43\linewidth}
                \centering
                \includegraphics[width=\linewidth]{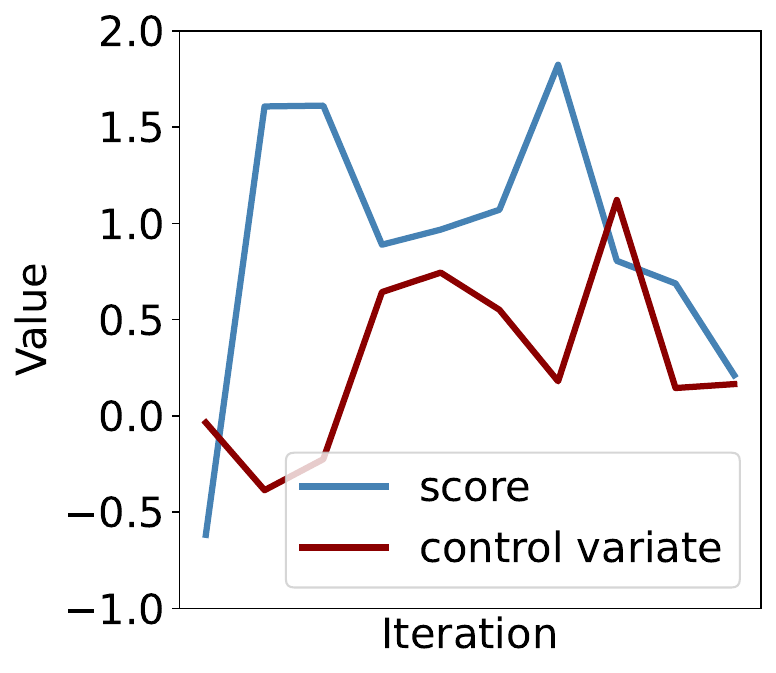}
                \caption{Seed 5}
            \end{subfigure} \\
            \begin{subfigure}{0.43\linewidth}
                \centering
                \includegraphics[width=\linewidth]{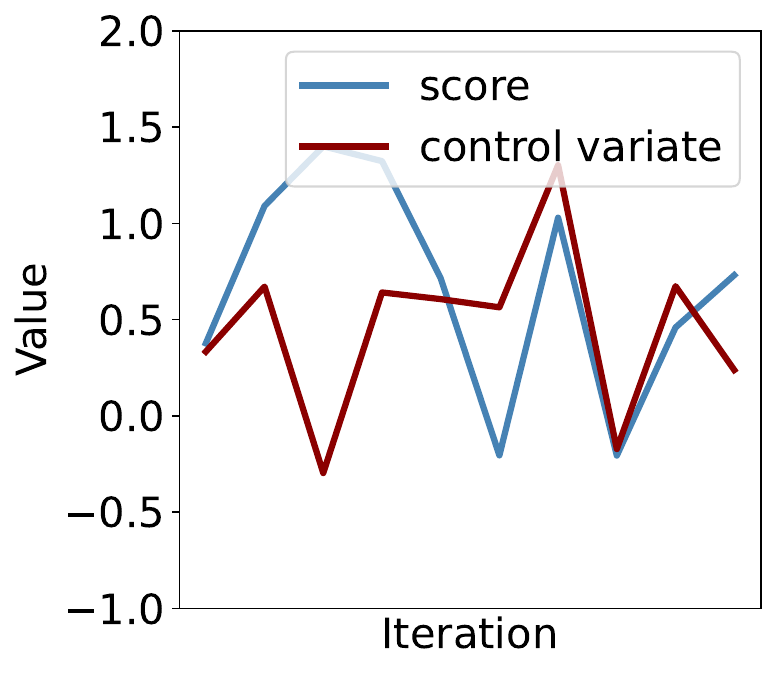}
                \caption{Seed 6}
            \end{subfigure} &
            \begin{subfigure}{0.43\linewidth}
                \centering
                \includegraphics[width=\linewidth]{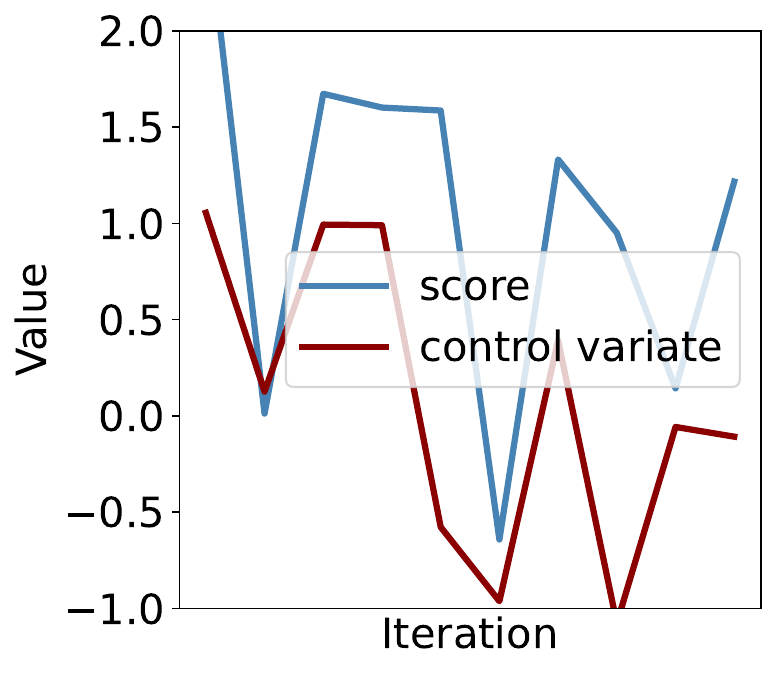}
                \caption{Seed 7}
            \end{subfigure} \\
            \begin{subfigure}{0.43\linewidth}
                \centering
                \includegraphics[width=\linewidth]{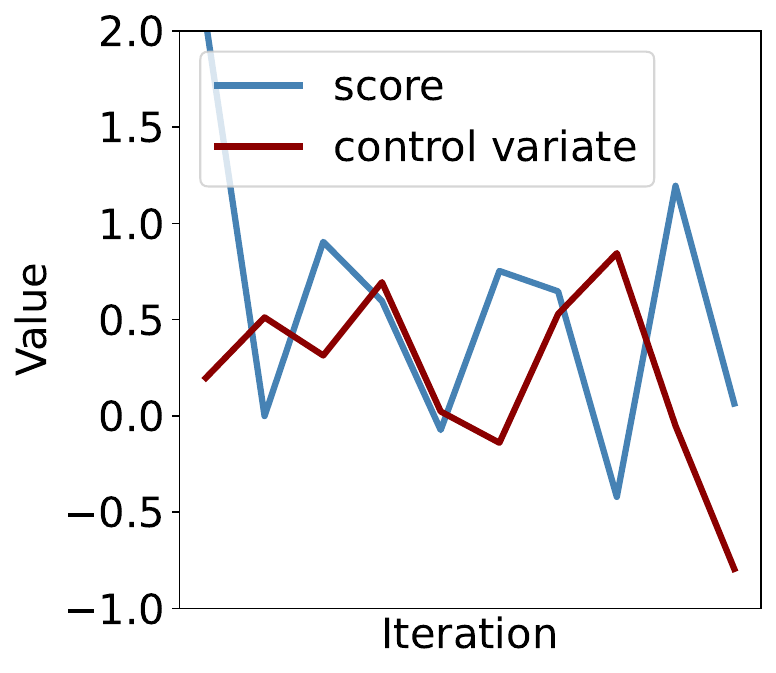}
                \caption{Seed 8}
            \end{subfigure} &
            \begin{subfigure}{0.43\linewidth}
                \centering
                \includegraphics[width=\linewidth]{imgs/analysis/cv-ns/jsd_ns-ns_9.pdf}
                \caption{Seed 9}
            \end{subfigure}
        \end{tabular}
        \end{adjustbox}
        \centerline{JSD}
    \end{subfigure}
    % Right Side (10 Subfigures)
    \begin{subfigure}[t]{0.49\textwidth}
        \centering
        \begin{adjustbox}{minipage=\linewidth,scale=1}
        \setlength{\tabcolsep}{1pt}
        \begin{tabular}{cc}
            \begin{subfigure}{0.40\linewidth}
                \centering
                \includegraphics[width=\linewidth]{imgs/analysis/cv-ns/sds_ns-ns_0.pdf}
                \caption{Seed 0}
            \end{subfigure} &
            \begin{subfigure}{0.40\linewidth}
                \centering
                \includegraphics[width=\linewidth]{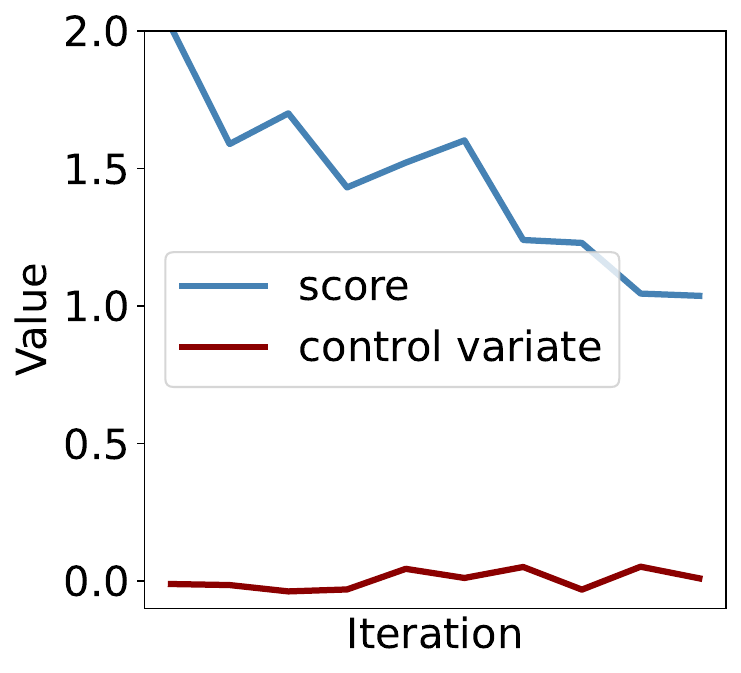}
                \caption{Seed 1}
            \end{subfigure} \\
            \begin{subfigure}{0.40\linewidth}
                \centering
                \includegraphics[width=\linewidth]{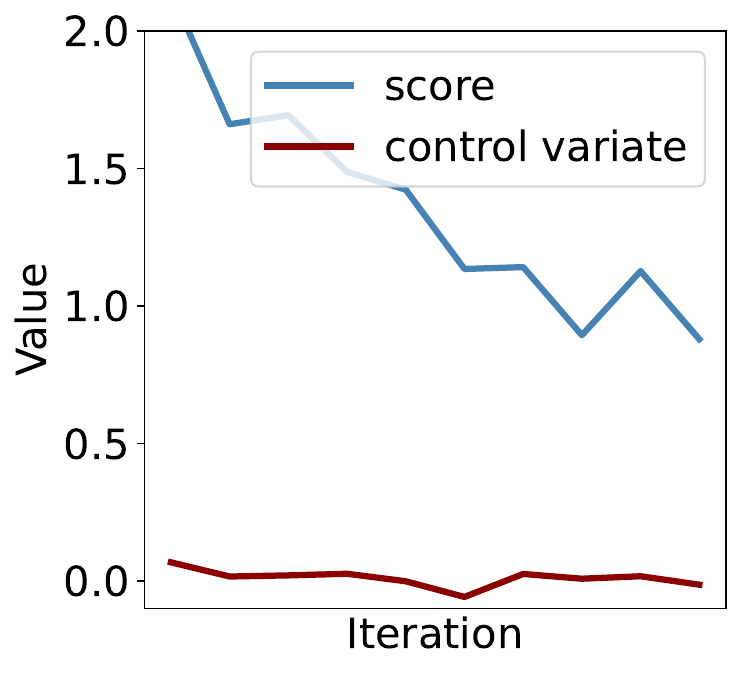}
                \caption{Seed 2}
            \end{subfigure} &
            \begin{subfigure}{0.40\linewidth}
                \centering
                \includegraphics[width=\linewidth]{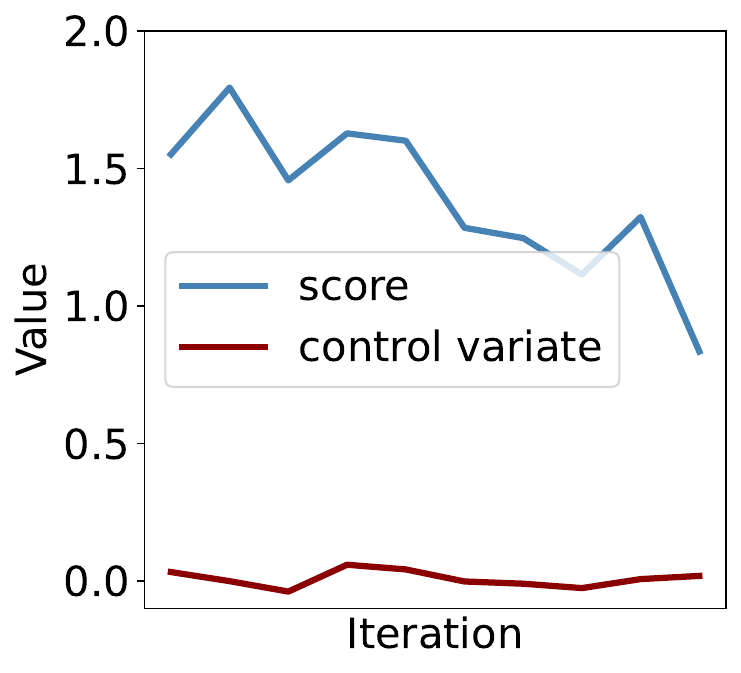}
                \caption{Seed 3}
            \end{subfigure} \\
            \begin{subfigure}{0.40\linewidth}
                \centering
                \includegraphics[width=\linewidth]{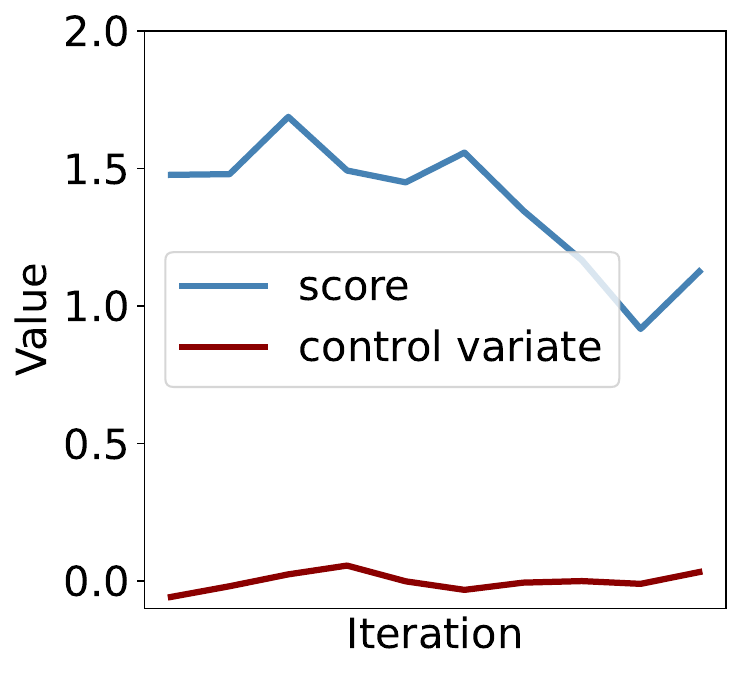}
                \caption{Seed 4}
            \end{subfigure} &
            \begin{subfigure}{0.40\linewidth}
                \centering
                \includegraphics[width=\linewidth]{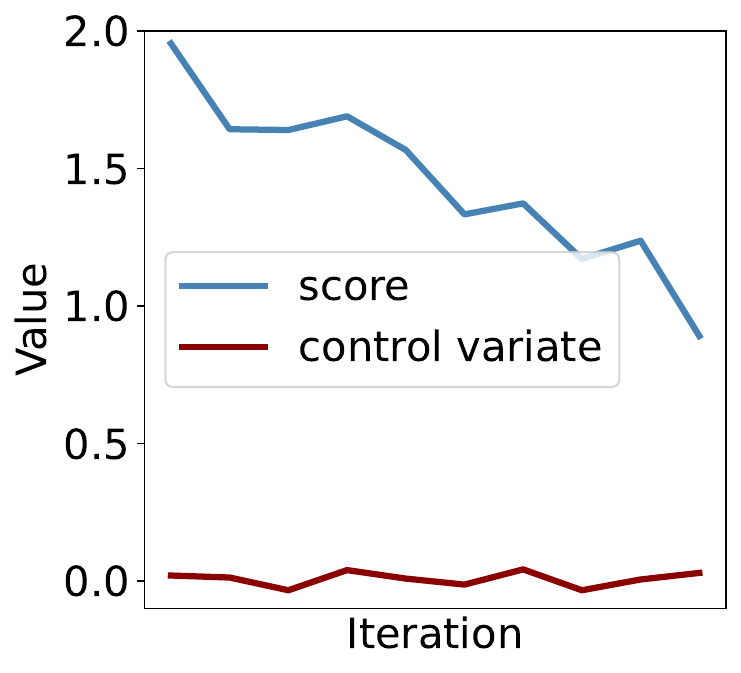}
                \caption{Seed 5}
            \end{subfigure} \\
            \begin{subfigure}{0.40\linewidth}
                \centering
                \includegraphics[width=\linewidth]{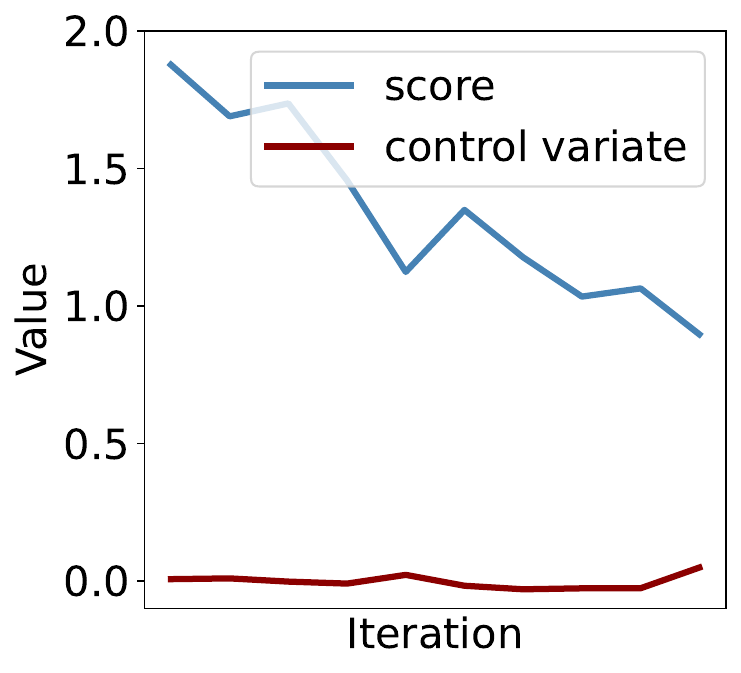}
                \caption{Seed 6}
            \end{subfigure} &
            \begin{subfigure}{0.40\linewidth}
                \centering
                \includegraphics[width=\linewidth]{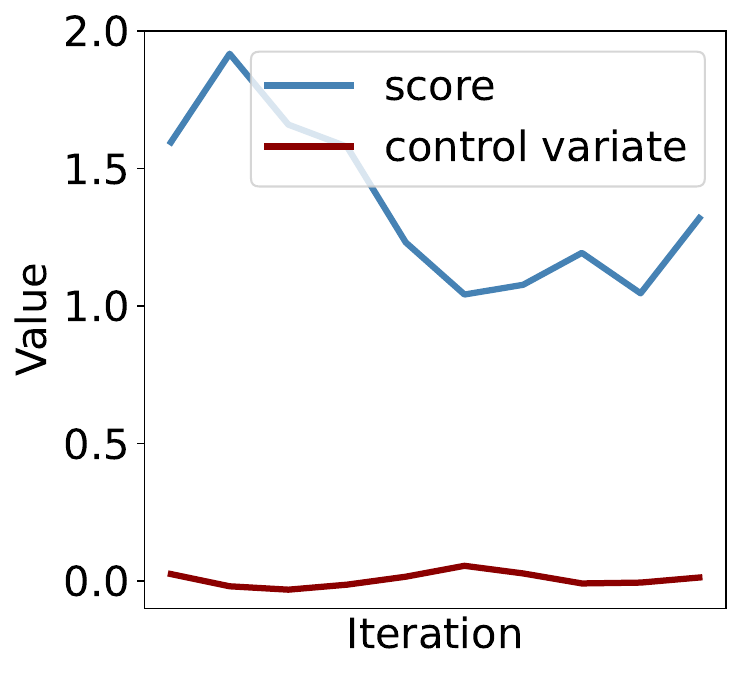}
                \caption{Seed 7}
            \end{subfigure} \\
            \begin{subfigure}{0.40\linewidth}
                \centering
                \includegraphics[width=\linewidth]{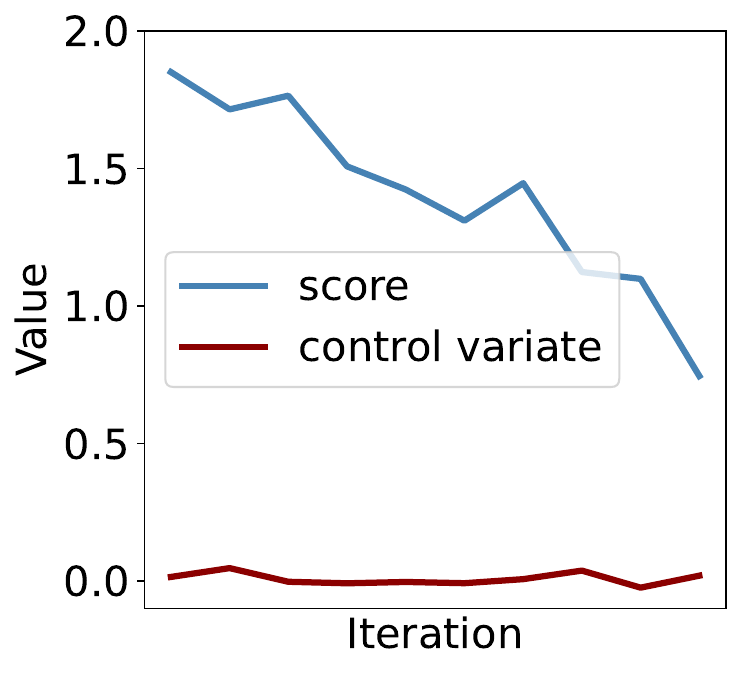}
                \caption{Seed 8}
            \end{subfigure} &
            \begin{subfigure}{0.40\linewidth}
                \centering
                \includegraphics[width=\linewidth]{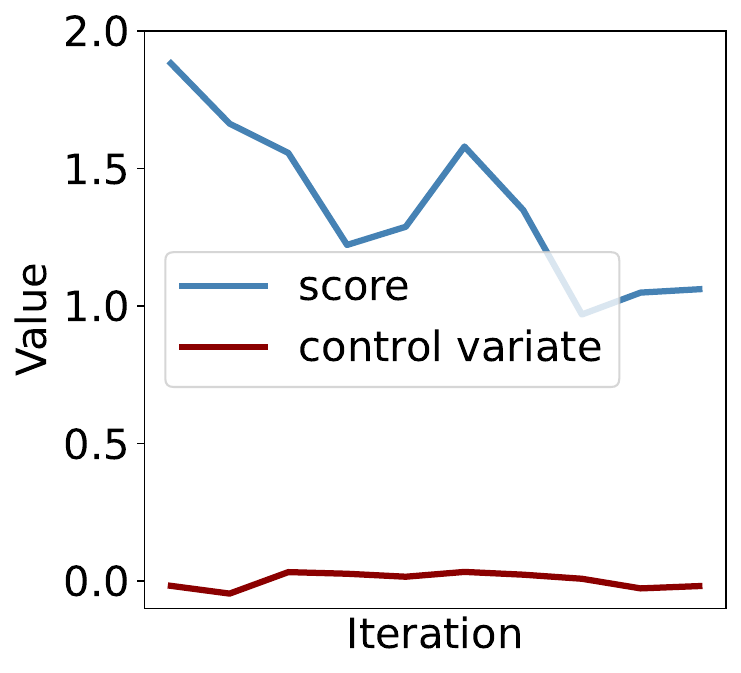}
                \caption{Seed 9}
            \end{subfigure}
        \end{tabular}
        \end{adjustbox}
        \centerline{SDS}
    \end{subfigure}
    \caption{Comparison between the estimated noise and the control variate in our proposed method (left) and SDS (right).}
    \label{fig:apx-estimated-noise-vs-control-variate}
\end{figure*}

\hd{Correlation between estimated noise and control variate.}
In this experiment, we compare the estimated noise and control variate in both SDS and our proposed method. The obtained results show that our estimated noise and the control variate are highly correlated. This confirms that the gradients become more stable than SDS, thus improving the convergence. The results of the first 10 seeds are shown in Fig.~\ref{fig:apx-estimated-noise-vs-control-variate}.

\section{More Qualitative Results}
\label{apx:more_qualitative}

\subsection{Failed Cases}
Like other text-to-3D methods, our method can sometimes produce multi-faced models. We show more examples of the Janus problem and other anomaly cases in Fig.~\ref{fig:janus}. 

\subsection{More Qualitative Results with Neural Radiance Field}
\begin{figure*}
    \centering
    \includegraphics[width=0.83\textwidth]{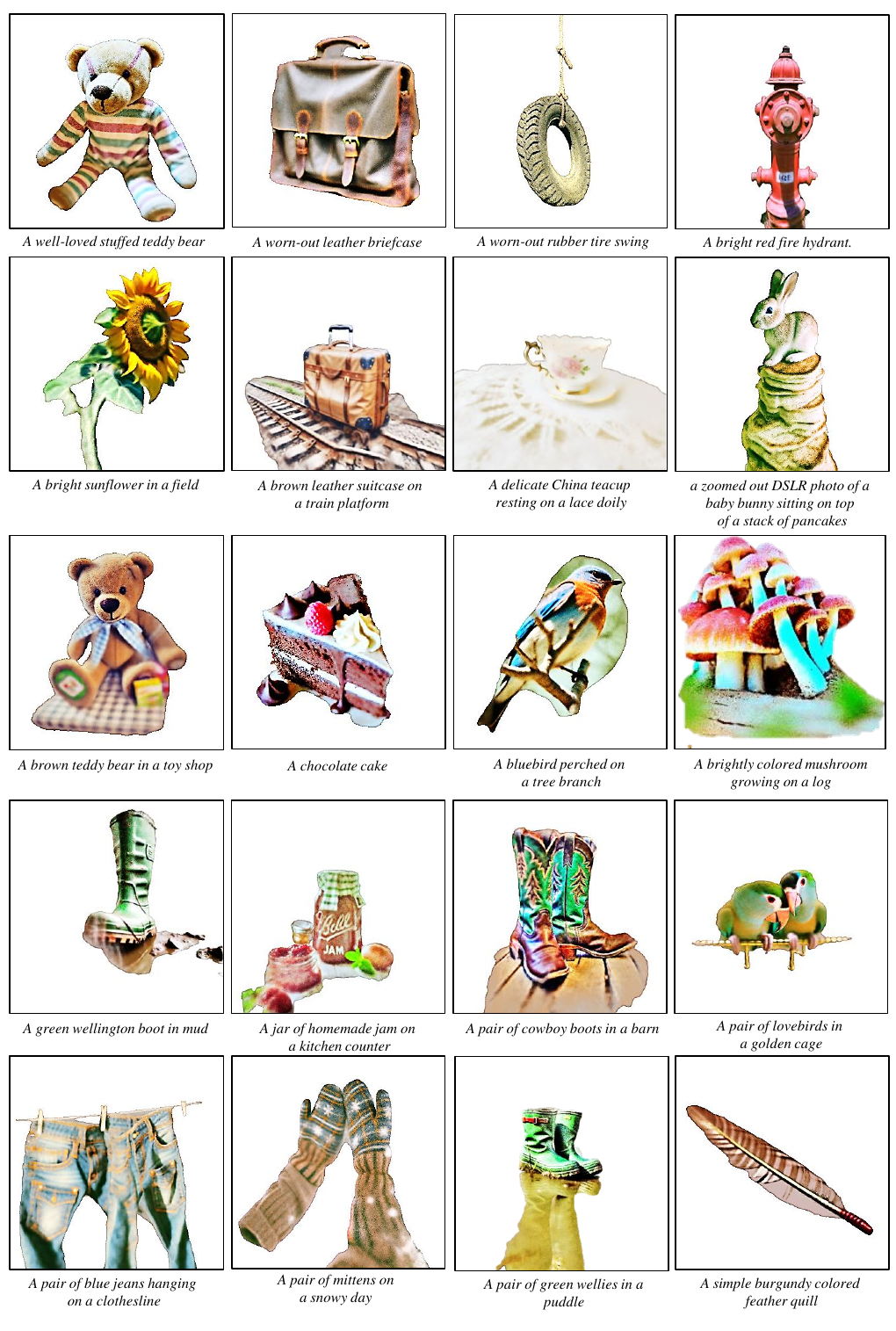}
    \caption{Additional qualitative results (1/3)}
    \label{fig:qualitative1}
\end{figure*}

\begin{figure*}
    \centering
    \includegraphics[width=0.83\textwidth]{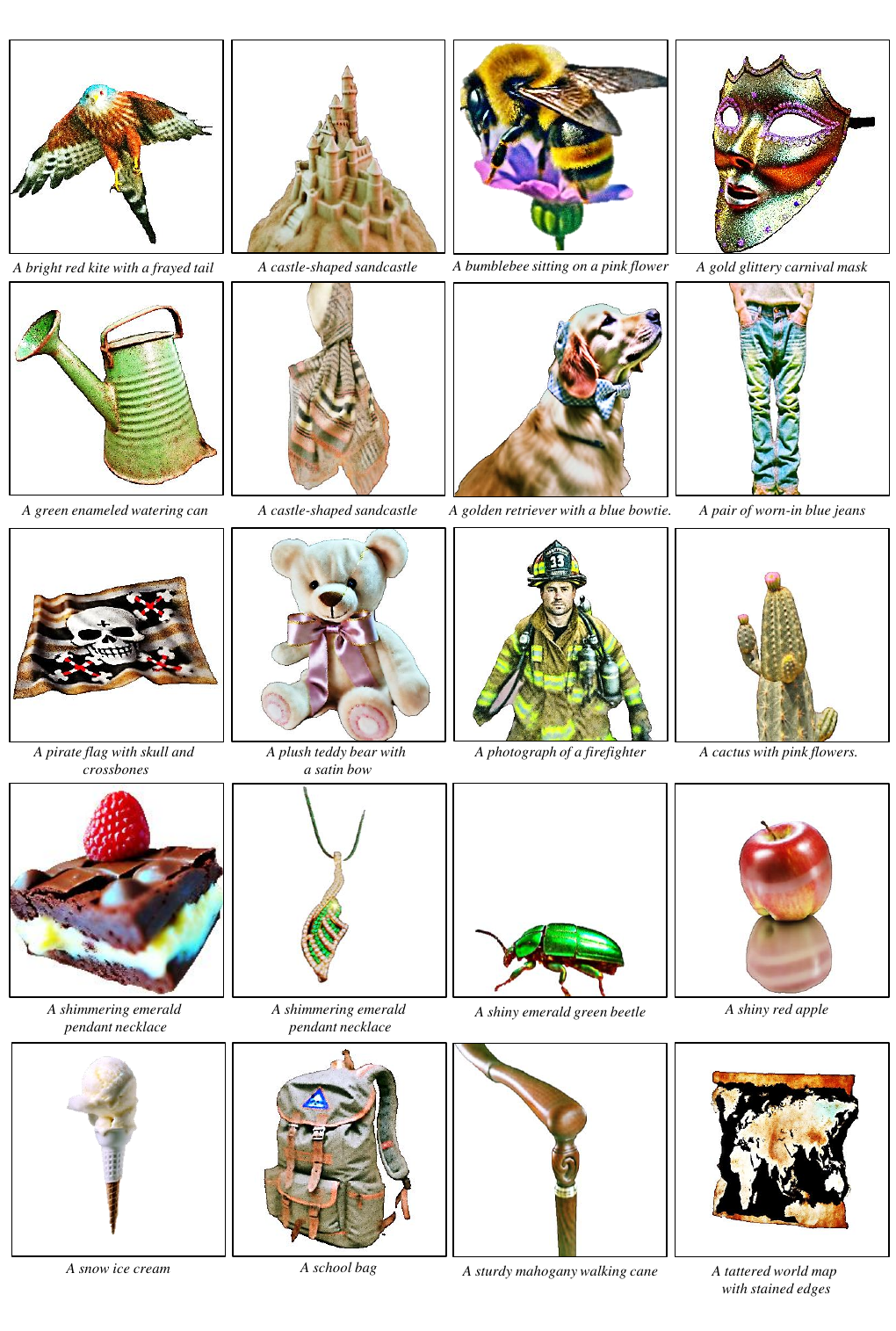}
    \caption{Additional qualitative results (2/3)}
    \label{fig:qualitative2}
\end{figure*}

\begin{figure*}
    \centering
    \includegraphics[width=0.83\textwidth]{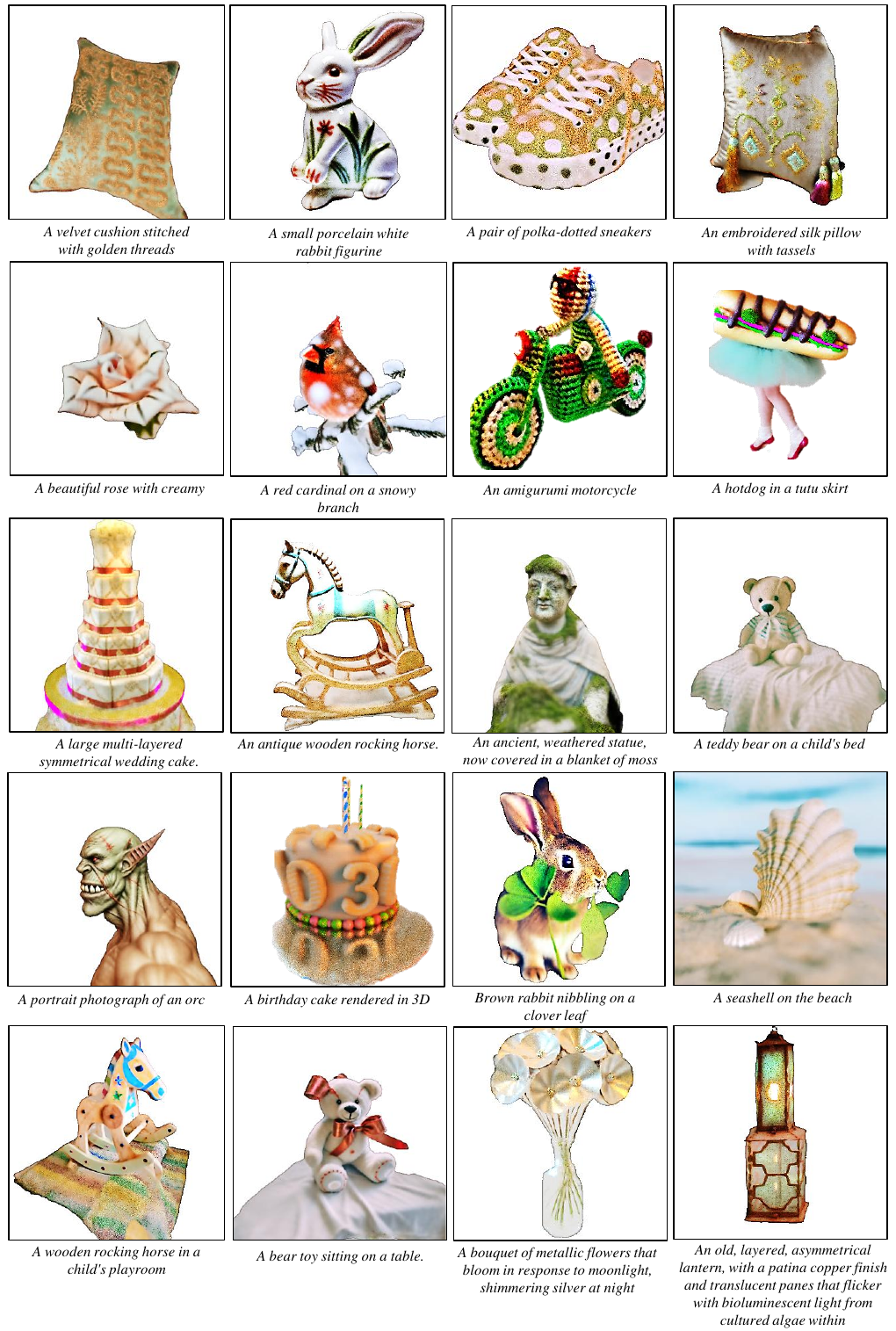}
    \caption{Additional qualitative results (3/3)}
    \label{fig:qualitative3}
\end{figure*}

In Figs.~\ref{fig:qualitative1},~\ref{fig:qualitative2}, and ~\ref{fig:qualitative3}, we present the visual results of the generated objects using NeRF from various text prompts. 

\subsection{More Qualitative Comparison with SOTAs}

\begin{figure*}
    \centering
    \includegraphics[width=0.65\linewidth]{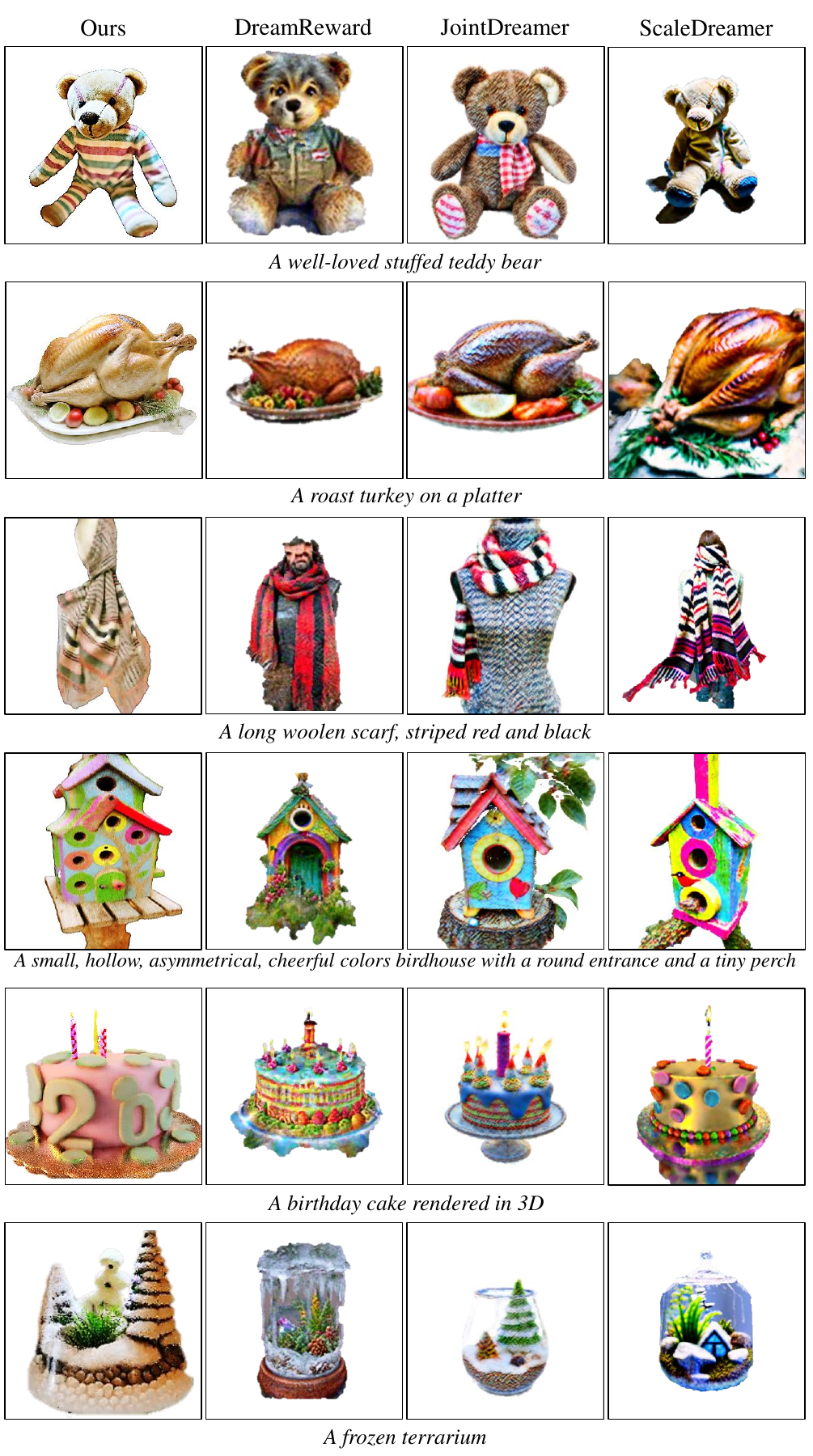}
    \caption{Additional qualitative comparisons (1/4)}
    \label{fig:compare-apx-0}
\end{figure*}

\begin{figure*}
    \centering
    \includegraphics[width=0.65\linewidth]{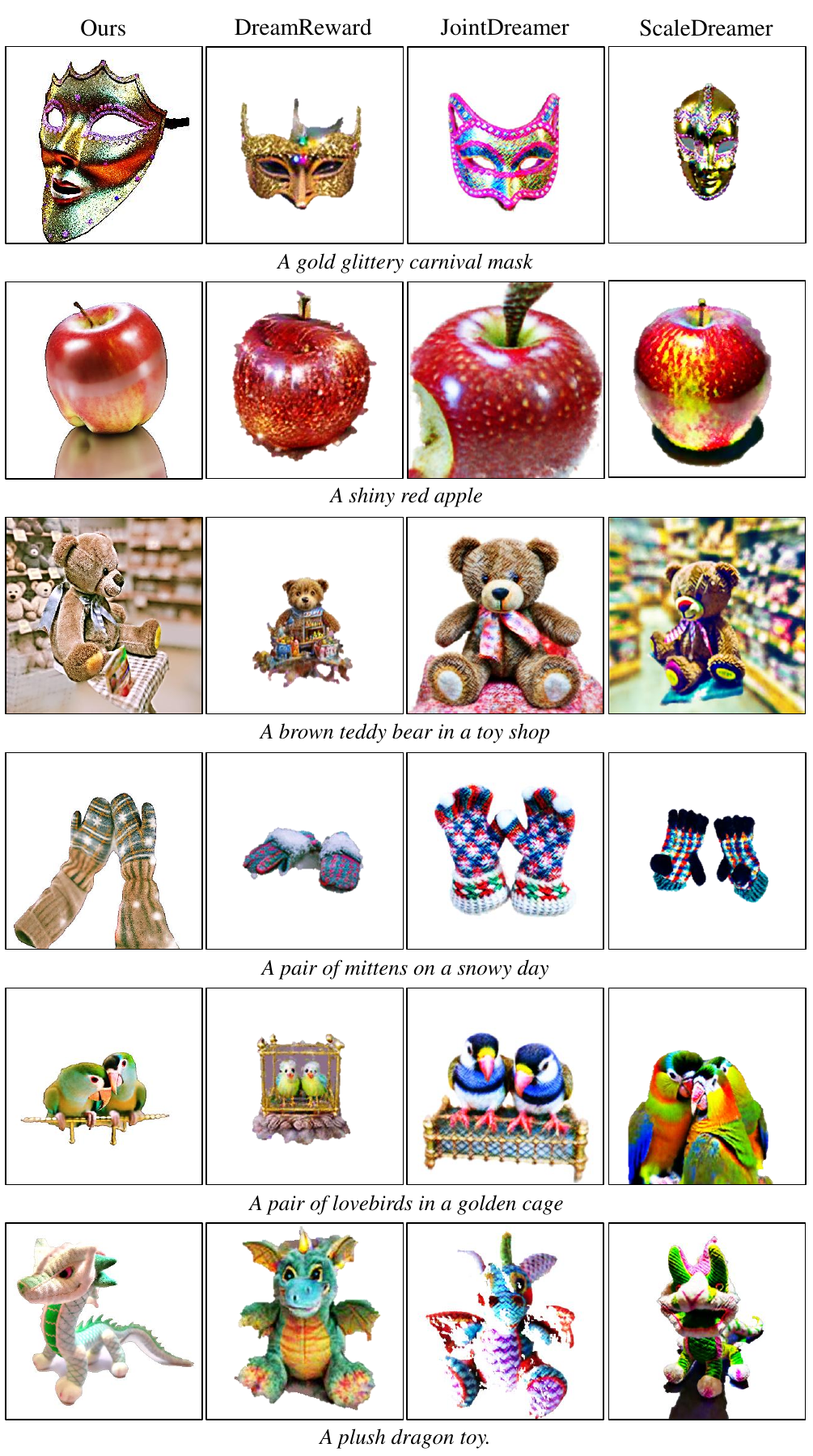}
    \caption{Additional qualitative comparisons (2/4)}
    \label{fig:compare-apx-1}
\end{figure*}

\begin{figure*}
    \centering
    \includegraphics[width=0.65\linewidth]{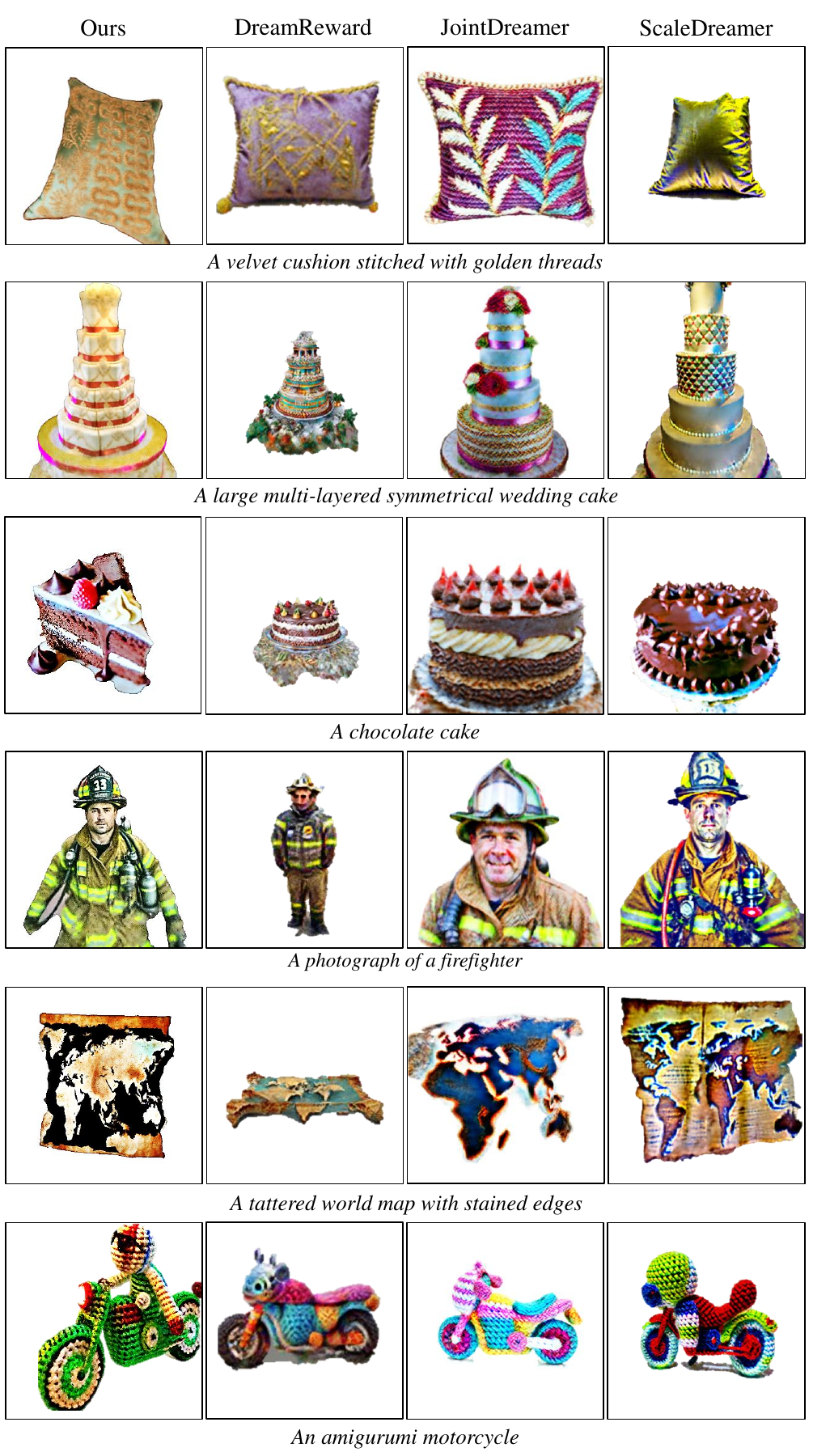}
    \caption{Additional qualitative comparisons (3/4)}
    \label{fig:compare-apx-2}
\end{figure*}

\begin{figure*}
    \centering
    \includegraphics[width=0.65\linewidth]{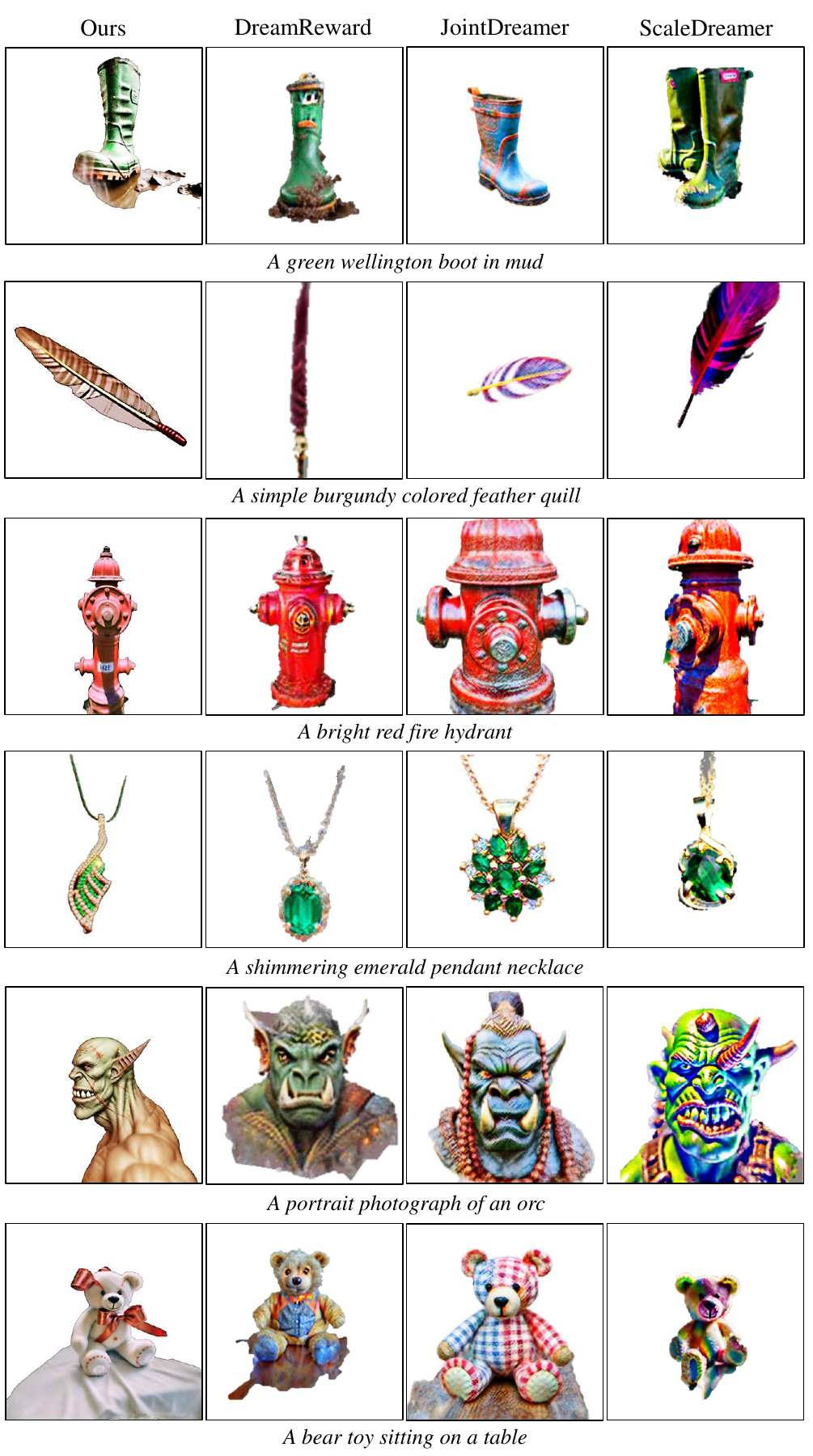}
    \caption{Additional qualitative comparisons (4/4)}
    \label{fig:compare-apx-3}
\end{figure*}

In Fig.~\ref{fig:compare-apx-0},~\ref{fig:compare-apx-1}, ~\ref{fig:compare-apx-2}, and~\ref{fig:compare-apx-3}, we provide external qualitative comparisons between our approach and the other SOTAs.

\clearpage
\onecolumn
\section{Addtional Experimental Settings}

\subsection{Training Settings}

\textbf{Camera viewpoints} are sampled within a distance range of 1.0 to 1.5 units and field‑of‑view angles between 40° and 70°, ensuring diverse spatial coverage. Elevation angles vary from –10° to 45°, allowing both low‑angle and overhead perspectives to be captured. To maintain geometric stability during training, we do not use camera perturbation. Illumination sampling uses the DreamFusion~\cite{sds} strategy. During evaluation, we fix the camera distance at 1.5 units and the field‑of‑view at 70° to standardize quantitative comparisons across models.

\noindent\textbf{Geometry Representation.} We employ an \textit{implicit-volume} representation for geometry, setting the scene radius to 2.0 units and utilizing analytic normals to ensure precise surface orientation. For density initialization. To encode spatial positions, we implement a coarse-to-fine multiresolution hash grid encoding. This configuration comprises 16 levels, each with 2 features per level. The base resolution is set to 16. A per-level scale factor of approximately 1.447 is applied, enabling the encoding to capture both global structures and fine-grained details effectively.

\noindent\textbf{Optimizer Configuration.} We utilize the Adam optimizer with a base learning rate of 0.01, employing $\beta_1 = 0.9$ and $\beta_2 = 0.99$ for the first and second moment estimates, respectively. A small epsilon value of $1 \times 10^{-15}$ is used to enhance numerical stability during training. To facilitate fine-grained control over different components of the model, we assign distinct learning rates: the geometry parameters are updated with a learning rate of 0.01, while the background parameters use a reduced learning rate of 0.001. Training is conducted for a total of 10,000 steps. To improve computational efficiency and reduce memory usage, we adopt mixed-precision training with 16-bit floating-point precision.

\subsection{List of prompts}

In this section, we provided the list of prompts which were used in the paper, including 300 prompts from T3Bench~\cite{t3bench}, 110 prompts from GPTEval3D~\cite{gpteval3d}, and 70 prompts from DiverseDream~\cite{diversedream}.

\subsubsection{T3Bench}

\textbf{Single Object prompts}
\begin{verbatim}
A cactus with pink flowers
A rainbow-colored umbrella
An antique wooden rocking horse
A golden retriever with a blue bowtie
An ivory candlestick holder
A pair of polka-dotted sneakers
A steaming mug of hot chocolate with whipped cream
A bright red fire hydrant
A gleaming silver saxophone
A leather-bound book with gold details
A vibrant sunflower with green leaves
A castle-shaped sandcastle
A neon green skateboard with black wheels
A pirate flag with skull and crossbones
A plush teddy bear with a satin bow
A ripe watermelon sliced in half
A sparkling diamond ring in a velvet box
A vintage porcelain doll with a frilly dress
A chameleon perched on a tree branch
A tarnished brass pocket watch
A ceramic teapot with floral patterns
An antique ruby-studded brooch
A simple burgundy colored feather quill
A vintage iron-cast typewriter
A shiny emerald green beetle
A crystal glass paperweight with abstract design
A velvet cushion stitched with golden threads
A small porcelain white rabbit figurine
A left-handed electric guitar painted black
A bright blue plastic swimming goggles
A partly broken shell of a tortoise
A long woolen scarf, striped red and black
A tattered old explorer’s map
A well-used black iron frying pan
A crumpled silver aluminum soda can
A thick, green-spined book with yellowed pages
A shimmering emerald pendant necklace
A well-worn straw sun hat
A tarnished silver letter opener
An antique glass perfume bottle
A polished mahogany grand piano
A dented brass trumpet
A pristine white wedding gown
A chipped porcelain teacup
A rustic wrought-iron candle holder
A vibrant, handmade patchwork quilt
A plush velvet armchair
A sleek, black top hat
A paint-splattered easel
A bent steel crowbar
A crisp paper airplane
A worn-out rubber tire swing
An intricately-carved wooden chess set
A bright red kite with a frayed tail
A smooth, round opal stone
A rusty, vintage metal key
A delicate, handmade lace doily
A sturdy mahogany walking cane
A sparkling crystal chandelier
A worn-out red flannel shirt
A cracked porcelain doll's face
A dusty classic typewriter
A glossy grand black piano
A faux-fur leopard print hat
A futuristic, sleek electric car model
A cherry red vintage lipstick tube
A cobweb-covered old wooden chest
A gold glittery carnival mask
A tattered world map with stained edges
A shiny red apple
A worn-out leather briefcase
An antique gold pocket watch
A sleek, slim smartphone
A wet, vibrant beach ball
A rusty, abandoned bicycle
A fluffy, orange cat
Crisp, folded origami paper
A shiny, new electric guitar
A weather-beaten wooden bat
A delicate crystal champagne flute
An old, frayed straw hat
A scuffed up soccer ball
A pair of worn-in blue jeans
A well-loved stuffed teddy bear
A chipped, white coffee mug
A bright, yellow rubber duck
A sleek stainless steel teapot
A water-streaked glass window pane
An intricate ceramic vase with peonies painted on it
A fuzzy pink flamingo lawn ornament
A blooming potted orchid with purple flowers
An old bronze ship's wheel
A sparkling diamond tiara
A vintage plaid woolen blanket
A pair of shiny black patent leather shoes
An elegant feather-quill ink pen
A fragrant pine Christmas wreath
A silver mirror with ornate detailing
A green enameled watering can
A classic leatherette radio with dials
\end{verbatim}

\textbf{Single Object with Surroundings}

\begin{verbatim}
A white seashell on a sandy beach
A vintage clock hanging on a brick wall
A pair of glasses on an open notebook
A green cactus in a clay pot
A colorful kite flying in a clear sky
A wooden rocking chair on a porch
A silver spoon in a bowl of soup
A black cat sleeping on a windowsill
A bluebird perched on a tree branch
A pair of sneakers hanging from a power line
A yellow umbrella on a rainy day
A pink bicycle leaning against a fence
A snowman wearing a scarf in a winter landscape
A leather-bound book on a wooden table
A red rose in a crystal vase
A jar of homemade jam on a kitchen counter
A pair of ballet shoes on a dance floor
A green turtle swimming in a clear pond
A rainbow-colored kite soaring in the sky
A chocolate cake on a white plate
A pair of binoculars on a mountain peak
A teddy bear on a child's bed
A pair of blue jeans hanging on a clothesline
A bright sunflower in a field
A colorful parrot on a jungle tree
A lighthouse on a rocky shore
A pair of hiking boots on a trail
A red and white lighthouse on a cliff
A yellow school bus on a city street
A pair of lovebirds in a golden cage
A white sailboat on a calm sea
A vintage typewriter on a desk
A ripe watermelon on a picnic table
A stack of pancakes on a breakfast table
A green frog on a lily pad
A pair of mittens on a snowy day
A blue butterfly on a pink flower
A red fire hydrant on a sidewalk
A pair of cowboy boots in a barn
A black and white photo in a silver frame
A rainbow over a waterfall
A white dove flying in a blue sky
A pink piggy bank on a shelf
A pair of scissors on a craft table
A red barn in a green field
A pair of flip flops on a beach
A hot air balloon in a clear sky
A green apple on a teacher's desk
A yellow rubber duck in a bathtub
A white swan on a tranquil lake
A red cardinal on a snowy branch
A black and white soccer ball on a field
A silver teapot on a dining table
A red and white candy cane on a Christmas tree
A pair of red boxing gloves hanging on a wall
A white wedding dress on a hanger
A brown leather suitcase on a train platform
A pink flamingo in a zoo
A pair of green wellies in a puddle
A black grand piano in a concert hall
A white picket fence around a garden
A blue whale in the deep ocean
A brown horse in a green pasture
A brown teddy bear in a toy shop
A brown leather suitcase at an airport
A green wellington boot in mud
A blue and white china cup on a saucer
A silver laptop sitting on a wooden desk
A blue ceramic mug filled with steaming coffee
A vintage pocket watch with a golden chain
A green leather-bound book on an antique shelf
A pair of white sneakers on a black mat
A golden locket with a delicate chain
A white porcelain teapot on a lace tablecloth
A silver telescope pointing towards the starry sky
A purple umbrella left on a park bench
A wooden rocking horse in a child's playroom
A red apple resting on a white ceramic plate
A green paperback novel lying on a beige couch
A white cotton t-shirt hanging on a wooden hanger
A silver wristwatch ticking on a glass bedside table
A pink ceramic vase filled with fresh white lilies
A gold necklace with a diamond pendant displayed in a velvet box
A black and white photograph framed in dark mahogany
A pair of red high-heeled shoes sitting in a shoebox
A blue denim jacket draped over a wooden chair
A purple yoga mat rolled up in a gym bag
A silver toaster with two slices of bread browning
A red ceramic coffee mug sitting on a wooden table
A green plastic watering can filled with fresh water
A shiny gold pocket watch ticking away in a velvet box
A blue denim jacket hanging on a metal coat rack
A pair of brown leather boots standing at the front door
A crystal chandelier sparkling in the grand foyer
A fluffy white pillow resting on a red velvet couch
A vibrant orange pumpkin sitting on a hay bale
A striped beach umbrella standing tall on a sandy beach
A stainless steel toaster sitting on a marble countertop
A delicate china teacup resting on a lace doily
A pair of neon running shoes waiting by the treadmill
\end{verbatim}

\textbf{Multiple Objects}
\begin{verbatim}
An artist is painting on a blank canvas
A student is typing on his laptop
A young gymnast trains with a balance beam
A chef is making pizza dough in the kitchen
A footballer is kicking a soccer ball
A man is holding an umbrella against rain
A girl is reading a hardcover book in her room
A woman putting lipstick on in front of a mirror
A worker is climbing a ladder to repair a roof
A florist is making a bouquet with fresh flowers
A boy is flying a colorful kite in the sky
A gardener is watering plants with a hose
A photographer is capturing a beautiful butterfly with his camera
A scientist is examining a specimen under a microscope
A drummer is beating the drumsticks on a drum
A fisherman is throwing the fishing rod in the sea
A baby is reaching for a teddy bear on the bed
A candle burns beside an ancient, leather-bound book
Colorful beads are scattered around a small, yellow sewing box
A skateboard leans casually against a brightly-marked graffiti wall
An apple lays nestled next to a vintage, brass pocket watch
Two seashells gleam under the soft light of the moon
A pair of worn ballet shoes rest next to a glossy violin
A quill pen lies across a stack of unmarked parchment paper
A baseball glove is forgotten next to a half-eaten hot dog
A wine bottle and two empty glasses glisten under a chandelier
A magnifying glass sits on top of a mysterious, printed map
A black cat sleeps peacefully beside a carved pumpkin
A weathered straw hat hangs beside a freshly picked sunflower
An open diary lays flat, a single dried rose on its pages
A twinkling star ornament hangs closely with a snow globe
A chessboard is set up, the king and queen standing in opposition
An elegant masquerade mask sits besides a velvet-handled brush
A vintage typewriter shares space with a half-full bottle of whisky
A teddy bear cuddles against a softly worn out children's book
A sandy hourglass and a rugged compass lay side by side
A pair of spectacles lies open on a dog-eared paperback
A tarnished key lies next to a fading photograph
Ripe apples cluster next to a gleaming knife
An empty notebook rests beside a rustling quill
An antique clock sits next to a fleeting hourglass
Weathered boots stand next to a worn hiking stick
A stack of vinyl records leans against an old gramophone
A half-eaten sandwich sits next to a lukewarm thermos
A flickering candle casts a dim light on a dusty mirror
Luxurious perfume bottles sit next to a dainty jewelry box
Burnt-out lightbulbs lie beside a discarded screwdriver
Sandwiched between two hardbound novels lies a bookmark
A broken tablespoon lies next to an empty sugar bowl
A polished flute gleams next to hastily-scrawled sheet music
A wine bottle rests besides a tall glass on a glossy surface
A novel is sprawled open next to a pair of reading glasses
An untouched slice of pizza is cooling beside a half-drunk can of soda
A tangle of yarn rests beside a pair of silver knitting needles
A violin reclines on a chair next to a music sheet filled with notes
An exploded bottle of ink is smeared across a pile of papers
A child's teddy bear is left abandoned near a frequently-read bedtime story book
A brightly lit birthday cake is set near a closed, wrapped gift
A dripping paintbrush stands poised above a half-finished canvas
A silent alarm clock is glaring at a tousled pillow
Two worn paddles rest next to an overturned canoe
A crumpled sheet of paper lies beside a blue ink pen
Two worn-out gym shoes are placed by the front door
An abandoned teddy bear leans against a discarded toy car
A bottle of red wine stands alongside an empty wine glass
An old brass key sits next to an intricate, dust-covered lock
Two silken, colorful scarves are knotted together
A flickering candle stands beside a vintage photograph in a wooden frame
A worn violin and a broken bow rest casually on a chair
A chewed pencil rests alongside a frustration-filled sudoku puzzle
A steaming teapot is standing next to a china teacup with roses
A sprouting potted plant sits comfortably next to a watering can filled with water
A camera with a telephoto lens rests atop a folded map of a city
A novel rests atop a worn-out diary
The golden trophy shines brightly next to a ruffled blue ribbon
An antique clock and a retro radio compete for attention on the dusty shelf
An artist's brush is soaked inwater, awaiting its turn beside a vibrant color palette
A candle, half-burnt, emits an inviting scent near a generous plate of cookies
Two wine glasses, one filled with red, the other white, touch in a graceful toast
An old-fashioned typewriter aspires to outshine a sleek laptop
A well-worn skateboard rests against a gleaming silver bike
A ceramic elephant stands guard over an intricate puzzle yet to be completed
The fragrance of an open perfume bottle mingles with the smell of a single red rose
A pyramid of stacked poker chips dwarfs a pair of playing cards
An oversharpened pencil argues with a ballpoint pen over a stash of blank paper
A colorful array of spices in tiny jars sits next to an unused cooking book
A warm, glowing lantern rests beside a well-read paperback book
A treasure map, crumped and faded, cocooned beside an ornate compass
Two vinyl records rest idly on the sputtering gramophone
An envelope sealed with love stands demurely beside an ancient inkwell and 
    a feather quill
An open book sits beside a vintage brass spectacles
Hot popcorn jump out from the red striped popcorn maker
A rustic cowboy hat hangs on the rugged saddle
A half-eaten slice of pizza forgotten beside a remote control
A dripping paintbrush strokes a vibrant palette of colors
A shimmering diamond necklace lays elegantly on a black velvet box
A thick novel is accompanied by a steaming cup of cocoa
A woolen knitted scarf is wrapped around a carved pumpkin
A scratched surfboard lies in the sandy shadow of a worn-out straw hat
An old brass key longs for the dusted antique chest it could unlock

\end{verbatim}

\subsubsection{GPTEval3D}

\begin{verbatim}
A torn hat
A soft sofa
A rusty boat
A rough rock
A thorny rose
A brick house
A crying sofa
A sleeping cat
A twisted tower
A shouting leaf
A rubbery cactus
A wooden bicycle
A dancing elephant
A pen leaking blue ink
A pair of worn-out shoes
A lamp casting a warm glow
A book with a leather cover
A teddy bear with a red bow
A book left on a park bench
Four ripe apples in a basket
A boat floating on calm water
A chair made from polished oak
A bicycle leaning against a wall
An origami crane made from a map
A mug filled with steaming coffee
A book bound in mysterious symbols
A dog creating sand art on a beach
Three vibrant balloons tied together
A cat with two different colored eyes
An orange tabby cat shaped cookie jar
A quartet of mugs that sing in harmony
Floating bonsai tree, roots in mid-air
Brown rabbit nibbling on a clover leaf
An embroidered silk pillow with tassels
A guitar resting against an old oak tree
A pen sitting atop a pile of manuscripts
A bicycle that leaves a trail of flowers
A weathered hiking backpack with patches
Sand hourglass, sand glitters like stars
A velvet diary, locks with a fingerprint
Spotted ladybug crawling on a green leaf
Gray squirrel with an acorn in its mouth
A teddy bear, fur matted, one eye missing
A golden retriever plush toy, floppy-eare
A carved wooden bear with a salmon in mouth
A cat pondering the mysteries of the universe
A lamp casting shadows on an old, forgotten map
Orange monarch butterfly resting on a dandelion
Clownfish peeking out from sea anemone tendrils
Caterpillar with a keyboard pattern on its back
A quill pen, feather shifts through rainbow hues
Green tree frog clinging to a rain-soaked window
An old-fashioned rotary phone with a tangled cord
Tortoise with a shell that looks like stained glass
A plush octopus whose arms are gently waving pencils
An assortment of vintage, fragrant perfumes on display
A group of vibrant, chattering parrots perched together
Swan with feathers resembling soft, white origami folds
A quaint, little house nestled at the end of a winding path
Flamingo balancing on a sphere instead of standing in water
A bright yellow rubber duck gently floats in a sudsy bathtub
A collection of fresh vegetables arranged in a wicker basket
An ancient, weathered statue, now covered in a blanket of moss
A mesmerizing dance performed by a kaleidoscope of butterflies
A lone, ancient tree stands tall in the middle of a quiet field
A plush teddy bear, sitting alone with a slight tear in its seam
A pair of hiking boots caked with mud at the doorstep of a cabin
A worn leather recliner with a knitted throw draped over the back
A small, intricately carved antique wooden box filled with mystery
Frog with a translucent skin displaying a mechanical heart beating
A small, rustic cabin sits alone in a peaceful, snow-covered forest
Jellyfish with bioluminescent tentacles shaped like lightning bolts
A floating teapot, pouring a stream of endless, steaming jasmine tea
A smoldering campfire under a clear starry night, embers glowing softly
A smartphone with a cracked screen lying on a coffee-stained office desk
An ice cream scoop that serves up scoops of cloud fluff instead of ice cream
A velvet-lined violin case, which opens to reveal a garden of miniature roses
A hammock strung between two skyscrapers, swaying high above a neon cityscape
A delicate porcelain teacup, painted with intricate flowers, rests on a saucer
A sleek red sports car with chrome finishes parked by a bustling city sidewalk
An ensemble of jellyfish-like hanging lamps, pulsing with soft bioluminescence
A cluster of tents pitched near a forest, campfire smoke curling into the evening sky
A colorful kite tangled in the branches of an oak tree, fabric fluttering in the wind
A dragon-shaped kite, with scales that shimmer in the sunlight as it dances in the wind
A stone bridge arching over a babbling brook, 
    encrusted with moss and echoing with stories
A sequence of street lamps, casting pools of light on 
    cobblestone paths as twilight descends
A bouquet of metallic flowers that bloom in response to moonlight, 
    shimmering silver at night
An array of small, solid, symmetrical, pastel-colored eggs, each revealing a miniature, 
    enchanted forest scene when cracked open
A heavy, layered, asymmetrical winter quilt, with a patchwork of plaid fabrics in reds 
    and greens, folded at the foot of a well-made bed
Various hollow, asymmetrical, textured seashells, collected in a sand-filled, 
    clear glass jar with a twine-tied neck, displayed on a windowsill
An old, solid, asymmetrical bronze bell, its contours irregular from centuries of use, 
    with a green patina, sitting silent in an abandoned temple
A small, hollow, asymmetrical birdhouse, painted in cheerful colors, 
    with a round entrance and a tiny perch, swaying gently in a backyard apple tree
An ensemble of hollow, irregularly shaped musical instruments, including a saxophone, 
    a violin, and a drum, resting on a stage before a jazz concert
An assortment of solid, symmetrical, smooth marbles, each one a different color with 
    a unique swirl pattern, scattered playfully across a hardwood floor
A medium-sized, layered, radially symmetrical conch shell, with a rough texture 
    on the outside, fading from pink to cream, sitting alone on a sandy beach
Several large, solid, symmetrical hay bales, with a rough, golden texture, 
    scattered across a rural, open field, with the setting sun casting long shadows
An old, layered, asymmetrical lantern, with a patina copper finish and 
    translucent panes that flicker with bioluminescent light from cultured algae within
A collection of solid, irregularly shaped hand tools, with wooden handles 
    and metal ends, well-used and slightly rusty, hanging on a pegboard in a workshop
Several solid, spherical, weathered cannonballs, with a rough cast-iron texture, 
    lying beside a rusted cannon in a historical fort overlooking a serene bay
A small, solid, geometrically spherical, metallic orb, with a glossy ruby finish, 
    nestled in a nest of black velvet, untouched and gleaming under a spotlight
An intricately carved, solid, wooden figurine, with jagged contours depicting 
    an ancient deity, the wood grain visible under a matte finish, on a stone altar
A large, multi-layered, symmetrical wedding cake, with smooth fondant, delicate 
    piping, and lifelike sugar flowers in full bloom, displayed on a silver stand
A large, hollow, asymmetrically shaped amphitheater, with jagged stone seating, 
    nestled in a natural landscape, a classical play being performed as the sun sets
A medium-sized, hollow, asymmetrical teapot, crafted to look like a slumbering 
    dragon, with a scaly, rough texture and smoke gently wafting from its snout-spout
A solid, smooth, symmetrical porcelain teapot, with a cobalt blue dragon design, 
    steam rising from the spout, suggesting it's just been filled with boiling water
A compact, cylindrical, vintage pepper mill, with a polished, ornate brass body, 
    slightly worn from use, placed beside a porcelain plate on a checkered tablecloth
Several large, solid, cube-shaped parcels, wrapped in brown paper and tied with string, 
    each labeled with a different destination, awaiting dispatch in a post office
An oversized, porous, sphere-shaped birdcage, made of woven golden wires, 
    with a matte finish, housing a small, mechanical, singing bird that 
    flutters in a lifelike manner
A small, solid, radially symmetrical, iridescent abalone shell, with jagged contours, 
    hosting a miniature, tranquil Zen garden complete with tiny, 
    raked sand and micro bonsai
A solid, symmetrical, smooth stone fountain, with water cascading over its edges 
    into a clear, circular pond surrounded by blooming lilies, 
    in the center of a sunlit courtyard
\end{verbatim}

\subsubsection{Diversity prompts}

\begin{verbatim}
A DSLR photo of a green monster truck
A DSLR photo of a corgi puppy
A 20-sided die made out of glass
A DSLR photo of a classic Packard car
A completely destroyed car
A dog made out of salad
A DSLR photo of a bald eagle
A DSLR photo of a bagel filled with cream cheese and lox
A blue poison-dart frog sitting on a water lily
A DSLR photo of a fox taking a photograph using a DSLR
A DSLR photo of a cat lying on its side batting at a ball of yarn
A beagle in a detective’s outfit
A brightly colored mushroom growing on a log
A tarantula, highly detailed
A DSLR photo of a bulldozer
A DSLR photo of a humanoid robot holding a human brain
A DSLR photo of a cat wearing a bee costume
A DSLR photo of a basil plant
A DSLR photo of a bear dressed in medieval armor
A cat with a mullet
A confused beagle sitting at a desk working on homework
A dachshund dressed up in a hotdog costume
A bumblebee sitting on a pink flower
A DSLR photo of a fox holding a videogame controller
An amigurumi motorcycle
A bichon frise wearing academic regalia
A cocktail
A robot made out of vegetables
A toy robot
A DSLR photo of a frog wearing a sweater
A dragon-cat hybrid
A blue tulip
A ceramic lion
A DSLR photo of a chimpanzee dressed like Napoleon Bonaparte
A snail on a leaf
A DSLR photo of a baby dragon hatching out of a stone egg
A 3D model of an adorable cottage with a thatched roof
A sliced loaf of fresh bread
A DSLR photo of a group of dogs eating pizza
A capybara wearing a top hat, low poly
A glowing lantern
A DSLR photo of a ghost eating a hamburger
A banana peeling itself
An edible typewriter made out of vegetables
A delicious hamburger
A ceramic upside-down yellow octopus holding a blue-green ceramic cup
A delicious croissant
A snow terrarium
A high-quality photo of an ice cream sundae
A bunch of colorful marbles spilling out of a red velvet bag
A DSLR photo of a car made out of cheese
A DSLR photo of a group of dogs playing poker
A DSLR photo of a goose made out of gold
A car made out of sushi
A DSLR photo of a chow chow puppy
A DSLR photo of a Christmas tree with donuts as decorations
A pineapple
A DSLR photo of a baby grand piano viewed from far away
A DSLR photo of a cat wearing a lion costume
A DSLR photo of a corgi sneezing
A DSLR photo of a car made out of pizza
A DSLR photo of a hippo wearing a sweater
A cute steampunk elephant
A beautiful rainbow fish
A crocodile playing a drum set
A small saguaro cactus planted in a clay pot
A birthday cake rendered in 3D
The Imperial State Crown of England
A DSLR photo of a knight chopping wood
\end{verbatim}
\clearpage
\twocolumn

\end{document}